\documentclass[twocolumn]{svjour3}          %
\RequirePackage{fix-cm}

\smartqed 

\usepackage{graphicx}
\usepackage{amssymb}
\usepackage{xspace}

\usepackage{newtxtext,newtxmath}

\DeclareMathAlphabet{\mathcal}{OMS}{cmsy}{m}{n}

\usepackage{xcolor}
\usepackage{amsmath}
\usepackage{multirow}
\usepackage{booktabs}
\usepackage{hyperref}
\usepackage{etoolbox}
\usepackage{makecell}
\usepackage{stfloats}

\newcommand{\PAR}[1]{\vskip1pt \noindent {\bf #1~}}

\DeclareMathOperator*{\argmax}{argmax}

\newtheorem{MOTAproblem}{Problem with MOTA}
\newtheorem{IDFproblem}{Problem with IDF1}
\newtheorem{mAPproblem}{Problem with Track-mAP}

\newcommand*{\eg}{\emph{e.g.}\@\xspace}

\newcommand*{\etal}{\emph{et al.}\@\xspace}

\newcommand*{\wrt}{\emph{w.r.t}\@\xspace}

\makeatletter
\renewcommand\subsection{\@startsection{subsection}{2}{\z@}%
    {-21dd plus-8pt minus-4pt}{10.5dd}
     {\normalsize\bfseries\boldmath\upshape}}
\makeatother

\journalname{International Journal of Computer Vision}
\begin{document}
\emergencystretch 3em

\title{HOTA: A Higher Order Metric for Evaluating Multi-Object Tracking}

\author{Jonathon Luiten         \and
         Aljo\u{s}a O\u{s}ep         \and
         Patrick Dendorfer   \and
        Philip Torr         \and
        Andreas Geiger         \and
        Laura Leal-Taix\'{e}         \and          
        Bastian Leibe     
}

\institute{J. Luiten, B. Leibe \at
             RWTH Aachen University, Germany \\
              \email{\{luiten,leibe\}@vision.rwth-aachen.de}
           \and
           A. O\u{s}ep, P. Dendorfer, L. Leal-Taix\'{e} \at
             Technical University Munich, Germany \\
              \email{\{aljosa.osep,patrick.dendorfer,leal.taixe\}@tum.de}
           \and
           P. Torr \at
             University of Oxford, UK \\
              \email{phst@robots.ox.ac.uk}
           \and
           A. Geiger \at
             Max Planck Institute for Intelligent Systems, T{\"u}bingen, and University of T{\"u}bingen, Germany  \\
              \email{andreas.geiger@tue.mpg.de}
}

\date{Received: date / Accepted: date}

\maketitle

\begin{abstract}
Multi-Object Tracking (MOT) has been notoriously difficult to evaluate.
Previous metrics overemphasize the importance of either detection or association.
To address this, we present a novel MOT evaluation metric, HOTA (Higher Order Tracking Accuracy), which explicitly balances the effect of performing accurate detection, association and localization into a single unified metric for comparing trackers.
HOTA decomposes into a family of sub-metrics which are able to evaluate each of five basic error types separately, which enables clear analysis of tracking performance.
We evaluate the effectiveness of HOTA on the MOTChallenge benchmark, and show that it is able to capture important aspects of MOT performance not previously taken into account by established metrics. Furthermore, we show HOTA scores better align with human visual evaluation of tracking performance.\footnote{Pre-print. Accepted for Publication in the International Journal of Computer Vision, 19 August 2020.\newline Code is available at \newline \mbox{\url{https://github.com/JonathonLuiten/HOTA-metrics}.}}

\end{abstract}

\begin{figure}[t!]
\centering
\includegraphics[width=\linewidth]{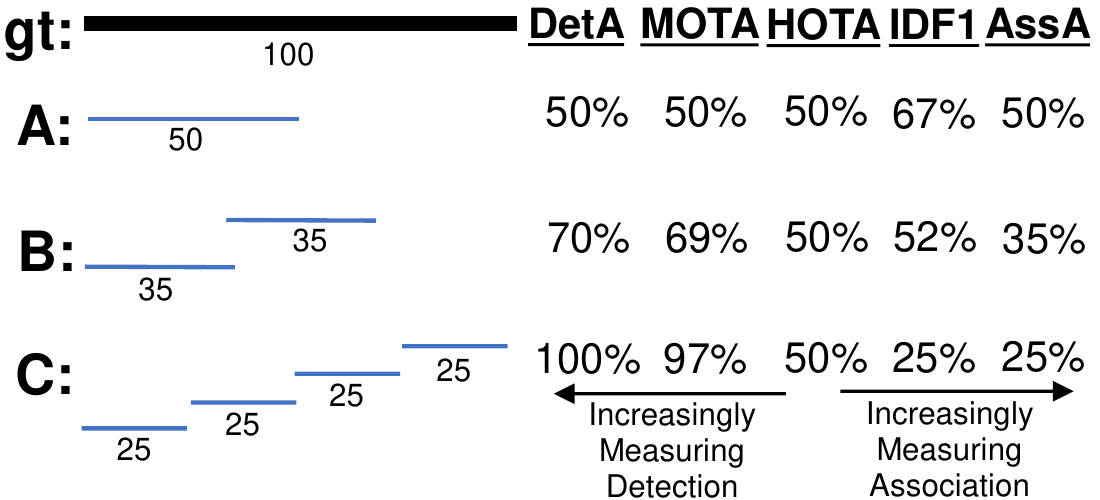}
\caption{A simple tracking example highlighting one of the main differences between evaluation metrics. Three different trackers are shown in order of increasing detection accuracy and decreasing association accuracy. MOTA and IDF1 overemphasize the effect of accurate detection and association respectively. HOTA balances both of these by being an explicit combination of a detection score DetA and an association score AssA.}
\label{fig:1stpage}
\end{figure}

\section{Introduction}

Multi-Object Tracking (MOT) is the task of detecting the presence of multiple objects in video, and associating these detections over time according to object identities.
The MOT task is one of the key pillars of computer vision research, and is essential for many scene understanding tasks such as surveillance, robotics or self-driving vehicles. 
Unfortunately, the evaluation of MOT algorithms has proven to be very difficult. MOT is a complex task, requiring accurate detection, localisation, and association over time. 

This paper defines a metric, called HOTA (Higher Order Tracking Accuracy), which is able to evaluate all of these aspects of tracking. We provide extended analysis as to why HOTA is often preferable to current alternatives for evaluating MOT algorithms.
As can be seen in Fig.~\ref{fig:1stpage}, currently used metrics MOTA \cite{CLEARMOT} and IDF1 \cite{IDF1} overemphasize detection and association respectively. HOTA explicitly measures both types of errors and combines these in a balanced way. HOTA also incorporates measuring the localisation accuracy of tracking results which isn't present in either MOTA or IDF1. 

HOTA can be used as a single unified metric for ranking trackers, while also decomposing into a family of sub-metrics which are able to evaluate different aspects of tracking separately. This enables clear understanding of the different types of errors that trackers are making and enables trackers to be tuned for different requirements.

The HOTA metric is also intuitive to understand. This can be seen clearly in Fig.~\ref{fig:1stpage}. The detection accuracy, DetA, is simply the percentage of aligning detections. The association accuracy, AssA, is simply the average alignment between matched trajectories, averaged over all detections. The final HOTA score is the geometric mean of these two scores averaged over different localisation thresholds.

In this paper we make four major novel contributions:
(i) We propose HOTA as a novel metric for evaluating multi-object tracking (Sec.~\ref{sec:HOTA});
(ii) We provide thorough theoretical analysis of HOTA as well as previously used metrics MOTA, IDF1 and Track-mAP, highlighting the benefits and shortcomings of each metric (Sec.~\ref{sec:analysis} and \ref{sec:comparison});
(iii) We evaluate HOTA on the MOTChallenge benchmark and analyse its properties compared to other metrics for evaluating current state-of-the-art trackers (Sec.~\ref{sec:evaluation});
(iv) We perform a thorough user-study comparing how different metrics align with human judgment of tracking accuracy and show that HOTA aligns closer with the desired evaluation properties of users compared to previous metrics (Sec.~\ref{sec:humanstudy}).

\section{Related Work}
\label{sec:related}

\PAR{Early History of MOT Metrics.}
Multi-Object tracking has a long history dating back to at least the 70s \cite{70sTrack1,70sTrack2,70sTrack3,70sTrack4}. 
Early work tended to evaluate using their own simple evaluation metrics, such that comparison wasn't possible between groups.
In the early 2000s a number of different groups sought to define standard MOT evaluation metrics. This included the PETS (Performance Evaluation of Tracking and Surveillance) workshop series \cite{PETSmetrics}, the VACE (Video Analysis and Content Extraction) program \cite{VACE}, and a number of other groups \cite{CHIL,AMI,ETISEO}. In 2006, the CLEAR (CLassification of Events, Activities and Relationships) workshop \cite{CLEARworkshop} brought together all of the above groups and sought to define a common and unified framework for evaluating MOT algorithms. This became the CLEAR MOT metrics \cite{CLEARMOT} which positions the MOTA metric as the main metric for tracking evaluation alongside other metrics such as MOTP. MOTA was adopted for evaluation in the PETS workshop series \cite{PETS} and remains, to this day, the most commonly used metric for evaluating MOT algorithms, although it has often been highly criticised \cite{GMME,OSPA-ST,Monotonicity,TrackingTrackers,TrackingChallenges,IDF1,TAO,MOTSFusion,sas_mot,etc,NonMarkovianGloballyConsistantObjectTracking,SolutionPath,MOT20,MOTlitreview} for its bias toward overemphasizing detection over association (see Fig. \ref{fig:1stpage}), as well as a number of other issues (see Sec. \ref{sec:comparison}).

\PAR{Benchmarks' use of Metrics.}
In the last five years the two most commonly used benchmarks for evaluating MOT have been the MOTChallenge \cite{MOT15,MOT16,MOT19} and KITTI \cite{KITTI} benchmarks. Both of these have ranked trackers using the MOTA metric, contributing to the general widespread use of MOTA in the community. 

Within the last few years the multi-object tracking community has grown enormously due in part to large investment from the autonomous vehicle industry. This has resulted in a large number of new MOT benchmarks being proposed. Many of these rank trackers using the MOTA metric (PANDA \cite{PANDA}, BDD100k \cite{BDD}, Waymo \cite{Waymo}, ArgoVerse \cite{Argoverse}, PoseTrack \cite{PoseTrack}, MOTS \cite{MOTS}), or a variation of MOTA (nuScenes \cite{nuScenes}, UA-DETRAC \cite{UADETRAC}).

Two other metrics have recently been adopted by some MOT benchmarks. The IDF1 metric \cite{IDF1} which was proposed specifically for tracking objects through multiple cameras has been used by `multi-camera MOT' benchmarks such as Duke-MTMC \cite{IDF1}, AI City Challenge \cite{AICityChallenge} and LIMA \cite{LIMA}. IDF1 has also recently been implemented as a secondary metric on the MOTChallenge benchmark, and has become preferred over MOTA for evaluation by a number of single camera tracking methods \cite{sas_mot,NonMarkovianGloballyConsistantObjectTracking,etc} due to its focus on measuring association accuracy over detection accuracy (see Fig.\ref{fig:1stpage}). IDF1 however exhibits unintuitive and non-monotonic behaviour in regards to detection (see Sec. \ref{sec:comparison}). 

The Track-mAP metric (also called 3D-IoU) was introduced for tracking evaluation on the ImageNet-Video benchmark \cite{ImageNet}. Recently it has been adapted by a number of benchmarks such as YouTube-VIS \cite{VIS}, TAO \cite{TAO} and VisDrone \cite{VisDrone}. Track-mAP differs from previously described metrics in that it doesn't operate on a set of fixed tracks, but rather requires a set of tracks ranked by the tracker's confidence that each track exists. This makes Track-mAP incompatible with many of the commonly used benchmarks (such as MOTChallenge and KITTI) which do not require trackers to output a confidence score with their predictions. Track-mAP suffers from many of the same drawbacks as IDF1 due to its use of global track-based matching, while also having a number of other drawbacks related to the use of ranking-based matching (See Sec. \ref{sec:comparison}).

A number of extensions to the MOTA metric, such as PR-MOTA \cite{UADETRAC} and AMOTA \cite{AMOTA}, have previously been proposed to adapt the MOTA metric to handle confidence ranked tracking results as is done in Track-mAP. We present a simple extension to our HOTA metric in Sec. \ref{sec:extensions} which similarly extends HOTA to confidence ranked results, and which reduces to the standard HOTA metric when taking a fixed set of detections above a certain confidence threshold.

A number of other metrics \cite{GMME,OSPA-ST,MDAM,InfoMetric,PurityMetric,TrackClustering} have been been proposed for MOT evaluation, but to the best of our knowledge none of them have been adopted by any MOT benchmarks and thus have not become widely used for evaluation. 

Other metrics such as the trajectory-based metrics (Mostly-Tracked, Partially-Tracked, Mostly-Lost, Fragmentation) \cite{TrajectoryOrig,TRAJECTORY}, and the metrics of Leichter and Krupka (False Negative Rate, False Positive Rate, Fragmentation Index, Merger Index, Mean Deviation) \cite{Monotonicity} are commonly shown as secondary metrics on benchmarks but are never used to comprehensively rank trackers as they are too simple, often focusing on only a single type of error each, and easy to be gamed if desired.

\PAR{Meta-Evaluation of Metrics.}
PETS vs VACE \cite{PETSvsVACE} discusses the trade-off between presenting multiple evaluation metrics vs a single unifying metric. Their conclusion is that multiple metrics are useful for researchers to debug algorithms and identify failure components, while a unified metric is useful for end-users wishing to easily choose highly performant trackers from a large number of options. We resolve this conflict by presenting both a unified metric, HOTA, and its decomposition into simple components which can be used to analyse different aspects of tracking behaviour (See Sec. \ref{sec:decomposition}).

Milan \etal \cite{TrackingChallenges} analyse the tracking metrics available in 2013 (CLEAR MOT \cite{CLEARMOT} and Trajectory-based Metrics \cite{TrajectoryOrig}) and identify a number of deficits present in these metrics, particularly in MOTA. Based on this analysis they find none of the metrics that were analysed to be suitable as a single unified metric and recommend to present results over all available metrics. We present HOTA as a solution to such issues, as a metric suitable for unified comparison between trackers.

Leichter and Krupka \cite{Monotonicity} present a theoretical framework for analysing MOT evaluation metrics, which involves two components. The first is a characterisation of five error types that can occur in MOT (False negatives, False positives, Fragmentation, Mergers and Deviation). The second component is the description of two fundamental properties that MOT evaluation metrics should have: monotonicity and error type differentiability. In \cite{Monotonicity}, they show that all previous metrics (including MOTA) don't have either of these properties. They propose a set of five separate simple metrics, one for each error type, however they make no effort to combine these into one unified metric, and thus the usefulness of these metrics in comparing trackers is limited.
In Sec.~\ref{sec:decomposition} we show how HOTA can be decomposed into components which correspond to each of these five error types (Detection Recall, Detection Precision, Association Recall, Association Precision and Localisation Accuracy, respectively), and as such HOTA has the property of error type differentiability which requires that the metrics are informative about the tracker's performance with respect to each of the different basic error types. 
In Sec.~\ref{sec:analysis} we show how the combined HOTA metric is strictly monotonic with regards to each of these five types of errors, thus having the second required property.
In Sec.~\ref{sec:comparison} we show that no other recently used metric has these desirable properties.

Tracking the Trackers \cite{TrackingTrackers} sought to analyse different evaluation metrics using human evaluators, in order to find out how well different metrics reflect human perception of the quality of tracking results. They only evaluated the set of CLEAR MOT \cite{CLEARMOT} and trajectory-based \cite{TrajectoryOrig,TRAJECTORY} metrics. From these metrics they found that MOTA remains the most representative measure that coincides to the highest degree with human visual assessment, despite pointing out its many limitations. 
In Sec.~\ref{sec:humanstudy} we seek to repeat this study and compare our HOTA metric with MOTA and IDF1 using human visual evaluation. We find that HOTA performs much better in this user-study than both MOTA and IDF1, particularly among MOT researchers.

\section{Preliminaries}
\label{sec:terminology}

In this section, we lay the framework needed for understanding the content of this paper. This includes describing the multi-object task, the role of evaluation metrics, the ground-truth and prediction representation and the notation we will be using, as well as other fundamental concepts that will be used throughout the paper.

\PAR{What is Multi-Object Tracking?}
Multi-Object Tracking (MOT) is one of the core tasks for understanding objects in video. 
The input to the MOT task is a single continuous video (although we present extensions for our metric for multi-camera tracking in Sec.~\ref{sec:extensions}), split up into discrete frames at a certain frame rate. Each discrete frame could be an image, or could be another representation such as point cloud from a laser scanner. 

The output of the MOT task is a representation that encodes the information about: (a) what objects are present in each frame (detection), where they are in each frame (localisation) and whether objects in different frames belong to the same or different objects (association).

\PAR{Evaluation Metrics and Ground-Truth.}
In order to evaluate how well a tracker performs, we need to compare its output to a ground-truth set of tracking results. The purpose of an evaluation metric is to evaluate the similarity between the given ground-truth and the tracking results. This is not a well defined problem, as there are many different ways of scoring such a similarity (especially between complex representations such as sets of trajectories). However, the choice of evaluation metric is extremely important, as the properties of the metric determine how different errors contribute to a final score, as such it is favourable that metrics have certain properties. The choice of metric also has the ability to heavily influence the direction of research within the research community. In the age of competitive benchmarks, a lot of research (for better or for worse) is evaluated on its ability to improve the scores on these benchmarks. If benchmarks are using metrics to evaluate these scores which are biased towards only certain aspects of a task, this will also bias research and methods towards focusing more on these aspects.

\PAR{MOT Ground-Truth and Prediction Format.}
The set of ground-truth tracks is represented as a set of detections (\mbox{gtDets}) in each video frame, where each gtDet is assigned an id (gtID), such that the gtIDs are unique within each frame and the gtIDs are consistent over time for detections from the same ground-truth object trajectory (gtTraj).

For most evaluation metrics (MOTA, IDF1 and HOTA) a tracker's prediction is in the same format as the ground-truth data. It consists of a set of predicted detections (prDets) in each frame, each assigned a predicted id (prID), such that the prIDs are unique within each frame and consistent over time for detections from the same predicted object trajectory (prTraj).

For the Track-mAP metric, in addition to the prDets with PrIDs as above, each prTraj is assigned a confidence score estimating how likely it is that this trajectory exists. If each prDet is assigned a confidence score instead of each prTraj, the confidence score for the prTraj is simply the average of the confidence scores over the prDets that belong to the prTraj.

For tracking multiple object classes, each gtTraj and prTraj may also be assigned a class id (gtCl / prCl). Previous metrics have all been applied per class and averaged over classes. Thus for simplicity we can ignore this class id when defining metrics and assume that metrics are calculated only over a single class of objects at a time. However, we also present an extension to HOTA in Sec.~\ref{sec:extensions} which explicitly deals with multi-class tracking. 

\PAR{Types of Tracking Errors.}
The potential errors between a set of predicted and ground-truth tracks can be classified into three categories: detection errors, localisation errors and association errors \cite{Monotonicity}. Detection errors occur when a tracker predicts detections that don't exist in the ground-truth, or fails to predict detections that are in the ground-truth.
Association errors occur when trackers assign the same prID to two detections which have different gtIDs, or when they assign different prIDs to two detections which should have the same gtID.
Localisation errors occur when prDets are not perfectly spatially aligned with gtDets. 

There are other ways of defining basic error types for MOT, such as identification errors \cite{IDF1}. However detection, association and localisation errors are the most commonly used error types \cite{Monotonicity} and are widely applicable for evaluating tracking for a wide range of different tracking scenarios (see Sec.~\ref{sec:decomposition} for examples). As such we only consider these error types when analysing and comparing evaluation metrics. 

\PAR{Different Detection Representations.}
Object detections within each frame (gtDets and prDets) may have a number of different representations depending on the domain and application. Commonly used representations include 2D bounding boxes \cite{MOT15,KITTI}, 3D bounding boxes \cite{Waymo,Argoverse}, segmentation masks \cite{MOTS,VIS}, point estimates in 2D or 3D \cite{PETS}, and human pose skeletons \cite{PoseTrack}. In general tracking evaluation metrics are agnostic to the specific representation except for the need to define measures of similarity for each representation. 

\PAR{Measures of Similarity.}
Most metrics require the definition of a similarity score between two detections, $\mathcal{S}$. Track-mAP requires such a similarity score between trajectories, $\mathcal{S}_{\text{tr}}$.

This similarity score should be chosen based on the detection representation used, but should be constrained to be between $0$ and $1$, such that when $\mathcal{S}$ is $1$ the prDet and gtDet perfectly align, and when $\mathcal{S}$ is $0$ there is no overlap between detections.
The most commonly used similarity metric for 2D boxes, 3D boxes and segmentation masks is $\text{IoU}_{\text{Loc}}$, which is the spatial intersection of two regions divided by the union of the regions. For point representations (and human joint locations), a score of one minus the Euclidean distance is often used, such that points are said to have zero overlap when they are more than one meter apart.

\PAR{Bijective Matching.}
A common procedure in MOT evaluation metrics (MOTA, IDF1, HOTA) is to perform a bijective (one-to-one) matching between gtDets and prDets. This ensures that all gtDets are matched to at most one prDet and vice versa, and that any extra or missed predictions are penalised. Such a bijective mapping can be easily calculated by calculating a matching score between all pairs of gtDet and prDet and using the Hungarian algorithm for finding the matching that optimises the sum of the matching score. Usually there is a minimum similarity $\mathcal{S} \geq \alpha$ requirement for a match to occur.
After matching has occurred, we have some gtDets and prDets that are matched together. We call these pairs as true positives (TP). These are considered correct predictions. Any gtDets that are not matched (missed) are false negatives (FN). Any prDets that are not matched (extra predictions) are false positives (FP). FNs and FPs are two types of incorrect predictions.

\PAR{Matching vs Association.}
The words \textit{match} and \textit{association} are used throughout this paper to refer to two different things. 
We refer to a \textit{match} as a pair consisting of a matching ground-truth detection and a predicted detection. 
On the other hand \textit{association} refers to a number of detections with the same ID such that they are associated to the same trajectory.

\PAR{Jaccard Index and IoU.}
The Jaccard Index is a measure of the similarity between two sets. It is defined as follows:
\begin{equation}
\text{Jaccard Index} = \frac{|\text{TP}|}{|\text{TP}| + |\text{FN}| + |\text{FP}|}
\end{equation}
The Jaccard index is commonly called the IoU (intersection over union) because it is calculated as the intersection of the two discrete sets divided by their union. However, in tracking IoU is also often used for describing the spatial overlap between two spatial regions (\eg boxes, masks). In order to not confuse these terms we use `Jaccard index' to refer to the operation over discrete sets, and $\text{IoU}_{\text{Loc}}$ to refer to the operation over spatial regions.

\PAR{Mathematical Metric Definition.}
The term \textit{metric} has a strict mathematical definition.
For a distance measure (such as HOTA) between two sets (\eg, ground-truth tracks and predicted tracks) to be strictly a \textit{metric} in the mathematical sense, it must satisfy three conditions, (i) identity of indiscernibles, (ii) symmetry, and (iii) subadditivity (the triangle inequality). 

Within the computer vision community, the term metric is commonly used for functions which calculate a score by which algorithms can be ranked, without requiring the three conditions above. We use this definition throughout this paper, and for example, refer to MOTA as an evaluation metric even though it doesn't meet the last two requirements.

Note that for the purpose of evaluating the MOT task, it is not strictly necessary for metrics to be symmetric or subadditive. However, as we show in Sec.~\ref{sec:analysis} these are both useful properties to have. Of the commonly used metrics, HOTA is the only one to have these properties and thus technically be mathematically a \textit{metric}.

\section{Overview of Previous Metrics}
\label{sec:previous}

In this section we provide a brief overview of the calculation of the three metrics which are currently used by MOT benchmarks (see Sec.~\ref{sec:related}). Our HOTA metric builds upon many of the ideas of previous metrics (especially MOTA), and as such it is important for the reader to have an overview of how they work. 
An analysis comparing the properties and deficits of all of the methods is presented in Sec.~\ref{sec:comparison}, and Tab.~\ref{table:metricsoverview} gives an overview of the different design decisions between the metrics.
\subsection{CLEARMOT: MOTA and MOTP}
\PAR{Matching Predictions and Ground-Truth.}
In MOTA (Multi-Object Tracking Accuracy), matching is done at a detection level. A bijective (one-to-one) mapping is constructed between prDets and gtDets in each frame. Any prDets and gtDets that are matched (correct predictions) become true positives (TPs). Any remaining prDets that are not matched (extra predictions) become false positives (FPs). Any gtDets that are not matched (missing predictions) become false negative (FNs). prDets and gtDets can only be matched if they are adequately spatially similar. MOTA thus requires the definition of a similarity score, $\mathcal{S}$, between detections (e.g. $\text{IoU}_{\text{Loc}}$ for 2D bounding boxes), and the definition of a threshold, $\alpha$, such that detections are only allowed to match when $\mathcal{S} \geq \alpha$.
In practice, multiple matches could occur, the actual matching is performed such that the final MOTA and MOTP scores are optimised (see below).

\PAR{Measuring Association.}
In MOTA, association is measured with the concept of an Identity Switch (IDSW).
An IDSW occurs when a tracker wrongfully swaps object identities or when a track was lost and reinitialised with a different identity. 
Formally, an IDSW is a TP which has a prID that is different from the prID of the previous TP (that has the same gtID). 
IDSWs only measure association errors compared to the single previous TP, and don't count errors where the same prID swaps to a different gtID (ID Transfer).

\PAR{Scoring Function.}
MOTA measures three types of tracking errors. The detection errors of FNs and FPs, as well as the association error of IDSW.
The final MOTA score is calculated by summing up the total number of these errors, dividing by the number of gtDets, and subtracting from one.
\begin{align}
\textrm{MOTA} = 1 - \frac{|\textrm{FN}| + |\textrm{FP}| + |\textrm{IDSW}|}{|\textrm{gtDet}|}
\label{eq:MOTA}
\end{align}
\PAR{MOTP.}
Note that MOTA doesn't include a measure of localisation error. Instead the CLEAR MOT metrics define a secondary metric, MOTP (Multi-Object Tracking Precision), which solely measures localisation accuracy. It is simply the average similarity score, $\mathcal{S}$, over the set of TPs.
\begin{align}
\textrm{MOTP} = \frac{1}{|\text{TP}|}\sum_{\text{TP}}{ \mathcal{S}}
\end{align}
\PAR{Matching to Optimise MOTA and MOTP.}
The matching of prDets to gtDets is performed so that the final MOTA and MOTP scores are maximised. 
This is implemented, in each new frame, by first fixing matches in the current frame which have $\mathcal{S} \geq \alpha$ and don't cause an IDSW. For the remaining potential matches the Hungarian algorithm is run to select the set of matches that as a first objective maximises the number of TPs, and as a secondary objective maximises the mean of $\mathcal{S}$ across the set of TPs.
\subsection{The Identification Metrics: IDF1}
\PAR{Matching Predictions and Ground-Truth.}
IDF1 calculates a bijective (one-to-one) mapping between the sets of gtTrajs and prTrajs (unlike MOTA which matches at a detection level).
This defines new types of detection matches. IDTPs (identity true positives) are matches on the overlapping part (where $\mathcal{S} \geq \alpha$) of trajectories that are matched together. IDFNs (identity false negatives) and IDFPs (identity false positives) are the remaining gtDets and prDets respectively, from both non-overlapping sections of matched trajectories, and from the remaining trajectories that are not matched.
\PAR{Scoring Function.}
The ID-Recall, ID-Precision and IDF1 scores are calculated as follows:
\begin{align}
&\text{ID-Recall} = \frac{|\text{IDTP}|}{|\text{IDTP}| + |\text{IDFN}|} \\
&\text{ID-Precision} = \frac{|\text{IDTP}|}{|\text{IDTP}|  + |\text{IDFP}|} \\
&\text{IDF1} = \frac{|\text{IDTP}|}{|\text{IDTP}| + 0.5 \, |\text{IDFN}| + 0.5 \, |\text{IDFP}|}
\end{align}
\PAR{Matching to Optimise IDF1.}
The matching is performed such that IDF1 is optimised. This is performed by enumerating the number of IDFPs and FDFNs that would result from each match (non-overlapping sections), and from non-matched trajectories (number of prDet and gtDet in these trajectories, respectively). The Hungarian algorithm is used to select which trajectories to match so that the sum of the number of IDFPs and IDFNs is minimised. Note that the localisation accuracy is not minimised during IDF1 matching unlike in MOTA.

\subsection{Track-mAP}
\PAR{Matching Predictions and Ground-Truth.}
Track-mAP (mean average precision) matches predictions and ground-truth at a trajectory level. It requires the definition of a trajectory similarity score, $\mathcal{S}_{\text{tr}}$, between trajectories (in contrast to MOTA and IDF1 which use a detection similarity score, $\mathcal{S}$), as well as a threshold $\alpha_{\text{tr}}$ such that trajectories are only allowed to match if $\mathcal{S}_{\text{tr}} \geq \alpha_{\text{tr}}$.
A prTraj is matched with a gtTraj if it has the highest confidence score of all prTrajs with $\mathcal{S}_{\text{tr}} \geq \alpha_{\text{tr}}$. If one prTraj should match with more than one gtTraj, it is matched with the one for which it has the highest  $\mathcal{S}_{\text{tr}}$, and the other gtTrajs can be matched by the prTraj with the next highest confidence score. We define the matched prTrajs as true positive trajectories (TPTr), and the remaining prTrajs as false positive trajectories (FPTr).
\PAR{Trajectory Similarity Scores.}
$\mathcal{S}_{\text{tr}}$ is commonly defined for 2D bounding box tracking in two different ways.

In \cite{ImageNet,VisDrone}, the set of TPs is defined as pairs of detections in the trajectories where $\mathcal{S} \geq \alpha$ (with $\mathcal{S}$ being $\text{IoU}_{\text{Loc}}$, and $\alpha$ being 0.5). FNs and FPs are the remaining gtDets and prDets respectively. $\mathcal{S}_{\text{tr}}$ is then equal to $|\text{TP}| / (|\text{TP}| + |\text{FN}| + |\text{FP}|)$.

In \cite{VIS,TAO}, $\mathcal{S}_{\text{tr}}$ is defined as the sum of the spatial intersection of the boxes across the whole trajectories, divided by the sum of the spatial union of the boxes across the whole trajectories.
\PAR{Scoring Function.}
Track-mAP follows the calculation of the average precision metric \cite{pascalvoc,ImageNet,COCO} over trajectories (instead of detections as commonly used for evaluating object detection).

PrTrajs are ordered by decreasing confidence score. Let the index of this ordering (starting at one) be $n$. Let the number of TPTrs in this list up to index $n$ be $|$TPTr$|_n$. For each value of $n$, the precision (Pr$_n$) and recall (Re$_n$) can be calculated as:
\begin{align}
&\text{Pr}_n = \frac{|\text{TPTr}|_n}{n} \\
&\text{Re}_n = \frac{|\text{TPTr}|_n}{|\text{gtTraj}|}
\end{align}
The precision values are interpolated (InterpPr$_n$) so that they are monotonically decreasing.
\begin{align}
&\text{InterpPr}_n = \max_{m \geq n}(\text{Pr}_m)
\end{align}
The Track mAP score is then the integral under the interpolated precision-recall curve created by plotting InterpPr$_n$ against Re$_n$ for all values of $n$. This integral is approximated by averaging over a number of fixed recall values.
\PAR{Threshold and Class Averaging.}
Track-mAP is sometimes calculated at a fixed value of $\alpha_{\text{tr}}$ \cite{TAO}, and sometimes averaged over a range different $\alpha_{\text{tr}}$ values \cite{VIS}.

When multiple classes are to be tracked, Track-mAP is usually calculated per class separately and then the final score is averaged over the classes.

\begin{figure*}[t!]
\centering
\includegraphics[width=0.7\linewidth]{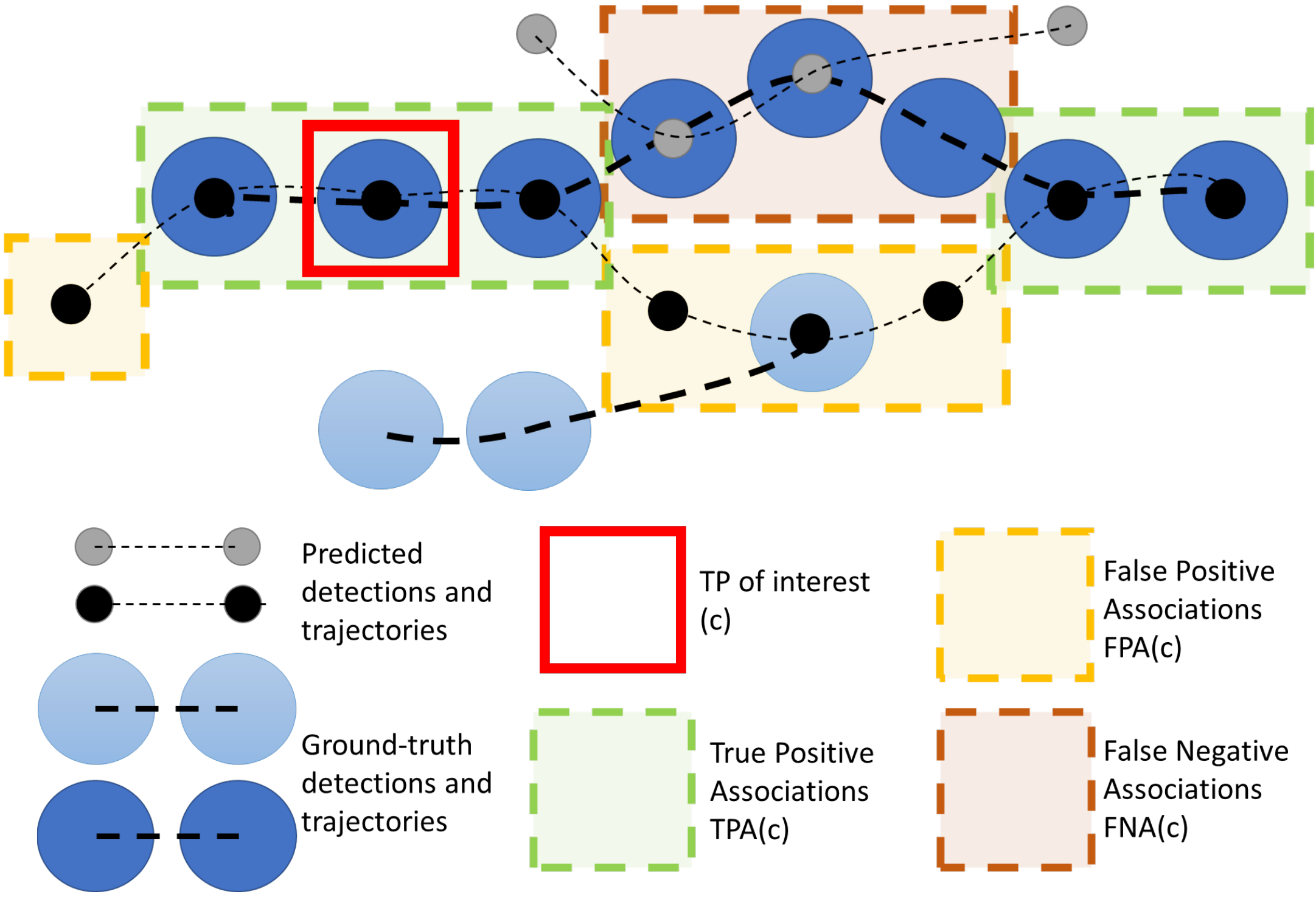}
\caption{A visual explanation of the concepts of TPA, FPA and FNA. The different TPAs, FPAs and FNAs are highlighted for the TP of interest c. 
The TPAs (green) for c (red) are the matches which have the same prID and the same gtID. The FPAs have the same prID but either a different or no gtID. The FNAs have the same gtID but either a different or no prID. In the diagram, $c$ has five TPAs, four FPAs and three FNAs. Conceptually these concepts are trying to answer the question: \textit{For the matched TP $c$, how accurate is the alignment between the gtTraj for this TP (large dark blue circles) and the prTraj for this TP (small black circles).}
}
\label{ass-explain}
\end{figure*}

\section{The HOTA Evaluation Metric}
\label{sec:HOTA}
The main contribution of this paper is a novel evaluation metric for evaluating Multi-Object Tracking (MOT) performance. 
We term this evaluation metric HOTA (Higher Order Tracking Accuracy).
HOTA builds upon the previously used MOTA  metric (Multi-Object Tracking Accuracy) \cite{CLEARMOT}, while addressing many of its deficits.

HOTA is designed to: (i) provide a single score for tracker evaluation which fairly combines all different aspects of tracking evaluation, (ii) evaluate long-term higher-order tracking association, and finally, (iii) decompose into sub-metrics which allow analysis of the different components of tracker's performance.

In this section, we provide a definition of the HOTA metric.
In Sec.~\ref{sec:decomposition} we show how HOTA can be decomposed into a set of sub-metrics which can be used to analyse different aspects of tracking performance.
In Sec.~\ref{sec:analysis} we analyse different properties of HOTA, and examine the design decisions inherent to the HOTA formulation.
In Sec.~\ref{sec:comparison} we present a comparison of HOTA to MOTA, IDF1 and Track-mAP, and show how HOTA addresses many of the deficits of previous metrics.
\PAR{Matching Predictions and Ground-Truth.}
In HOTA, matching occurs at a detection level (similar to MOTA). 
A true positive (TP) is a pair consisting of a gtDet and a prDet, for which the localisation similarity $\mathcal{S}$ is greater than or equal to the threshold $\alpha$.
A false negative (FN) is a gtDet that is not matched to any prDet.
A false positive (FP) is a prDet that is not matched to any gtDet.
The matching between gtDets and prDets is bijective (one-to-one) in each frame. Multiple different combinations of matches could occur, the actual matching is performed to maximise the final HOTA score (see below).
\PAR{Measuring Association.}
The concepts of TPs, FNs and FPs are commonly used to measure detection accuracy. 
In order to evaluate the success of association in a similar way, we propose the novel concepts of TPAs (True Positive Associations), FNAs (False Negative Associations) and FPAs (False Positive Associations), which are defined for each TP. 
For a given TP, $c$, the set of TPAs is the set of TPs which have both the same gtID and the same prID as $c$:
\begin{align}
\begin{aligned}
\text{TPA}&(c) = \{k\}, \\
&k \in \{ \text{TP} \, | \, \text{prID}(k) = \text{prID}(c) \land \text{gtID}(k) = \text{gtID}(c) \}
\end{aligned}
\label{eq:TPA}
\end{align}
For a given TP, $c$, the set of FNAs is the set of gtDets with the same gtID as $c$, but that were either assigned a different prID as $c$, or no prID if they were missed:
\begin{align}
\begin{aligned}
\text{FNA}&(c) = \{k\}, \\
\multirow{3}{*}{ $k \in$ }  & \{ \text{TP} \, | \, \text{prID}(k) \neq \text{prID}(c) \land \text{gtID}(k) = \text{gtID}(c) \} \\ & \cup \,  \{ \text{FN} \, | \, \text{gtID}(k) = \text{gtID}(c) \}
\end{aligned}
\label{eq:FNA}
\end{align}
Finally, for a given TP, $c$, the set of FPAs is the set of prDets with the same prID as $c$, but that were either assigned a different gtID as $c$, or no gtID if they did not actually correspond to an object:
\begin{align}
\begin{aligned}
\text{FPA}&(c) = \{k\}, \\
\multirow{3}{*}{ $k \in$ }  & \{ \text{TP} \, | \, \text{prID}(k) = \text{prID}(c) \land \text{gtID}(k) \neq \text{gtID}(c) \} \\ & \cup \,  \{ \text{FP} \, | \, \text{prID}(k) = \text{prID}(c) \}.
\end{aligned}
\label{eq:FPA}
\end{align}
A visual example explaining the concept of TPAs, FNAs and FPAs is shown in Fig.~\ref{ass-explain}.

Note that although TPAs, FPAs, and FNAs are measured between pairs of detections, these measures can be easily and efficiently calculated by counting the
number of matches per each prID-gtID pair, and there is no need to explicitly iterate over all pairs of detections.
\PAR{Scoring Function.}
Now we have defined concepts used to measure successes and errors in detection (TPs, FPs, FNs) and association (TPAs, FPAs, FNAs), we can define the HOTA$_\alpha$ score for a particular localisation threshold $\alpha$:
\begin{align}
   &\text{HOTA}_{\alpha} = \sqrt{\frac{\sum_{c \in \{\text{TP}\}} \mathcal{A}(c) }{|\text{TP}| + |\text{FN}| + |\text{FP}|}} &
   \label{equ:hotaalpha} \\
    &\mathcal{A}(c) = \frac{|\text{TPA}(c)|}{|\text{TPA}(c)| + |\text{FNA}(c)| + |\text{FPA}(c)|}
    \label{eq:assoc} &
\end{align}

We call this a `double Jaccard' formulation, where a typical Jaccard metric is used over detection concepts of TPs/FPs/FNs with each of the TPs in the numerator being weighted by an association score $\mathcal{A}$ for that TP, which is equal to another Jaccard metric, but this time over the association concepts of TPAs/FPAs/FNAs. 

$\mathcal{A}$ measures the alignment between the gtTraj and prTraj which are matched at the TP $c$. This alignment is calculated using the same formulation (Jaccard) as is used to measure the alignment between the whole set of all gtTrajs and all prTrajs for detection, but in this case only over the subset of trajectories that are matched at a TP.

Note that concepts such as $\mathcal{A}$, TP, FN, FP, TPA, FNA and FPA, are all calculated for a particular value of $\alpha$. However, the $\alpha$ subscript is omitted for clarity. We will continue to omit this $\alpha$ throughout the paper except where it is needed for understanding.

\PAR{Matching to Optimise HOTA.}
Like in MOTA (and IDF1) the matching occurs in HOTA to maximise the final HOTA score. The Hungarian algorithm is run to select the set of matches, such that as a first objective the number of TPs is maximised, as a secondary objective the mean of the association scores $\mathcal{A}$ across the set of TPs is maximised, and as a third objective the mean of the localisation similarity $\mathcal{S}$ across the set of TPs is maximised.
This is implemented with the following scoring for potential matches, MS, between each gtDet $i$ and each prDet $j$.
\begin{align}
\text{MS}(i,j) = 
\begin{cases}
\frac{1}{\epsilon} + \mathcal{A}_\text{max}(i,j) + \epsilon\mathcal{S}(i,j)	& \text{if}\ \mathcal{S}(i,j) \geq \alpha \\
\end{cases}
\label{eq:matching}
\end{align}
where $\epsilon$ is small number such that the 
components have different magnitudes.
$\mathcal{A}_\text{max}$ is the maximum $\mathcal{A}$ score if detections are not required to be bijectively matched, e.g. if each prDet and each gtDet are allowed to match with multiple others. 
$\mathcal{A}_\text{max}$ is optimised as a proxy for $\mathcal{A}$. This is because $\mathcal{A}$ depends upon which matches are selected, and thus cannot be optimised using a linear assignment formulation. This approximation is valid, because $\mathcal{A}$ approaches $\mathcal{A}_\text{max}$ for the optimal assignment. 
\PAR{Integrating over Localisation Thresholds.}
The formulation for HOTA$_{\alpha}$ in Eq.~\ref{equ:hotaalpha} accounts for both detection and association accuracy, but doesn't take into account localisation accuracy. In order for HOTA to measure localisation, the final HOTA score is the integral (area under the curve) of the HOTA score across the valid range of $\alpha$ values between $0$ and $1$. This is approximated by evaluating HOTA at a number of different distinct $\alpha$ values ($0.05$ to $0.95$ in $0.05$ intervals) and averaging over these values. For each $\alpha$ value the matching between gtDets and prDets is performed separately.
\begin{equation}
    \text{HOTA} = \int_{0}^{1}{ \text{HOTA}_\alpha \; d\alpha } \approx \frac{1}{19} \sum_{\alpha \in \{ \substack{0.05, \; 0.1, \; ... \\ 0.9, \; 0.95} \} } \text{HOTA}_\alpha
    \label{eq:integrate}
\end{equation}

\PAR{HOTA in One Sentence.}
If HOTA were to be described in one sentence it would be:

\textit{HOTA measures how well the trajectories of matching detections align, and averages this over all matching detections, while also penalising detections that don't match.}

\section{Decomposing HOTA into Different Error Types.}
\label{sec:decomposition}

\begin{figure}[t!]
\centering
\includegraphics[width=\linewidth]{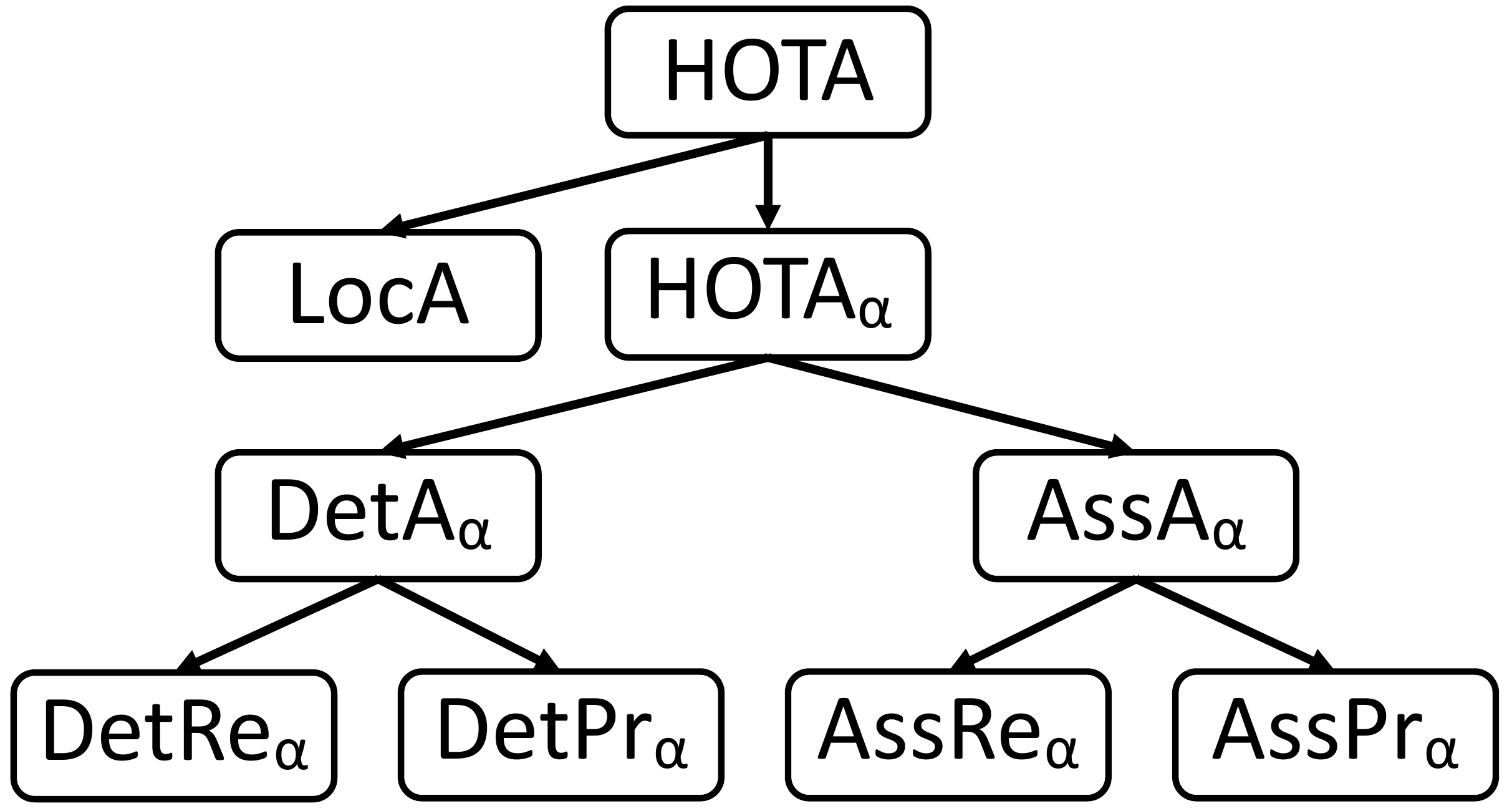}
\caption{Diagrammatic representation of how HOTA can be decomposed into separate sub-metrics which are able to differentiate between different types of tracking errors.}
\label{fig:decomposition}
\end{figure}

A set of evaluation metrics has two main purposes. 
The first purpose is to enable simple comparison between methods to determine which perform better than others. For this purpose it is important that there exists a single metric by which methods can be ranked and compared, for this we propose the HOTA metric (see Sec~\ref{sec:HOTA}). 
The second purpose of evaluation metrics is to enable the analysis and understanding of the different types of errors that algorithms are making, in order to understand how algorithms can be improved, or where they are likely to fail when used. In this section, we show that the HOTA metric naturally decomposes into a family of sub-metrics which are able to separately measure all of the different aspects of tracking. Fig.~\ref{fig:decomposition} shows each of these sub-metrics and the relations between them.

HOTA solves the long-held debate in the tracking community \cite{PETSvsVACE,TrackingChallenges} about whether it is better to have a single evaluation metric or multiple different metrics. HOTA simultaneously gives users the benefits of both options, a single metric for ranking trackers, and the decomposition into sub-metrics for understanding different aspects of tracking behaviour.

\PAR{Taxonomy of Error Types.}
We classify potential tracking errors into three categories: detection errors, association errors and localisation errors. Detection errors can be further categorised into errors of detection recall (measured by FNs) and detection precision (measured by FPs). Association errors can be further categorised into errors of association recall (measured by FNAs) and association precision (measured by FPAs).

Detection recall errors occur when trackers fail to predict detections that are in the ground-truth (misses).
Detection precision errors occur when trackers predict extra detections that don't exist in the ground-truth.
Association recall errors occur when trackers assign two different prIDs to the same gtTraj.
Association precision errors occur when trackers assign the same prID to two different gtTrajs. 
Localisation errors occur when prDets are not perfectly spatially aligned with gtDets.

Our taxonomy aligns with previous work to classify different tracking errors \cite{Monotonicity}, which defines the five basic error types as: false negatives, false positives, fragmentations, mergers and deviations. These are equivalent to detection recall, detection precision, association recall, association precision, and localisation, respectively.

Leichter and Krupka \cite{Monotonicity} argue that any set of tracking metrics must be both error type differentiable and monotonic with respect to these five basic error types. HOTA meets both criteria as it naturally decomposes into separate sub-metrics measuring each basic error type, and is designed to ensure monotonicity (see Sec.~\ref{sec:analysis}). 

\PAR{Metrics for Different Tracking Scenarios with Different Requirements.}
The decomposition of HOTA into different sub-metrics has the further advantage that it enables users to select algorithms or tune algorithms' hyper-parameters based on the nuances of their particular use-case. 

For example, in human motion analysis, crowd analysis or sports analysis \cite{TrackingChallenges,TrackingTrackers} it may be far more important to predict correct, identity preserving trajectories (association recall and precision), than to find all present objects (detection recall).
Whereas for a driving assistance system, it is crucial to detect every pedestrian to avoid collision (detection recall) while also not predicting objects that are not present to avoid unnecessary evasive action (detection precision). However, correctly associating these detections over time may be less crucial (association recall and precision).
In surveillance scenarios, it is typically more important to ensure all objects are found (detection recall), whereas extra detections can easily be ignored by human observers (detection precision).
For short-term future motion prediction it is important to have accurate trajectories of recent object motion, without mixing trajectories of multiple objects (association precision), in order to extrapolate to future motion. Whereas it doesn't matter if trajectories are not correctly merged into long-term consistent tracks (association recall). 

With HOTA, different aspects of tracking can easily be analysed and optimised for, which was not as easily or intuitively possible with previous metrics.

\PAR{Measuring Localisation.}
HOTA is calculated at a number of different localisation thresholds $\alpha$, and the final HOTA score is the average of the HOTA$_{\alpha}$ scores calculated at each threshold. This formulation ensures that the final HOTA score takes the actual localisation accuracy into account. 

A localisation accuracy score (LocA) can be measured separately from other aspects of tracking, as follows:
\begin{align}
  \text{LocA} = \int_{0}^{1}{ \frac{1}{|\text{TP}_\alpha|} \sum_{c \in \{\text{TP}_\alpha\}} \mathcal{S}(c) \; d\alpha }
\end{align}
Where $\mathcal{S}(c)$ is the spatial similarity score between the prDet and gtDet which make up the TP $c$.
This is similar to MOTP \cite{CLEARMOT}, however it is evaluated over multiple localisation thresholds $\alpha$, similarly to how HOTA is calculated.

\PAR{Separating Detection and Association.}
HOTA can be naturally decomposed into a separate detection accuracy score (DetA) and an association accuracy score (AssA) as follows: 
\begin{align}
&\text{DetA}_\alpha = \frac{|\text{TP}|}{|\text{TP}| + |\text{FN}| + |\text{FP}|} &
\label{eq:DetA} \\
&\text{AssA}_\alpha = \frac{1}{|\text{TP}|} \sum_{c \in \{\text{TP}\}} \mathcal{A}(c) &
\label{eq:assa} \\
&\mathcal{A}(c) = \frac{|\text{TPA}(c)|}{|\text{TPA}(c)| + |\text{FNA}(c)| + |\text{FPA}(c)|} &
\\
&\begin{aligned} 
\text{HOTA}_\alpha 
&= \sqrt{\frac{\sum_{c \in \{\text{TP}\}} \mathcal{A}(c)  }{|\text{TP}| + |\text{FN}| + |\text{FP}|}} &\\
&= \sqrt{\text{DetA}_\alpha\cdot\text{AssA}_\alpha}&
\end{aligned}&
\end{align}
We see that HOTA is equal to the geometric mean of a detection score and an association score. This formulation ensures that both detection and association are evenly balanced, unlike many other tracking metrics, and that the final score is somewhere between the two.

It also ensures that both the detection score and association score have the same structure. Both are calculated using a Jaccard index formulation, and both are calculated to ensure that each detection contributes equally to the final score.

The detection score is simply the standard Jaccard index, which is commonly used for evaluating detection. The association score is a sum of Jaccard indices over the TPs, where each is equal to a standard Jaccard index evaluated only between the trajectories which are a part of that TP.

In practice (see Sec.~\ref{sec:evaluation}) this often results in trackers having DetA and AssA scores that are quantitatively similar.

Final DetA and AssA scores can be calculated by integrating over the range of $\alpha$ values in the same way as done for the HOTA score in Eq.~\ref{eq:integrate}.
Note that DetA and AssA should only be combined into HOTA before integrating over a range of $\alpha$ values, not afterwards.

\PAR{Separating Recall and Precision.}
HOTA is further decomposable in that each of the detection and association components can be simply decomposed into a recall and precision component.

\PAR{Detection Recall/Precision.} 
The detection recall/precision are defined as follows:
\begin{align}
    \text{DetRe}_\alpha &= \frac{|\text{TP}|}{|\text{TP}| + |\text{FN}| } \\
    \text{DetPr}_\alpha &= \frac{|\text{TP}|}{|\text{TP}| + |\text{FP}| } \\
    \text{DetA}_\alpha &= \frac{\text{DetRe}_\alpha \cdot \text{DetPr}_\alpha}{\text{DetRe}_\alpha + \text{DetPr}_\alpha - \text{DetRe}_\alpha \cdot \text{DetPr}_\alpha}
\end{align}
These are equivalent to the concepts commonly used in the field of object detection~\cite{pascalvoc}. 
Detection recall is the percentage of ground-truth detections that have been correctly predicted, and detection precision is the percentage of detection predictions made which are correct.
DetRe and DetPr are easily combined into DetA.
Final DetRe and DetPr scores can be calculated by integrating over the range of $\alpha$ values in the same way as done for the HOTA score in Eq.~\ref{eq:integrate}.

\PAR{Association Recall/Precision.}
The association recall/precision are defined as follows:
\begin{align}
    \text{AssRe}_\alpha &= \frac{1}{|\text{TP}|} \; \sum_{c \in \{\text{TP}\}} \frac{|\text{TPA}(c)|}{|\text{TPA}(c)| + |\text{FNA}(c)|}  \\
    \text{AssPr}_\alpha &= \frac{1}{|\text{TP}|} \; \sum_{c \in \{\text{TP}\}} \frac{|\text{TPA}(c)|}{|\text{TPA}(c)| + |\text{FPA}(c)|}\\
    \text{AssA}_\alpha &= \frac{\text{AssRe}_\alpha \cdot \text{AssPr}_\alpha}{\text{AssRe}_\alpha + \text{AssPr}_\alpha - \text{AssRe}_\alpha \cdot \text{AssPr}_\alpha}
\end{align} 
Unlike the detection equivalent, these are novel concepts.
Association recall measures how well predicted trajectories cover ground-truth trajectories. E.g. a low AssRe will result when a tracker splits an object up into multiple predicted tracks.
Association precision measures how well predicted trajectories keep to tracking the same ground-truth trajectories. E.g. a low AssPr will result if a predicted track extends over multiple objects.

AssRe and AssPr are easily combined into AssA.
The introduction of association precision and recall are very powerful tools for measuring different aspects of MOT performance and are a natural extension to the similar widely used detection concepts.
Final AssRe and AssPr scores can be calculated by integrating over the range of $\alpha$ values in the same way as done for the HOTA score in Eq.~\ref{eq:integrate}.

\section{Analysing the Design Space of HOTA}
\label{sec:analysis}
In this section, we analyse a number of different design choices of the HOTA algorithm, and the effect of these choices on the properties of the evaluation metric.

\PAR{Higher-Order vs First-Order Association.}
In HOTA (and in MOTA) the concept of an association is measured for each detection. The association score in HOTA, and the number of IDSWs in MOTA seek to answer the question `how well is this detection associated throughout time?'. In MOTA an IDSW measures this association only one time-step back into the past. E.g. whether this detection has the correct association compared to the previous detection. Since association is measured only over one step, we term this \textit{first-order association}. A metric which considers associations over two time-steps could be called second order association. In contrast, HOTA measures association over all frames of a video. We term this concept \textit{higher-order association} and name our HOTA metric after it.
This property allows HOTA to measure long-term association, which is lacking from the MOTA metric.
\PAR{Higher-Order vs First-Order Matching.}
Just as one of the main conceptual differences between HOTA and MOTA is first-order vs higher-order association between detections, one of the main conceptual differences between HOTA and IDF1 can be thought of as first-order vs higher-order matching between trajectories.

In IDF1, each trajectory is matched only with a single other trajectory and scored by how well it aligns with this single trajectory. We call this \textit{first-order matching}. This is enforced by a unique bijective matching between prTrajs and gtTrajs. HOTA in contrast, is able to measure how well each trajectory matches with all possible matching trajectories, which we term \textit{higher-order matching}. This is done by performing matching at a detection level, which allows each trajectory to match to different trajectories in each time-step, and then scoring the alignment between each of these matching trajectories for each matching detection.

Thus HOTA can be thought of as being \textit{higher-order} in terms of both association and matching.
\PAR{Detection vs Trajectory Matching.}
Both HOTA and MOTA create matches between sets of detections, whereas other metrics like IDF1 and Track-mAP directly match whole trajectories with one another. 
Matching detections has the advantage over matching trajectories that all of the possible trajectory matches can be measured by the metric simultaneously. In IDF1, if a gtTraj is split between multiple prTrajs, only the best matching trajectories are considered correct, while all the remaining trajectories are considered incorrect. This causes the problem that the association accuracy of non-matched trajectories are ignored, no matter how well they are associated it will not affect the score. This also has the disadvantage that the score actually decreases as detection accuracy increases for non-matched segments. This is because such segments are considered negatives and decrease the score.
Matching at a detection level instead of a trajectory level is required in order to ensure that the association accuracy of all segments contributes to the final score, and that the metric monotonically increases as detection improves.

\PAR{Jaccard vs F1 vs MODA.}
In HOTA we use the Jaccard index to measure both detection and association accuracy. We compare this formulation with two possible alternatives, the F1 score and the MODA score.
\begin{align}
    &\text{Jaccard} = \frac{|\text{TP}|}{|\text{TP}| + |\text{FN}| + |\text{FP}|} &\\
    &\text{F1} = \frac{|\text{TP}|}{|\text{TP}| + 0.5|\text{FN}| + 0.5|\text{FP}|} &\\
    &\text{MODA} = \frac{|\text{TP}| - |\text{FP}|}{|\text{TP}| + |\text{FN}|} &
\end{align}
The F1 score (also called the Dice coefficient) is the harmonic mean of recall and precision and is used in IDF1 as well as other metrics such as PQ for panoptic segmentation \cite{PanSeg}.
The MODA score (Multi-Object Detection Accuracy) is the MOTA score without considering IDSWs and is often used for measuring detection accuracy in video. 

Of the three, the Jaccard formulation is the only one that meets all three requirements to mathematically be a \textit{metric}, and the only one that obeys the triangle inequality (MODA also isn't symmetric). Note that when using the Jaccard formulation the entire HOTA metric also obeys the triangle inequality, and is mathematically a \textit{metric}, because the mean of metrics over non-empty finite subsets of a metric space is also a metric \cite{AveragingMetrics}.

\begin{figure}[t!]
\centering
\includegraphics[width=\linewidth]{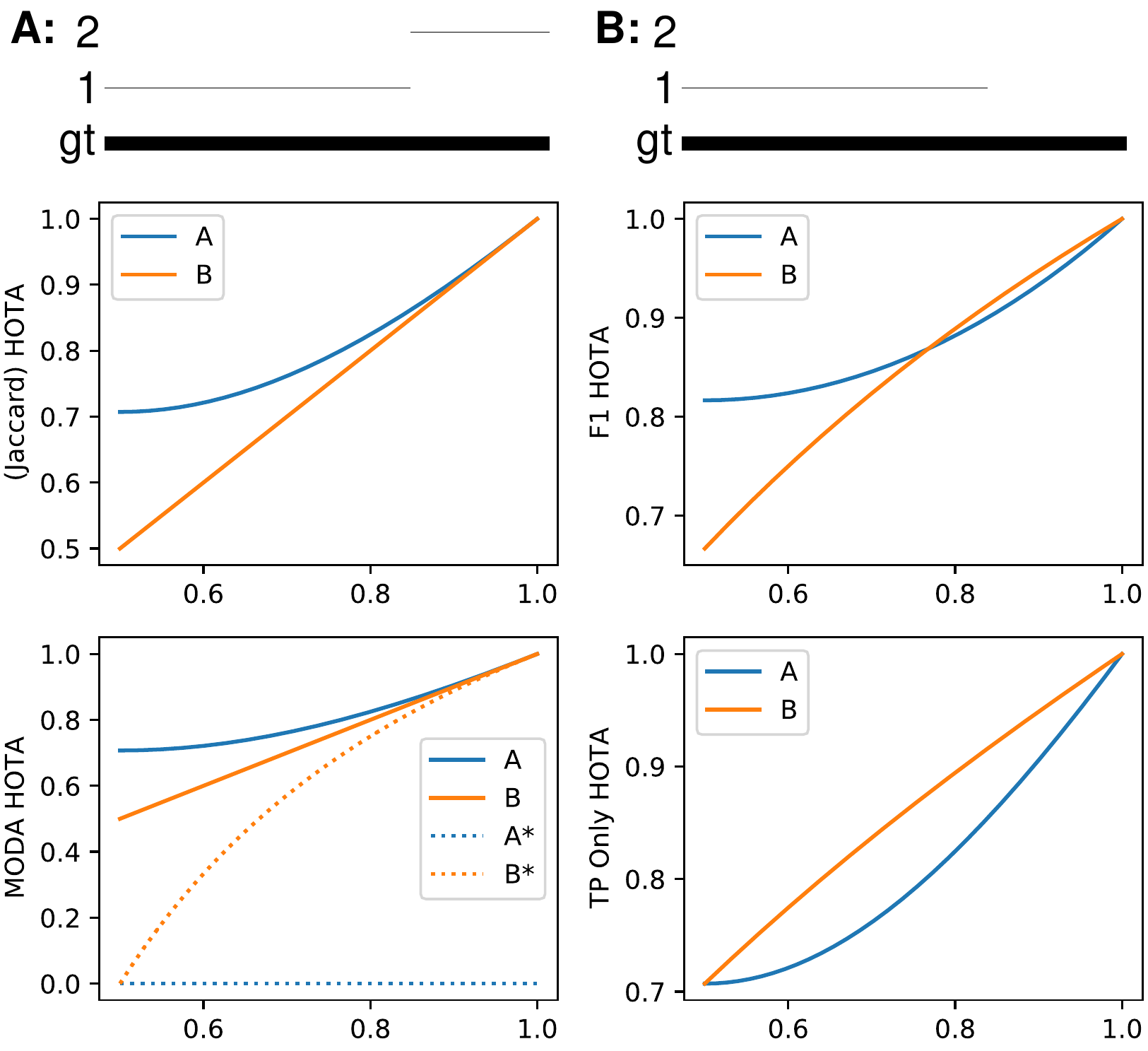}
\caption{A simple tracking example showing why the Jaccard formulation for HOTA is preferable to a number of others such as F1, MODA and a formulation which excludes FNs and FPs from the association score (see text). For tracking a single object present in all frames (bold line), two tracking results, A and B (thin lines) are presented. The x-axis of the plots is the ratio of the len(traj 1)/len(gt).
The tracking result of A should always have a higher score than B for the metric to be monotonic in detection, over all possible ratios of the len(traj 1)/len(gt), and also if the predictions and ground-truth are swapped. This is only valid for Jaccard based HOTA. Note that the MODA formulation is non-symmetric so the results when swapping the ground-truth and tracks are shown as dashed lines and labeled with an asterisk. The other formulations are symmetric.}
\label{fig:iou-f1}
\end{figure}

Fig.~\ref{fig:iou-f1} shows a simple tracking example designed to show the monotonicity of different formulations.
The HOTA score is evaluated using the three different formulations for both detection and association scores. If the evaluation measure is monotone, the tracking results in (A) should always be higher than (B) because (A) contains more correct detections. This should also be true when swapping which set is the ground-truth and which is the prediction. As can be seen of the three formulations, Jaccard is the only one to exhibit this monotone property. The F1 formulation scores B higher when the track labeled (1) is adequately long. The MODA formulation is non-symmetric. It acts the same as Jaccard in the absence of TPs and TPAs, but exhibits very undesirable behaviour when the ground-truth and predictions are switched. 
Both the monotone and symmetric properties result in the Jaccard formulation being preferable to the other two.

\PAR{Including vs Excluding Detection Errors in the Association Score.}
In HOTA, the association score for a TP is the Jaccard index score between the gtTraj and the prTraj that have the same gtID and prID as the TP, respectively. These trajectories could include FNs and FPs which are not matched. This can be seen in the definitions of TPA and TNA in Sec.~\ref{sec:HOTA}.
A potential alternative formulation would only calculate this Jaccard index over TPs such that any FNs or FPs in these trajectories are ignored, and do not count toward the count of FNAs and FPAs. 
This formulation may seem advantageous in that association is now calculated only over TPs, and detection errors would no longer decrease the association score. However, as seen in the fourth panel of Fig.~\ref{fig:iou-f1} this results in non-monotonic results where adding in correct detections decreases the overall score, as the AssA decreases faster than the DetA increases.
By including all detections of matching trajectories in the association score calculation, we ensure that DetA and AssA are perfectly balanced such that an improvement in either one cannot result in a larger decrease in the other, thus ensuring the monotonicity of the metric.

Note that in HOTA, the presence of these FNs and FPs in the association score does not influence the \textit{error type differentiability} of the metric. The AssA still measures only association and the DetA still measures only detection. These terms in the association score correspond to measuring associations to unmatched detections.

\PAR{With vs Without the Square Root.}
The HOTA formulation contains a square root operation after the double Jaccard formulation. This square root has three effects. The first is that it increases the magnitude and spread of trackers' scores. While the magnitude of the scores is in itself not important, it is nice to see that both the magnitude and spread of HOTA scores is in the same ball-park range as previous metrics (see Sec.~\ref{sec:evaluation} Fig.~\ref{trends}). This means that researchers' current intuitive understanding of how good certain scores are, still roughly holds.

The second effect of the square root is its interpretation as the geometric mean of a detection score and an association score. This is natural and intuitive as we wish for HOTA to evenly balance both detection and association, thus having HOTA as the geometric mean of these two scores is a good choice. The geometric mean has the advantage over other formulations such as an arithmetic mean that the score approaches $0$ as either of the two sub-scores approach zero. Thus when a tracker completely fails in either detection or association, a very low score will effectively represent this.

The third effect is that it accounts for double counting of similar error types. As discussed above, the association score also includes FP and FN detection errors. The use of the square root prevents double counting of these errors. This can be illustrated by a simple example. In cases where there is only one gtTraj and one prTraj, such as in single object tracking (SOT), the HOTA score reduces to the following equation:
\begin{align}
\begin{aligned}
    &\text{HOTA}_{\alpha}\{ \text{SOT} \} \\
    &\;\;= \sqrt{\text{DetA}.\text{AssA}} \\
    &\;\;=  \sqrt{\frac{|\text{TP}|}{|\text{TP}| + |\text{FN}| + |\text{FP}|}.
    \frac{1}{|\text{TP}|} \sum_{|\text{TP}|}\frac{|\text{TP}|}{|\text{TP}| + |\text{FN}| + |\text{FP}|}} \\
    &\;\;=  \frac{|\text{TP}|}{|\text{TP}| + |\text{FN}| + |\text{FP}|}
    \label{eq:SOT-HOTA}
\end{aligned}
\end{align}
In the single object tracking case, the association score and the detection score are the same. This makes sense because the association score can be thought of as the average detection score between matched trajectories, and where there is only one possible matching trajectory these scores are the same. The square root thus cancels out these same errors being counted twice, and results in an intuitive metric for SOT, which is just a Jaccard metric measuring the ratio of correct to incorrect predictions. 

\PAR{With vs Without the Detection Accuracy.}
As we have seen above, the AssA takes into account FP and FN detection errors for matched trajectories. A natural question to ask then, is whether we even need the DetA, or if the AssA is adequate by itself.
The DetA is critical for two reasons, the first is in accounting for non-matched trajectories. If we have gtTrajs or prTraj that are not matched at all, then these FNs and FPs will not at all be taken into account in the AssA. The DetA is needed to penalise these errors. 
The second reason is that the AssA by itself is non-monotonic. Since it is the average association over detections, if we add in one extra matching detection (TP), if this TP has a low association score, then the overall AssA will decrease. By including the DetA in the HOTA score, the DetA and AssA are perfectly balanced such that adding any correct match (TP) will always increase the overall score and thus HOTA is monotonic. 

\PAR{Averaging over Detections vs Averaging over Associations.}
In order to calculate the AssA we average the association scores between matching trajectories over matching detections. This is a natural formulation because it ensures that each detection contributes equally to the AssA, and it results in the intuitive understanding that the contribution for each detection is weighted by how accurately that detection is associated across time.

An alternative formulation would be to calculate AssA by averaging over all possible associations rather than averaging over detections. This would result in a quadratic dependence on the number of TPs such that longer matching trajectories would contribute quadratically more than shorter matching trajectories, which is undesirable. It would also lose the interpretation that the contribution of each TP is weighted by its association accuracy, as well as making HOTA non-monotonic.

\PAR{Final Tracks vs Potential Tracks with Confidence Scores.}
Taking into account the confidence of predictions has both advantages and disadvantages for metrics. One of the main disadvantages is that your metric is no longer evaluating an actual final tracking result, but rather a selection of potential tracking results which are ranked by how likely each one is.
Such an approach makes sense when the ground-truth is inherently ambiguous (such as trajectory forecasting) as we can't expect algorithms to predict the correct results and as such it is only fair to allow multiple ranked predictions to be evaluated.
It also makes sense when the task is too difficult for current algorithms to accurately predict the correct ground-truth (such as monocular 3D detection and tracking), and again it is fair to allow algorithms to predict multiple ranked results.

However, as algorithms become better at a task, it is better to evaluate an algorithm's ability to actually predict the correct ground-truth values. This also has other benefits such as enabling constraints between detection representations such as commonly used in segmentation mask tracking \cite{MOTS} where segmentation masks are not allowed to overlap. 

For these reasons, HOTA is designed to operate on final tracking results. We also present an extension to HOTA in Sec.~\ref{sec:extensions} in which we present a confidence ranked version of HOTA, which reduces to the default HOTA when taking a fixed set of detections above a certain confidence threshold.

\PAR{Drawbacks of HOTA.}
HOTA has two main potential drawbacks, which could make using it less than ideal in some situations.
The first is that it may not be ideal for evaluating online tracking. This is because association is measured over the entire video, and the association score for each TP depends on how well it is associated in the future, which is not a desirable feature for online evaluation.

The second potential drawback is that it doesn't take fragmentation of tracking results into account (see Fig.~\ref{fig:frag}). This is by design, as we wish for the metric to measure long-term global tracking. However, for applications where this is important, this could be a drawback of HOTA. 

In Sec.~\ref{sec:extensions} we present both an online version of HOTA and a fragmentation-aware version of HOTA, which address both of these issues.

\section{HOTA Extensions}
\label{sec:extensions}
In this section we provide a number of different extensions to HOTA for use in different tracking scenarios.

\PAR{Variety of MOT scenarios.}
HOTA is designed for a variety of MOT scenarios, from 2D box tracking, segmentation mask tracking, 3D tracking, human pose tracking or point tracking or beyond. HOTA can easily be adapted to any such representation, all the is required is measure of similarity $\mathcal{S}$ between objects in whichever representation is chosen (see Sec.~\ref{sec:terminology}).

\PAR{Multi-Camera HOTA.}
HOTA also extends trivially to evaluating Multi-Camera MOT. Both the ground-truth and predictions could contain trajectories in multiple different cameras with consistent ids across cameras. HOTA could then be applied without any changes. We recommend the use of HOTA for multi-camera tracking over the currently used IDF1 metric due to the many advantages of HOTA and drawbacks of IDF1 (see Sec.~\ref{sec:comparison}).  

\PAR{HOTA for Single Object Tracking.}
As can be seen in Eq.~\ref{eq:SOT-HOTA}, HOTA simplifies trivially in the single object tracking (SOT) case to being a Jaccard index (integrated over $\alpha$ values). Currently in SOT most evaluation procedures don't penalise FPs and only evaluate over frames where a gtDet is present. We believe that the HOTA formulation is also perfectly suited for SOT evaluation, both for effectively penalising FPs, as well as for promoting unification between the MOT and SOT communities.

\PAR{Online HOTA.}
By default, HOTA scores association globally over the whole video sequence. Thus the association score for a particular TP depends on how well it is associated both forward and backward in time.
For online tracking scenarios (such as autonomous vehicles) this is not ideal evaluation behaviour as the results from the tracker are used online in each time-step (for decision making) and should be evaluated in a similar way.

Thus we propose a simple extension of HOTA to the online case which we call Online HOTA (OHOTA).
OHOTA is calculated in the same way as HOTA but only time-steps up to the current time-step are used for calculating the association accuracy for each TP.

This is a natural way to extend HOTA to online scenarios, and has the further benefit that it can also be used to evaluate online tracking where trackers are able to update previous predictions in each new time-step. In this case, for each new time-step the TPs for this time-step will have their association scores calculated with the most up to date predictions from all previous time-steps.

\begin{figure}
\centering
\includegraphics[width=\linewidth]{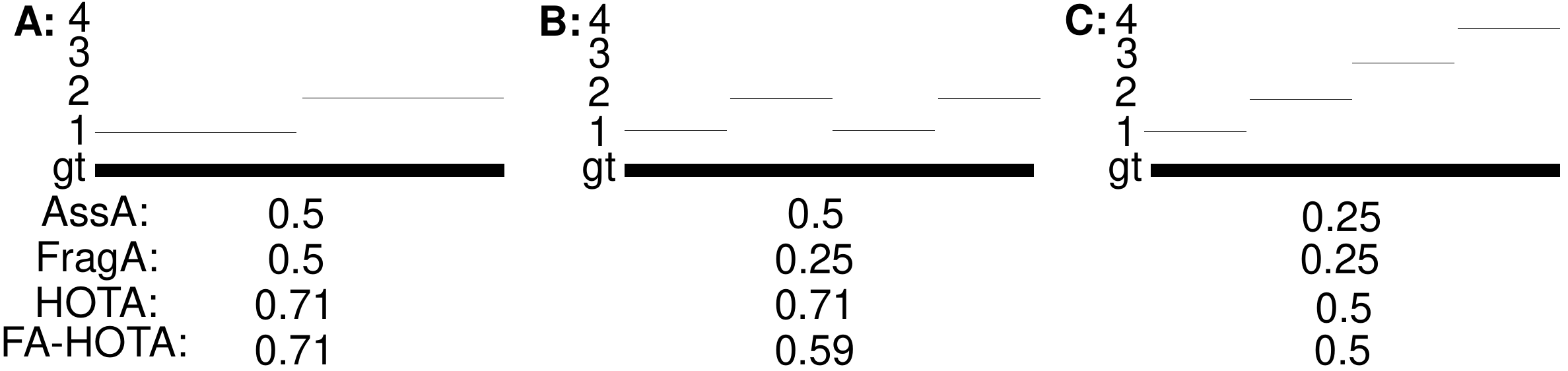}
\caption{
An example showing the difference between fragmentation and association. One gtTraj is present in all frames (bold line), and three tracking results, A, B and C (sets of thin lines) are presented.
A and B have equal association (global alignment), whereas they have different fragmentation (short-range alignment). B and C have equal fragmentation but different association. HOTA only measures association by design. FA-HOTA measures both association and fragmentation.
}
\label{fig:frag}
\end{figure}

\PAR{Fragmentation-Aware HOTA.}
HOTA is designed to evaluate global association alignment between gtTrajs and prTrajs. However, in some cases it is important to measure short-range alignment, which we call fragmentation. Fig.~\ref{fig:frag} clearly shows the difference between association and fragmentation.

In most tracking scenarios the default version of HOTA is preferable, however, for when measuring fragmentations is important, we present an extension which we call fragmentation-aware HOTA (FA-HOTA).
\begin{align}
    &\text{FA-HOTA}_{\alpha} =  \sqrt{\frac{\sum_{c \in \{\text{TP}\}} \Big(\sqrt{ {\mathcal{A}(c) \cdot \mathcal{F}(c)} } \Big) }{|\text{TP}| + |\text{FN}| + |\text{FP}|}} &\\
    &\mathcal{F}(c) = \frac{|\text{FrA}(c)|}{|\text{TPA}(c)| + |\text{FNA}(c)| + |\text{FPA}(c)|}&
\end{align}
where the set of fragment associations of $c$, $\text{FrA}(c)$, is the subset of TPA($c$) which belongs to the same fragment as $c$. A fragment is a set of TPAs for which there are no FNAs or FPAs between them.

We compute the geometric mean of fragmentation and association for each TP in order to concurrently measure both short-term fragmentation alignment and long-term association. 
As the fragmentation score is bounded by the association score, FA-HOTA equals HOTA when no fragmentation occurs.
This formulation also allows us to compute the overall fragmentation accuracy FragA:
\begin{equation}
    \text{FragA} = \frac{1}{|\text{TP}|} \sum_{c \in \{\text{TP}\}} \text{Frag}(c).    
\end{equation}

\PAR{Importance Weighted HOTA.}
As detailed in Sec.~\ref{sec:decomposition} different tracking applications can assign different importance to different aspects of tracking (detection/association recall/precision). We present an extension of HOTA, Weighted HOTA (W-HOTA) which allows users to apply different weights to each aspect depending on their requirements.
\begin{align}
   &\text{W-HOTA}_{\alpha} = \sqrt{\frac{\sum_{c \in \{\text{TP}\}} \mathcal{A}_{\text{w}}(c) }{|\text{TP}| + w_{\text{FN}}|\text{FN}| + w_{\text{FP}}|\text{FP}|}} & \\
    &\mathcal{A}_{\text{w}}(c) = \frac{|\text{TPA}(c)|}{|\text{TPA}(c)| + w_{\text{FNA}}|\text{FNA}(c)| + w_{\text{FPA}}|\text{FPA}(c)|} &
\end{align}
Where each of the weightings, $w_{\text{FN}}$, $w_{\text{FP}}$, $w_{\text{FNA}}$, $w_{\text{FNA}}$ are values between $0$ and $1$. When all weightings are $1$ we have the original HOTA. When any single weight is $0$ that component no longer contributes to the score.

The default weighting provides a balanced weighting between the different components, and should be used for tracking evaluation unless there is a strong reason to use a different weighting for a particular desired outcome.

\PAR{Classification-Aware HOTA.}
Traditionally, there are two ways to deal with evaluation for tracking multiple classes. The first option is to require trackers to assign each object to a class and then evaluate over each class separately before averaging the results over classes. A second option is to ignore the effect of classification all together and simply evaluate all classes together in a class-agnostic way as though they were all the same class.

We propose a third option for dealing with evaluating multiple classes which we call classification-aware HOTA (CA-HOTA).
We require that each prediction is assigned a probability that it belongs to each class such that these probabilities sum to one over all classes. CA-HOTA then becomes:
\begin{align}
    &\text{CA-HOTA}_{\alpha} = \sqrt{\frac{\sum_{c \in \{\text{TP}\}} \Big( {\mathcal{A}(c) \cdot \mathcal{C}(c)} \Big) }{|\text{TP}| + |\text{FN}| + |\text{FP}|}} &
    \label{eq:CA-HOTA}
\end{align}
where  $\mathcal{C}(c)$ is probability that the prDet of $c$ assigned to the class of the gtDet of $c$. 
This effectively weights the contribution of each TP by the classification score, in the same way that it is weighted by an association score.
In this setting we have to include the classification score in the matching procedure so that the matching still maximises the final score. Eq.~\ref{eq:matching} now becomes:
\begin{align}
\begin{aligned}
&\text{MS}(i,j) \\
& \;\;\;\;= 
\begin{cases}
\frac{1}{\epsilon} + \mathcal{A}_\text{max}(i,j) \cdot \mathcal{C}(i,j)	+ \epsilon\mathcal{S}(i,j) & \text{if}\ \mathcal{S}(i,j) \geq \alpha \\
0, & \text{otherwise}
\end{cases}&
\end{aligned}
\end{align}
Each detection, even those belonging to the same prTraj can have different class probabilities.

We can also compute the overall classification accuracy ClaA, in order to evaluate the success of classification separate from other aspects of tracking.
\begin{equation}
    \text{ClaA} =  \frac{1}{|\text{TP}|} \sum_{c \in \{\text{TP}\}} \mathcal{C}(c).
\end{equation}

\PAR{Class-Averaged Classification-Aware HOTA.}
We can also calculate a class-averaged classification-aware HOTA (CA$^2$-HOTA), by calculating a score for each class, Cls, as follows:
\begin{align}
\begin{aligned}
    &\text{CA}^2\text{-HOTA}_{\alpha} \{ \text{Cls} \} &\\
    &\;\;\;\;= \sqrt{\frac{\sum_{c \in \{\text{TP}_{\text{Cls}}\}} \Big( {\mathcal{A}(c) \cdot \mathcal{C}(c,\text{Cls})} \Big) }{|\text{TP}_{\text{Cls}}| + |\text{FN}_{\text{Cls}}| + \sum_{f \in \{\text{FP} \} } \mathcal{C}(f,\text{Cls}) }} &
\end{aligned}
\label{eq:CACA-HOTA}
\end{align}
where TP$_{\text{Cls}}$ and FN$_{\text{Cls}}$ are those which have a ground-truth class Cls, and the notation $\mathcal{C}(c,\text{Cls})$ is the classification score which prediction $c$ has assigned to the ground-truth class Cls. The final score is calculated by averaging over all classes before averaging over $\alpha$ thresholds.

For datasets with many classes \cite{TAO,VIS} we recommend the use of CA$^2$-HOTA as it adjusts for class bias during evaluation. 

\PAR{Federated HOTA.}
TAO \cite{TAO} uses a federated evaluation strategy. Not all objects are annotated in all images. Instead, each image is labeled with the set of classes for which there is confirmed no unannotated objects (for which FPs can be evaluated). Extra predictions of other classes should be ignored as they could be present but unannotated.

We propose a version of HOTA which adapts Eq.~\ref{eq:CACA-HOTA} to federated evaluation (Fed HOTA). 
\begin{align}
\begin{aligned}
    &\text{Fed}\text{-HOTA}_{\alpha} \{ \text{Cls} \} &\\
    &\;\;\;\;= \sqrt{\frac{\sum_{c \in \{\text{TP}_{\text{Cls}}\}} \Big( {\mathcal{A}(c) \cdot \mathcal{C}(c,\text{Cls})} \Big) }{|\text{TP}_{\text{Cls}}| + |\text{FN}_{\text{Cls}}| + \sum_{f \in \{\text{FP}\} } \mathcal{I}(f,\text{Cls}) \cdot \mathcal{C}(f,\text{Cls}) }} &
\end{aligned}
\end{align}
where $\mathcal{I}(f,\text{Cls})$ is $1$ if $f$ is from an image where the class Cls should be counted as false positive, and $0$ otherwise.

\PAR{Confidence-Ranked HOTA.}
HOTA operates on final tracking predictions rather than confidence-ranked potential tracks. However, for certain tracking scenarios such as monocular 3D tracking where it is difficult for trackers to accurately localise detections a confidence-ranked version (CR-HOTA) is more suitable. This follows other metrics such as Track mAP \cite{ImageNet} or sAMOTA \cite{AMOTA}.

When using CR-HOTA trackers must output a confidence score for each detection, $k$. Detections over the whole benchmark are ordered by decreasing confidence. Looping over the ordered detections, the detection recall score is calculated for each one considering all detections with a higher confidence. For $19$ fixed recall values ($0.05$ to $0.95$ in $0.05$ intervals), the HOTA score is calculated using Eq.~\ref{equ:hotaalpha} and Eq.~\ref{eq:integrate}, by taking into account all detections with a confidence score higher than the maximum confidence score needed to obtain a recall at that threshold. The final CR-HOTA score is given by:
\begin{align}
    &\text{CR-HOTA} = \int_{0}^{1}{ \frac{\text{HOTA}_k}{\text{DetRe}_k} \; dk } \approx \frac{1}{19} \sum_{k \in \{ \substack{0.05, \; 0.1, \; ... \\ 0.9, \; 0.95} \} } \frac{\text{HOTA}_k}{\text{DetRe}_k}
\end{align}
By integrating the value of HOTA/DetRe over a range of DetRe scores, we obtain a formulation which reduces to the original HOTA score when only evaluating detections above a given threshold. In this case the HOTA score would be the same for all values recall value from $0$ to DetRe and would be zero afterward.
Note that this formulation is the same as how MOTA is adapted to sAMOTA in  \cite{AMOTA}.

\section{Analysing Previous Evaluation Metrics.}
\label{sec:comparison}
\begin{table}
\setlength{\tabcolsep}{1.4pt}
\renewcommand{\arraystretch}{0.01}
\scriptsize
\centering

\begin{tabular}{ccccccc}
\toprule 
 &  & \textbf{MOTA} & \textbf{IDF1} & \textbf{Track-mAP} & \textbf{HOTA} & \tabularnewline
\midrule 
\makecell{\hspace{0.01pt} \\ \hspace{0.01pt}} & \textbf{Representation} & Final Tracks & Final Tracks & \makecell{Potential Tracks \\ with Conf. Score} & Final Tracks & \makecell{\hspace{0.01pt} \\ \hspace{0.01pt}}\tabularnewline
\makecell{\hspace{0.01pt} \\ \hspace{0.01pt}} & \textbf{Matching Mechanism} & Bijective & Bijective & Highest Conf. & Bijective & \makecell{\hspace{0.01pt} \\ \hspace{0.01pt}}\tabularnewline
\makecell{\hspace{0.01pt} \\ \hspace{0.01pt}} & \textbf{Matching Domain} & Detection & Trajectory & Trajectory & Detection & \makecell{\hspace{0.01pt} \\ \hspace{0.01pt}}\tabularnewline
\makecell{\hspace{0.01pt} \\ \hspace{0.01pt}} & \textbf{Association Domain} & Prev. One Det & Matched Dets & Matched Dets & All Dets & \makecell{\hspace{0.01pt} \\ \hspace{0.01pt}}\tabularnewline
\makecell{\hspace{0.01pt} \\ \hspace{0.01pt}} & \textbf{Scoring Function} & \small{$1-\frac{\sum{\text{Err}}}{|\text{GTDet}|}$} & F1 Score & Av. Precision & Doub. Jaccard & \makecell{\hspace{0.01pt} \\ \hspace{0.01pt}}\tabularnewline
\makecell{\hspace{0.01pt} \\ \hspace{0.01pt}} & \textbf{Bias Toward} & Detection & Association & Association & Balanced & \makecell{\hspace{0.01pt} \\ \hspace{0.01pt}}\tabularnewline
\bottomrule
\end{tabular}

\caption{An overview of different design choices and properties for each of the previously used three metrics and HOTA.}
\label{table:metricsoverview}
\end{table}
In this section, we analyse the previous evaluation metrics MOTA \cite{CLEARMOT}, IDF1 \cite{IDF1} and Track-mAP \cite{ImageNet}, identifying a number of drawbacks for each one and drawing comparisons to our HOTA metric. See Sec.~\ref{sec:previous} for descriptions of each of the previous metrics. Note that our analysis of each metric is with respect to the properties of detection, association and localization. This is not the only framework in which results can be analysed, but one that is common within the tracking community \cite{Monotonicity}.

\PAR{High-level Comparison Between Metrics.}
Table~\ref{table:metricsoverview} shows an overview of the key differences between the four metrics. 

HOTA can be thought of as a middle ground between MOTA and IDF1. MOTA performs both matching and association scoring at a local detection level which biases scores toward measuring detection, while IDF1 performs both at a trajectory level which ends up being biased towards association. HOTA matches at the detection level while scoring association globally over trajectories. This results in HOTA being balanced between measuring detection and association, exhibiting many of the benefits of each method without the drawbacks.

Track-mAP is similar to IDF1 in many ways in that it performs both matching and association at a trajectory level and as such is biased toward measuring association. However, Track-mAP differentiates itself in that it operates on potential tracks with confidence scores rather than final tracks and doesn't perform bijective mapping but matches based on the highest confidence valid matches.

\subsection{Problems with MOTA}
\label{MOTA}
MOTA has been the main MOT evaluation metric since 2006. It has served the community over the years, however we believe its drawbacks have restricted tracking research. Now that we are equipped with better tools and a better understanding of the tracking task, we are able to analyse all of the problems with MOTA, and ensure that new metrics, such as HOTA, don't have the same issues. We hope HOTA will quickly replace MOTA as the default standard for evaluating MOT algorithms.

Below we highlight $9$ separate problems of the MOTA metric, and show how these problems are addressed in HOTA.
 
\begin{MOTAproblem}
Detection performance significantly outweighs association performance.
\end{MOTAproblem}
MOTA measures detection errors as FNs and FPs, and association errors as IDSWs. The ratio of the effect of detection errors to association errors on the final score is given by |FN|+|FP| : |IDSW|.

For real trackers this ratio is extremely high. For the top ten trackers on the MOT17 benchmark \cite{MOT16} on the 1st April 2020 this ratio varies between $42.3$ and $186.4$, with an average of $98.6$.
This is not because trackers are $100$ times better at detection than association, but rather that MOTA is heavily biased towards measuring detection. In fact, on average the effect of detection on the final score is $100$ times as large as the effect of association. 

We also compared the MOTA and MODA (MOTA without IDSWs, which only measures detection) scores for $175$ trackers on the MOT17 benchmark. When fitting a linear regression model between MOTA and MODA the R$^2$ value is $99.4$ indicating that the detection only score MODA explains more than $99$\% of the variation in final MOTA score. In contrast, the R$^2$ value between MOTA and the number of IDSWs is only $23.7$.

This can potentially have significant negative effects on the tracking community. If researchers are tuning their trackers to optimise MOTA to increase scores on benchmarks, then such trackers will be tuned toward performing well for detection while mostly ignoring the requirement of performing successful association.

\begin{figure}
\centering
\includegraphics[width = 0.6\linewidth]{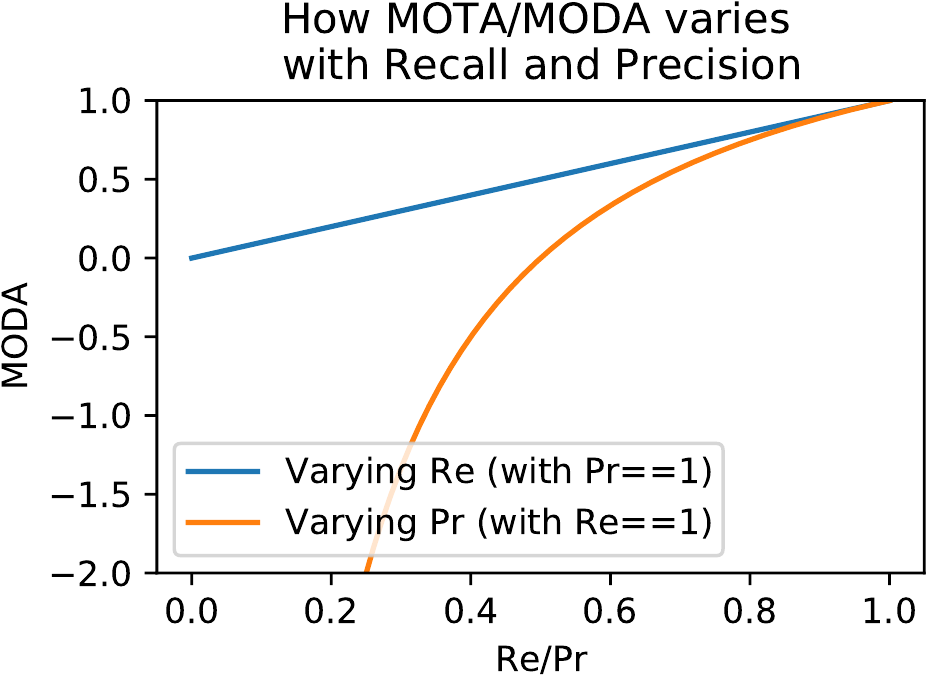}
\caption{Figure showing how MOTA and MODA vary with the Recall and Precision.}
\label{MODA_varying}
\end{figure}

HOTA is designed to not have this problem as it is composed of a detection and association score which both contribute equally. Tab.~\ref{Table:MOT17} and Fig.~\ref{trends} show that the numerical values for each of these scores are similar to one another for real trackers.
\begin{MOTAproblem}
 Detection Precision significantly outweighs the effect of the Detection Recall. 
\end{MOTAproblem}
Let us consider MOTA without considering IDSWs. This is MODA (multi-object detection accuracy). The equation for MOTA/MODA as shown in Eq.~\ref{eq:MOTA} can be rearranged as follows:
\begin{align}
\begin{aligned}
\textrm{MODA} &= 1 - \frac{|\textrm{FN}| + |\textrm{FP}|}{|\textrm{gtDet}|} &\\
& = \frac{|\textrm{TP}| - |\textrm{FP}|}{|\textrm{TP}| + |\textrm{FN}|}& \\
&= \text{DetRe} \cdot (2 - \frac{1}{\text{DetPr}}) &
\end{aligned}
\end{align}
It can be seen that this is not symmetric in terms of recall and precision. The score increases linearly with increasing recall and hyperbolically with increasing precision. This relation can be seen visually in Fig.~\ref{MODA_varying}. Poor precision has a much greater effect on the final score than poor recall.

Again this promotes researchers to tune trackers to optimise precision at the cost of recall, because poor precision values are penalised extremely heavily by MOTA. This can be seen starkly in Fig.~\ref{trends} where precision values for all trackers are much higher than recall values.

As described in Sec.~\ref{sec:decomposition}, for different tracking applications the importance of recall vs precision varies. For applications such as surveillance, recall is often much more important than precision, and as such, the bias imparted by MOTA is particularly harmful. Ideally a benchmark designed for evaluating trackers for a range of applications would evenly weight precision and recall. HOTA solves this issue by using a symmetric Jaccard formulation and thus ensuring that precision and recall are weighted evenly.
\begin{MOTAproblem}
Association errors in MOTA, measured in the form of IDSWs, only take into account short-term (or first order) association.
\end{MOTAproblem}
In MOTA, an IDSW is a TP which has a prID that is different from the prID of the previous TP (that has the same gtID). 
Formally a TP, $c$, is an IDSW under the following criteria:
\begin{align}
\begin{aligned}
c \in \text{IDSW} \hspace{10pt}
\text{if}\ & \hspace{10pt} \textrm{prev}(c) \neq \emptyset & \\
 \text{and}\ & \hspace{10pt} \textrm{prID}(c)  \neq \textrm{prID}(\textrm{prev}(c)) \\
\textrm{prev}(c) = & 
\begin{cases}
\argmax_{k}(\textrm{t}(k)), & \text{if}\ k \neq \emptyset \\
\emptyset, & \text{otherwise}
\end{cases} \\
&k \in \{\textrm{TP} \, | \, \textrm{t}(k) < \textrm{t}(c) \, \land \, \textrm{gtID}(k) = \textrm{gtID}(c) \} 
\end{aligned}
\end{align}

An IDSW only measures if there is an association tracking error to the previous gtDet. This is equal to a single FNA for each TP comparing to the previous TP from the same gtTraj. This can be thought of as a \textit{first-order approximation} to the global association score over the whole trajectory.

This is only able to capture algorithms' ability to perform `short-term tracking', and is unable to evaluate global long-term tracking over a whole video. HOTA in contrast is able to evaluate \textit{higher-order} global tracking by measuring FNAs and FPAs compared to all other detections in matched trajectories.

\begin{MOTAproblem}
MOTA doesn't take into account association precision (ID transfers).
\end{MOTAproblem}
The transpose of an IDSW is called an ID transfer (IDTR).
An IDTR is a TP which has a gtID that is different from the gtID of the previous TP that has the same prID.
Whereas IDSWs compare to the previous gtDet, IDTRs compare to the previous prDet. This is a first-order FPA, whereas an IDSW is a first-order FNA.
ID transfers commonly occur when a predicted track spreads over two ground truth tracks.
MOTA doesn't at all penalise such errors.

As an example consider a short video with only two frames. In scenario A, there is one ground-truth object present in both frames, the tracker correctly detects it in each frame but splits it into two tracks. In this case an IDSW occurs and the MOTA score is $0.5$. In scenario B, there are two ground-truth objects which are present only in one of the frames each, the tracker again detects both correctly but predicts that they are the same object. In scenario B an ID transfer has occurred, but this is not an IDSW and is not penalised and the MOTA score is a perfect $1.0$.

Earlier we saw how MOTA wasn't symmetric between detection precision and recall. Now we see that it also isn't symmetric between association precision and recall. In fact it doesn't measure association precision (ID transfer) errors at all.

This is again potentially extremely undesirable behaviour. Trackers can take advantage of this fact to `hack the metric' to improve their score while performing worse tracking by artificially merging their trajectories over multiple ground-truth objects.

HOTA solves this error by measuring both FPAs and FNAs when measuring association accuracy.

\begin{MOTAproblem}
MOTA does not reward trackers that correct their own association mistakes. In fact, it unfairly penalises such corrections.
\end{MOTAproblem}
\begin{figure}
\centering
\includegraphics[width=\linewidth]{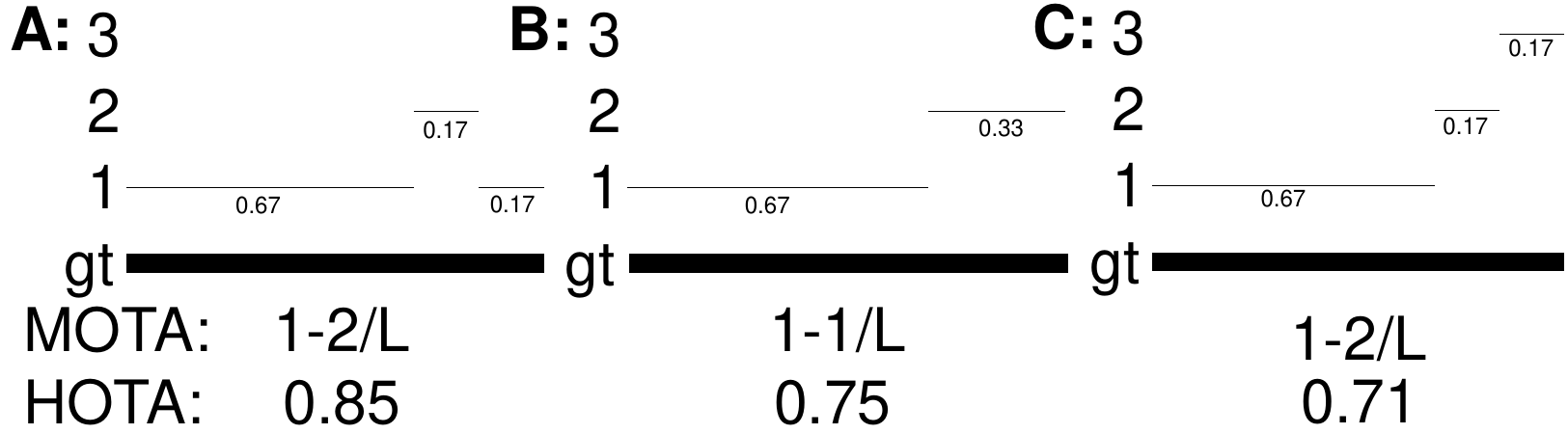}
\caption{An example showing how MOTA does not correctly evaluate tracking situations when a tracker corrects itself after making a mistake. The thick line is the gtTraj. Thin lines are prTrajs. All detections are TPs.}
\label{fig:track-example}
\end{figure}
MOTA is not able to successfully evaluate tracking when a tracker corrects itself after making an association mistake. In this case, MOTA will penalise the tracker twice, first for making a mistake, and then for correcting it. An example of this can be seen in Fig.~\ref{fig:track-example}.
In this example the ordering of the scores should be (A)$>$(B)$>$(C). This is because (A) corrects itself when making a tracking mistake. (B) does not correct itself and continues to track the object with the wrong ID. (C) is even worse making a further tracking mistake. However, the MOTA score does not follow this intuitive ranking. MOTA is unable to account for the tracker correcting itself and counts the correction as a further mistake. This means that under MOTA (A) and (C) are equal even though (A) is clearly much better than (C). 
This property of MOTA heavily disincentivises research into long-term trackers that are able to correct from mistakes.

HOTA solves this issue by measuring the association globally over the whole sequence. In the example HOTA ranks the trackers with the intuitive ranking of (A)$>$(B)$>$(C).

\begin{MOTAproblem}
MOTA does not reward trackers for having a greater alignment between predicted and ground-truth trajectories.
\end{MOTAproblem}
\begin{figure}
\centering
\includegraphics[width=\linewidth]{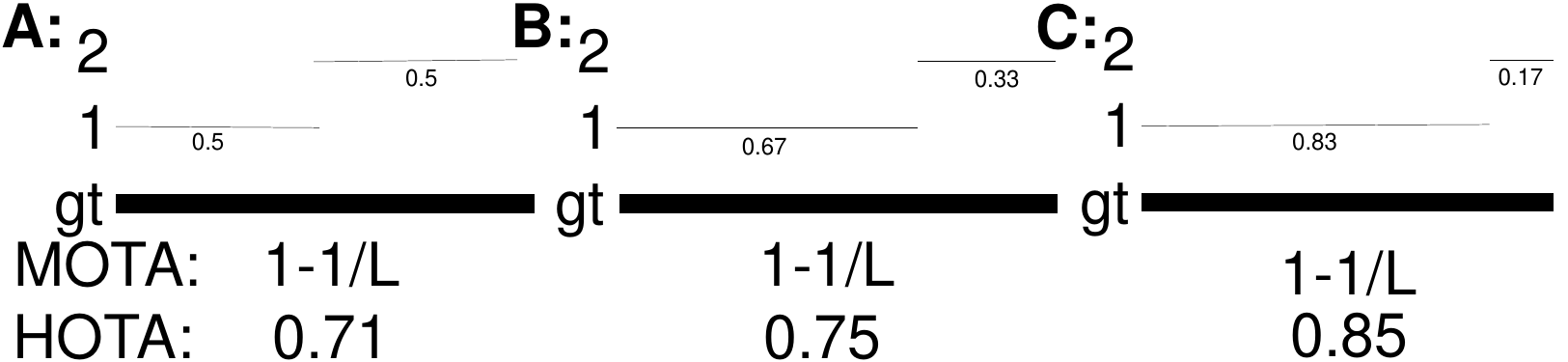}
\caption{An example showing how MOTA does not reward trackers for having a greater alignment between predicted and ground-truth trajectories. The thick line is the gtTraj. Thin lines are prTrajs. All detections are TPs.}
\label{fig:track-example2}
\end{figure}
Another problem that arises because MOTA only evaluates first-order short-term tracking, is that it does not reward trackers based on how well predicted trajectories and ground-truth trajectories align. 
This can be seen clearly in Figure~\ref{fig:track-example2}, where it is clear that (C)$>$(B)$>$(A) due to the fact that in (C) one of the prTrajs is $83$\% similar to the gtTraj, whereas in (A) at best one of the trajectories is only $50$\% similar to the gtTraj. However, because MOTA only evaluates short-term association, it is unable to differentiate between these cases and correctly measure the trajectory alignment.

This property is important, because as prTrajs and gtTrajs become more similar to one another the evaluation score should increase.

HOTA solves this problem by measuring association globally across the whole sequence. Thus HOTA correctly ranks these trackers, and is able to take into account the level of alignment between trajectories.

\begin{MOTAproblem}
In MOTA, the influence of association (IDSWs) on the final score is highly dependent on the video frame rate.
\end{MOTAproblem}

\begin{figure}
\centering
\includegraphics[width=0.9\linewidth]{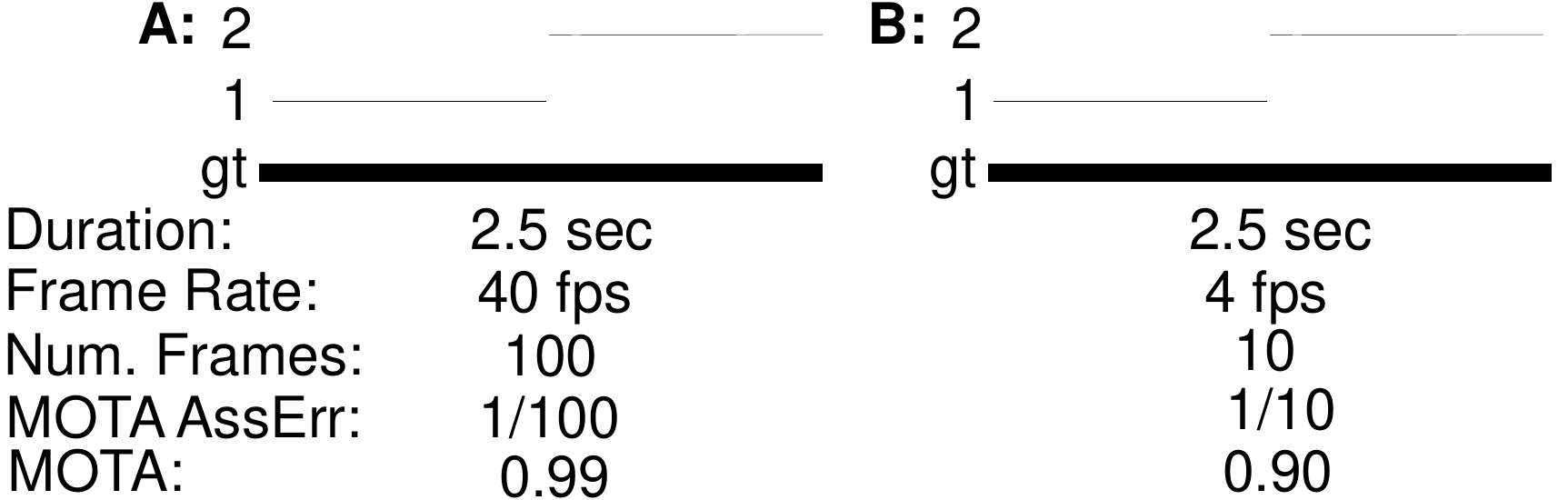}
\caption{An example on MOTAs variability with the frame rate. The thick line is the GT trajectory. Thin lines are predicted trajectories. All detections are TPs.}
\label{frame-rate-bad}
\end{figure}

This is easily seen with an example, as shown in Figure~\ref{frame-rate-bad}, where we show exactly the same sequence processed by the same tracker. In (A), however, it is evaluated at the original video frame rate of 40 fps, in (B) it is evaluated at a reduced frame rate of 4 fps. 
Different scenarios call for different frame rates, e.g., in a surveillance scenario the frame rate of the cameras tends to be low.
In Figure~\ref{frame-rate-bad} we show how the tracker is performing identically in both scenarios.
However, as can be seen, the MOTA score reflects a poorer performance of the tracker when used at a low frame rate, 0.90 MOTA vs 0.99. 
This occurs because MOTA registers only a single error in both cases (one IDSW), and yet this is normalised by the total number of GT detections over the video, which increases with the number of frames. 

Our HOTA does not have this problem, because it measures the association error for each TP and averages this over the TPs. Thus HOTA is independent of the frame rate, and the HOTA value will be the same ($0.707$) for both (A) and (B) in Figure~\ref{frame-rate-bad}.

\begin{MOTAproblem}
MOTA does not take into account localisation accuracy.
\end{MOTAproblem}
MOTA is calculated at a pre-set value of $\alpha$ for determining the matching between prDets and gtDets, but the value of MOTA is the same regardless of how correct these matches are as long as they are over a minimum localisation threshold. 
Thus, MOTA was proposed as one of two metrics that should be used in combination for measuring MOT accuracy. The second of these is MOTP (Multi-Object Tracking Precision), which is simply the average localisation similarity over the TP matches (ignoring detection and association errors). 
Since MOTA was designed to be used together with MOTP, it is not able to measure localisation together in a single metric with detection and association.

HOTA solves this issue by including the localisation accuracy into its calculation. By calculating the score over a range of $\alpha$ values, HOTA is able to include localisation along with detection and association together into one score.

\begin{MOTAproblem}
MOTA scores can be negative and are unbounded.
\end{MOTAproblem}
The final, and perhaps most frustrating, problem of MOTA is that scores are not between $0$ and $1$, as is typically expected for evaluation metrics. 
Although the maximum MOTA score is $1$ for perfect tracking, there is no lower limit to the MOTA score, and it can go down to negative infinite.
This is caused by the score decreasing linearly with the number of FPs, which can be continuously added forever.  
This leads to a score that is hard to interpret: how can we understand a negative MOTA score? Furthermore, a MOTA close to zero might not be as bad as it seems.
HOTA does not have this problem by conveniently being a score between $0$ and $1$. 

\subsection{Problems with IDF1}
\label{IDF1}
The IDF1 metric \cite{IDF1} was originally designed for evaluating tracking in a multi-camera setting, but is trivial to apply to a standard single camera setting. An overview of how IDF1 is calculated can be found in Sec.~\ref{sec:previous}.

IDF1 was designed to measure the concept of `identification'. This concept is related to, but distinct from the concepts of detection and association which we analyse in this paper. Identification is more about determining which trajectories are present, rather than detecting objects and associating them throughout time.

In recent years IDF1 has been adopted by a number of MOT papers \cite{sas_mot,NonMarkovianGloballyConsistantObjectTracking,etc} for MOT evaluation instead of MOTA, as these papers wish to adequately account for association in evaluation, which is lacking from MOTA.
However, IDF1 produces counter-intuitive and non-monotonic results for measuring detection. Due to this, no single-camera MOT benchmarks have adopted IDF1 as the main metric for evaluating trackers and new benchmarks are still choosing to use MOTA instead of IDF1 despite all of MOTAs drawbacks \cite{MOTS,Waymo,BDD,PANDA}. This is because detection is such an important part of tracking evaluation, and IDF1 isn't able to adequately measure it.

In IDF1, each gtTraj can be matched with a single prTraj and vice versa. This contrasts to HOTA where gtTrajs and prTrajs are evaluated as being matched if they are matched at any point in time at any detections.

Just like MOTA can be thought of as only measuring \textit{first-order association} (a single previous association for each detection) compared to HOTA measuring \textit{higher-order association} (all possible associations for each detection), IDF1 can be thought of as measuring \textit{first-order matching} (a single possible match for each trajectory), compared to HOTA measuring \textit{higher-order matching} (all possible matches for each trajectory).

Because IDF1 only allows a single best set of matching trajectories to be evaluated, any trajectory that doesn't end up in this matching set is counted as a negative and decreases the score, even if it contributes correct detections and associations.

Below we highlight $5$ separate problems of the IDF1 metric, and show how these problems are addressed in HOTA.

\begin{IDFproblem}
The best unique bijective mapping of whole trajectories is often not representative of the actual alignment between ground-truth and predicted trajectories.
\end{IDFproblem}

\begin{figure}[t!]
\centering
\includegraphics[width=\linewidth]{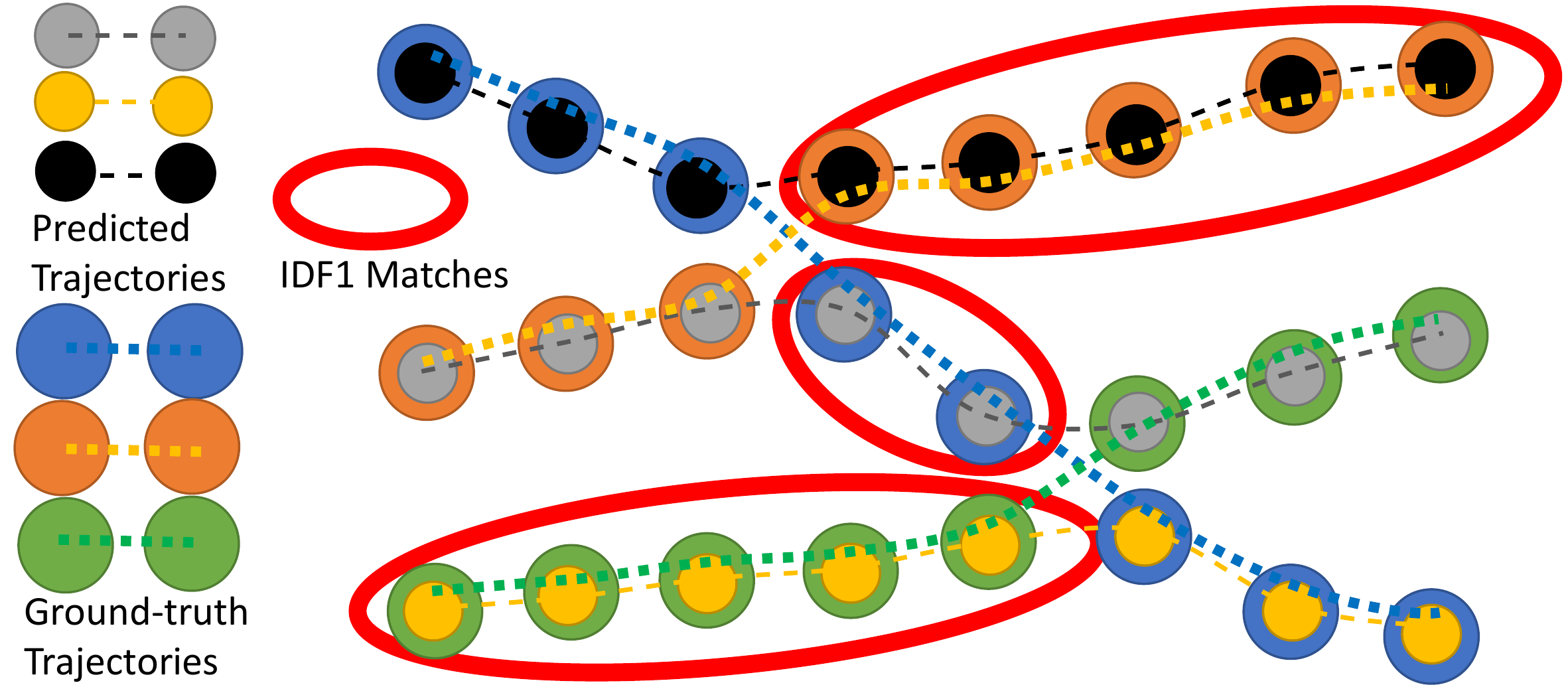}
\caption{A tracking example which shows how the single best trajectory matching, as performed by IDF1, can result in unintuitive matches between trajectories.}
\label{fig:idf1-example}
\end{figure}

This can be best understood with an example, as shown in Fig~\ref{fig:idf1-example}. 
In this example the best bijective mapping matches the blue gtTraj with the grey prTraj. This occurs even though this gtTraj is better matched with both other prTraj, and this prTraj is better matched with both other gtTraj. Thus in this case, matching these trajectories is the worst possible matching for both the gtTraj and the prTraj. However they are still matched here, as when considering all trajectories, all of the better matching options are better matched elsewhere and thus this worst fitting match is what ends up being evaluated.
This is a perfect example to show why a unique bijective mapping of trajectories does not make intuitive sense for evaluating tracking, which often results in complex overlaps between ground-truth and predicted trajectories.

HOTA avoids this problem by not forcing a single global matching between trajectories but rather evaluating over all combinations of ground-truth and predicted trajectories that overlap at any point.

\begin{IDFproblem}
IDF1 actually decreases with improving detection.
\end{IDFproblem}

\begin{figure}[t!]
\centering
\includegraphics[width=\linewidth]{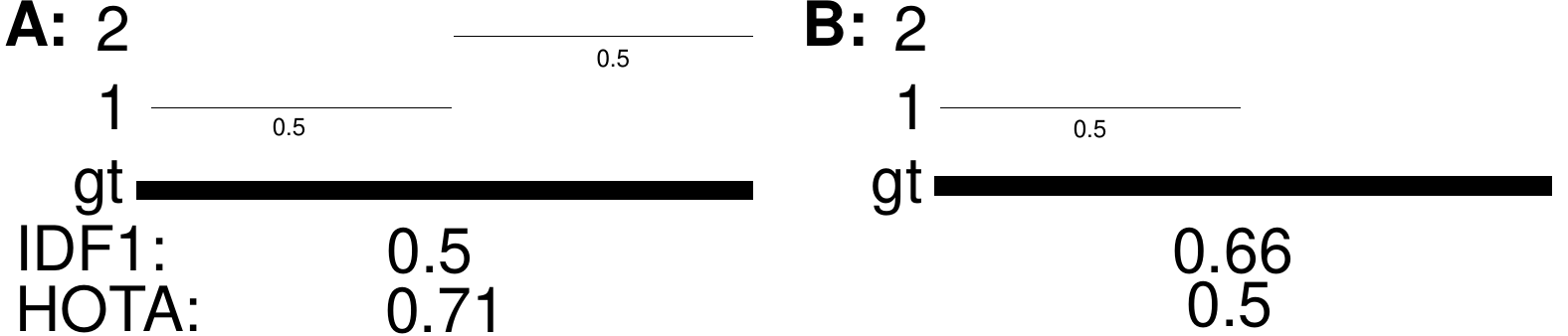}
\caption{A simple tracking example showing how IDF1 fails to correctly rank tracking performance because the score can decrease with improving detection. The thick line is the gtTraj. Thin lines are prTrajs. All detections are TPs, except the second half of the gt in B which are FNs.}
\label{fig:IDF-det}
\end{figure}
In Fig.~\ref{fig:idf1-example} it can be seen that in non-matched regions there are many correct detection results. These correct detections don't add positively to the final score. In fact, each one of these correct detections decreases the final IDF1 score. Thus IDF1 is non-monotonic in detection.

This can be seen more clearly in Fig.~\ref{fig:IDF-det}, where the score for (A) should be higher than for (B), but IDF1 ranks the two tracking results in the other order because it penalises the correct detections in the second trajectory.

This property is extremely counter productive for many tracking scenarios such as in autonomous driving where it is critical to correctly detect objects. 

HOTA does not have this problem and is strictly monotonic in detection such that improving detection always improves the HOTA score.

\begin{IDFproblem}
IDF1 ignores the effect of how good association is outside of matched sections.
\end{IDFproblem}

\begin{figure}[t!]
\centering
\includegraphics[width=\linewidth]{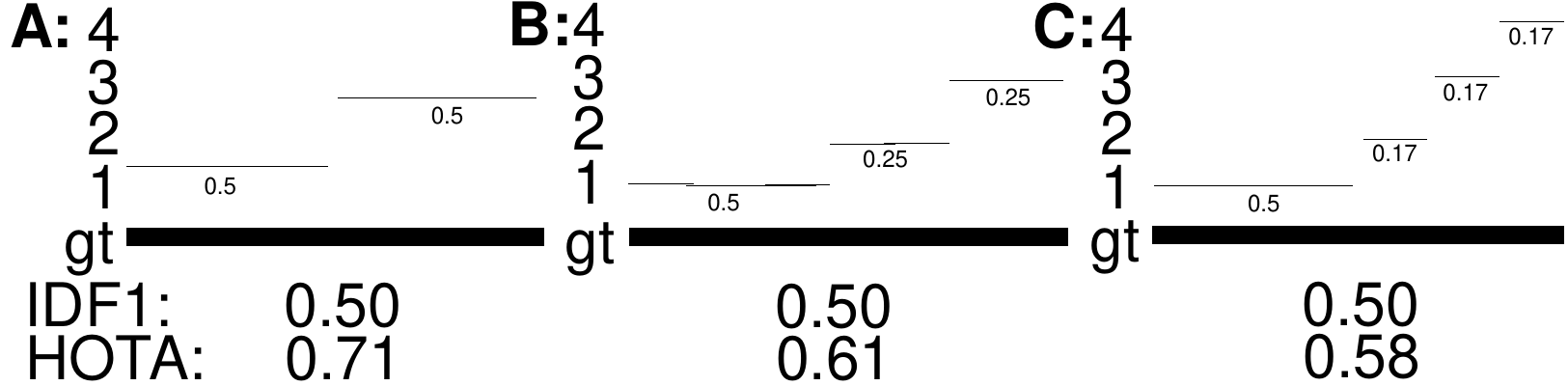}
\caption{A simple tracking example showing how IDF1 fails to correctly rank tracking performance because it ignores the effect of any association that is not included in the best matching trajectories. The thick line is the gtTraj. Thin lines are prTrajs. All detections are TPs.}
\label{IDF-ass}
\end{figure}

Due to the fact that only the matched sections count towards the score, the association can be trivially bad in non matched regions and this will have no effect on the IDF1 score. Thus creating better or worse association does not necessarily correlate to an increase or decrease in the IDF1 score. This can be seen most clearly in Fig.~\ref{IDF-ass} where the IDF1 scores for (A), (B) and (C) are identical where it is clear that the trackers should be evaluated such that (A) $>$ (B) $>$ (C). 

HOTA does not have the same problem as all detections are evaluated, not just the best matching ones. Thus HOTA ranks these trackers correctly.

\begin{IDFproblem}
Scoring highly on IDF1 is more about estimating the total number of unique objects in a scene than it is about good detection or association.
\end{IDFproblem}

Due to the fact that any extra trajectories that are not matched as one of the best matching trajectories are automatically counted as negatives, one of the key design goals for trackers that are optimizing for IDF1 becomes to estimate the total number of unique objects in the scene and only produce that many tracks, rather than performing good detection or good association. This is because if there are more (or less) predicted trajectories than ground-truth trajectories, the extra (missing) trajectories are automatically counted as negatives and severely decrease the score. This is a very different objective than the objective defined in this paper for multi-object tracking, which is to detect the presence of objects and to associate these consistently over time.

In contrast, optimizing for HOTA directly optimises for both accurate detection and accurate association as the final HOTA score is a combination of scores for each of these components.

\begin{IDFproblem}
IDF1 does not evaluate the localisation accuracy of trackers.
\end{IDFproblem}

Like MOTA, IDF1 is also evaluated at a fixed $\alpha$ threshold for how accurate localisation needs to be for detections to match, however the actual localisation of the detections is ignored as long as they are beyond the threshold.

HOTA is evaluated over a range of localisation thresholds $\alpha$ and as such HOTA increases as the localisation of trackers increases.

\subsection{Problems with Track-mAP}
\label{Track-mAP}

Track-mAP is an extension of the mAP metric, commonly used for evaluating detection \cite{pascalvoc,ImageNet,COCO}, to the video domain. An overview of Track-mAP can be found in Sec.~\ref{sec:previous}.
Track-mAP doesn't operate on final tracking results but on confidence-ranked potential tracking results. This results in it being non-trivial to compare to other metrics as it operates on a different, metric specific tracking output format.
Track-mAP is similar to IDF1 in that it also performs matching at a trajectory level. This results in it also being non-monotonic in detection.
Below we highlight $5$ problems of Track-mAP, and show how these are addressed in HOTA.

\begin{mAPproblem}
    The interpretation of tracking outputs is not trivial, nor easily visualisable. 
\end{mAPproblem}
With other metrics, when one wishes to understand a tracker, all they have to do is visualise the tracking results. One can easily identify each of the error types defined for each of the other metrics. 
This is not the case for Track-mAP. Here the output is likely many overlapping outputs, many of which have low confidence scores, with the actual influence of each trajectory on the final score being hidden behind the implicit confidence ranking.
This makes developing trackers that optimise Track-mAP a potentially frustrating experience. It also makes user-comparison, like the type we perform in Sec.~\ref{sec:humanstudy} impossible, as the representation doesn't allow meaningful analysis of the visual results.

\begin{mAPproblem}
It is possible to \textit{game} the metric by tuning trackers for quirks of the metric, which do not necessarily correspond to better tracking.
\end{mAPproblem}

There are a number of ways in which a trackers output can be tuned to increase Track-mAP without actually improving the tracking result.
One of these, as discussed in \cite{WinningVIS}, is that by producing many different predictions with low confidence scores, it greatly increases the chances of obtaining a correct trajectory match and thus improving the score. In \cite{WinningVIS} they do this by replicating trajectories for each class for all other classes.
Since this is possible to tune algorithms in this way, it becomes a requirement for methods to adequately compete. Trackers that don't do this will be heavily penalised.
Some benchmarks \cite{TAO} have attempted to mitigate this issue by restricting results to a maximum number of trajectories per video. However \cite{TAO} still allows $300$ trajectories per frame which still enables significant gaming.
HOTA and other metrics don't have this issue because trackers are required to produce final tracking results.

\begin{figure*}
\centering
\includegraphics[width=0.85\linewidth]{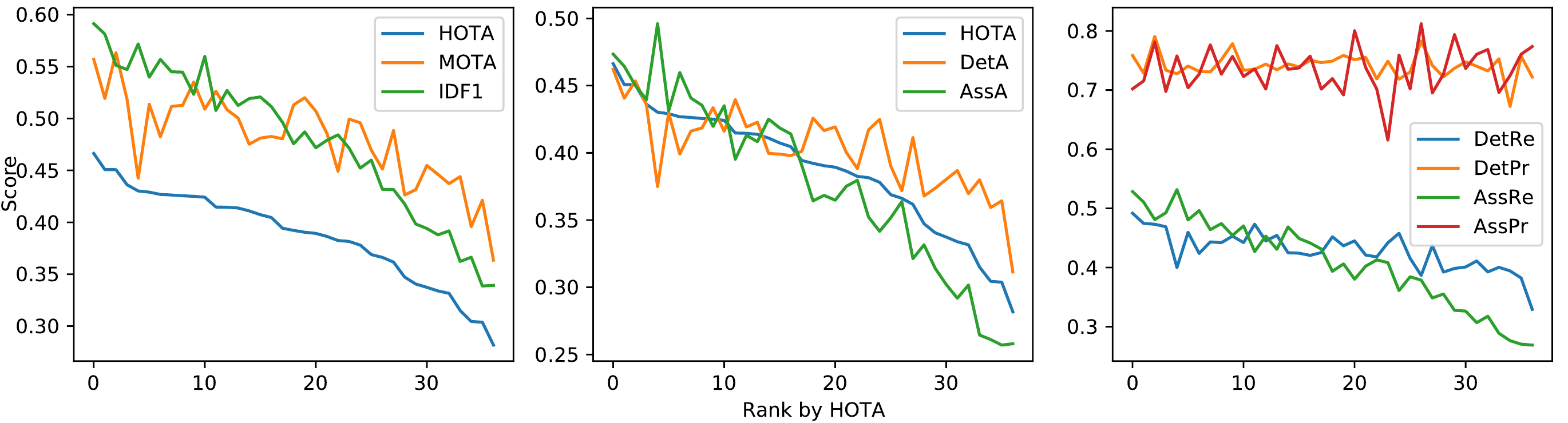}
\caption{Trends between various metrics and sub-metrics on \textbf{MOT17}. All $37$ trackers from Table~\ref{Table:MOT17} are shown in order of decreasing HOTA.}
\label{trends}
\end{figure*}

\begin{mAPproblem}
The threshold for being counted as a positive match is so high that a lot of improvement in detection, association and localisation is ignored by the metric.
\end{mAPproblem}
In the Track-mAP version used in \cite{ImageNet,VisDrone} for a trajectory to be counted as a positive it must be successfully detected and associated such that the Jaccard over detections is at least 50\%. This is a very high threshold for many tracking scenarios. We can see in Fig.~\ref{trends} that for all published trackers on the MOTChallenge leader-board the average Jaccard association alignment between trajectories (AssA) ranges between $0.25$ and $0.5$. This means that even for the very best tracker, more than half of its best guess predictions will be counted as errors in Track-mAP. 
This has the effect that as trackers improve significantly in terms of detection and association this is not shown by an improvement in metric scores. E.g. trajectories that align $5$\% or $45$\% are given the same score.
For the Track-mAP version used in \cite{TAO,VIS} this threshold is even harder to reach because although the threshold is still $50$\%, the score that must be above this threshold is effectively a multiplication of both the trajectory alignment and the average localisation score across the trajectory which is usually much lower (see Sec.~\ref{sec:previous}) 

HOTA doesn't have this issue as it measures the alignment between all trajectory pairs, not just those over a certain threshold.

\begin{mAPproblem}
Improving detection (adding correctly matching trajectories) can decrease the score.
\end{mAPproblem}
Track-mAP is non-monotonic in detection. This is because it matches at a trajectory level (like IDF1) such that extra trajectories are counted as negatives, even if they contain correct, previously unaccounted for detections. 
As described previously this is a non-intuitive and undesirable property for many applications such as autonomous driving where detection is critical.
HOTA on the other hand is monotonic in detection.

\begin{mAPproblem}
Track-mAP mixes association and detection (and localisation) in a way that is not error type differentiable.
\end{mAPproblem}

In the Track-mAP version used in \cite{ImageNet,VisDrone} the score used to measure whether trajectories match is a combination of both detection scores and association scores in a way that is not separable or interpretable.
Thus trajectories can match and add positively to the score if they have high detection accuracy and medium association accuracy, or medium detection accuracy and high association accuracy. The Track-mAP metric doesn't give any indication as to which of these situations is occurring and as such has very limited use for understanding and optimizing the behaviour of trackers.
The Track-mAP version used in \cite{TAO,VIS} is even worse in that the matching score also mixes localisation accuracy with detection and association. Furthermore the effect of localisation accuracy of the score can vary based on differences in object scale over time, which is even more unintuitive.
Note that the detection version of mAP \cite{pascalvoc,ImageNet,COCO} which the tracking version is based upon doesn't have this problem because the matching score only measures a single type of error, which is detection errors. The problem arises when this metric is extended to video to measure multiple error types simultaneously.

HOTA doesn't have this problem because it is designed to be decomposable into separate scores for each error type, such that the effect of each error type on the final metric is clear and the overall metric is error type differentiable.

\begin{table*}
\centering{}
\scriptsize
\renewcommand{\arraystretch}{0.4}

\begin{tabular}{ccccccccccccccc}
\toprule 
 & \multicolumn{3}{c}{\textbf{Ranking}} &  & \multicolumn{3}{c}{\textbf{Scores}} &  & \multicolumn{6}{c}{\textbf{Sub-Scores}}\tabularnewline
\cmidrule{2-4} \cmidrule{3-4} \cmidrule{4-4} \cmidrule{6-8} \cmidrule{7-8} \cmidrule{8-8} \cmidrule{10-15} \cmidrule{11-15} \cmidrule{12-15} \cmidrule{13-15} \cmidrule{14-15} \cmidrule{15-15} 
 & \textbf{HOTA} & \textbf{MOTA} & \textbf{IDF1} &  & \textbf{HOTA} & \textbf{MOTA} & \textbf{IDF1} &  & \textbf{DetA} & \textbf{AssA} & \textbf{DetRe} & \textbf{DetPr} & \textbf{AssRe} & \textbf{AssPr}\tabularnewline
\midrule
\textbf{MPNTrack17~\cite{mpntrack}} & 1 & 2 (\textcolor{green}{$\uparrow$} 1) & 1 (-) &  & 46.6 & 55.7 & 59.1 &  & 46.2 & 47.3 & 49.2 & 75.9 & 52.8 & 70.2\tabularnewline
\textbf{eTC17~\cite{etc}} & 2 & 6 (\textcolor{green}{$\uparrow$} 4) & 2 (-) &  & 45.1 & 51.9 & 58.1 &  & 44.1 & 46.4 & 47.5 & 72.8 & 51.1 & 71.6\tabularnewline
\textbf{Tracktor++v2~\cite{tracktor}} & 3 & 1 (\textcolor{red}{$\downarrow$} 2) & 6 (\textcolor{green}{$\uparrow$} 3) &  & 45.1 & 56.3 & 55.1 &  & 45.3 & 45 & 47.3 & 79.1 & 48.1 & 78.2\tabularnewline
\textbf{eHAF17~\cite{ehaf}} & 4 & 7 (\textcolor{green}{$\uparrow$} 3) & 7 (\textcolor{green}{$\uparrow$} 3) &  & 43.6 & 51.8 & 54.7 &  & 43.6 & 44 & 46.9 & 73.4 & 49.3 & 69.8\tabularnewline
\textbf{SAS\_MOT17~\cite{sas_mot}} & 5 & 31 (\textcolor{green}{$\uparrow$} 26) & 3 (\textcolor{red}{$\downarrow$} 2) &  & 43 & 44.2 & 57.2 &  & 37.5 & 49.6 & 40 & 72.8 & 53.2 & 75.8\tabularnewline
\textbf{YOONKJ17~\cite{yoonkj}} & 6 & 8 (\textcolor{green}{$\uparrow$} 2) & 10 (\textcolor{green}{$\uparrow$} 4) &  & 42.9 & 51.4 & 54 &  & 43 & 43.1 & 46 & 74.1 & 48 & 70.4\tabularnewline
\textbf{DMAN~\cite{dman}} & 7 & 21 (\textcolor{green}{$\uparrow$} 14) & 5 (\textcolor{red}{$\downarrow$} 2) &  & 42.7 & 48.2 & 55.7 &  & 39.9 & 46 & 42.4 & 73.2 & 49.6 & 72.7\tabularnewline
\textbf{jCC~\cite{jcc}} & 8 & 11 (\textcolor{green}{$\uparrow$} 3) & 8 (-) &  & 42.6 & 51.2 & 54.5 &  & 41.6 & 44.1 & 44.3 & 73.1 & 46.4 & 77.6\tabularnewline
\textbf{NOTA~\cite{nota}} & 9 & 10 (\textcolor{green}{$\uparrow$} 1) & 9 (-) &  & 42.6 & 42.6 & 54.5 &  & 41.9 & 43.5 & 44.2 & 75.1 & 47.4 & 72.7\tabularnewline
\textbf{Tracktor++~\cite{tracktor}} & 10 & 3 (\textcolor{red}{$\downarrow$} 7) & 12 (\textcolor{green}{$\uparrow$} 2) &  & 42.5 & 53.5 & 52.3 &  & 43.4 & 42 & 45.3 & 77.8 & 45.4 & 75.7\tabularnewline
\textbf{STRN\_MOT17~\cite{strn}} & 11 & 12 (\textcolor{green}{$\uparrow$} 1) & 4 (\textcolor{red}{$\downarrow$} 7) &  & 42.4 & 50.9 & 56 &  & 41.6 & 43.5 & 44.2 & 73.3 & 47 & 72.3\tabularnewline
\textbf{JBNOT~\cite{jbnot}} & 12 & 4 (\textcolor{red}{$\downarrow$} 8) & 17 (\textcolor{green}{$\uparrow$} 5) &  & 41.5 & 52.6 & 50.8 &  & 44 & 39.5 & 47.3 & 73.5 & 42.7 & 73.6\tabularnewline
\textbf{MOTDT17~\cite{motdt}} & 13 & 13 (-) & 11 (\textcolor{red}{$\downarrow$} 2) &  & 41.5 & 50.9 & 52.7 &  & 41.9 & 41.3 & 44.5 & 74.4 & 45.3 & 70.2\tabularnewline
\textbf{EDMT17~\cite{edmt}} & 14 & 15 (\textcolor{green}{$\uparrow$} 1) & 15 (\textcolor{green}{$\uparrow$} 1) &  & 41.4 & 50 & 51.3 &  & 42.3 & 40.8 & 45.4 & 73.4 & 43.1 & 77.5\tabularnewline
\textbf{MHT\_bLSTM~\cite{mht_blstm}} & 15 & 24 (\textcolor{green}{$\uparrow$} 9) & 14 (\textcolor{red}{$\downarrow$} 1) &  & 41.1 & 47.5 & 51.9 &  & 40 & 42.5 & 42.5 & 74.4 & 46.9 & 73.5\tabularnewline
\textbf{AM\_ADM17~\cite{am_adm}} & 16 & 22 (\textcolor{green}{$\uparrow$} 6) & 13 (\textcolor{red}{$\downarrow$} 3) &  & 40.7 & 48.1 & 52.1 &  & 39.9 & 41.8 & 42.4 & 73.9 & 44.9 & 73.8\tabularnewline
\textbf{HAM\_SADF17~\cite{ham_intp15}} & 17 & 20 (\textcolor{green}{$\uparrow$} 3) & 16 (\textcolor{red}{$\downarrow$} 1) &  & 40.5 & 48.3 & 51.1 &  & 39.8 & 41.4 & 42.1 & 75 & 44.1 & 75.7\tabularnewline
\textbf{PHD\_GSDL17~\cite{phd_gsdl}} & 18 & 23 (\textcolor{green}{$\uparrow$} 5) & 18 (-) &  & 39.4 & 48 & 49.6 &  & 40.1 & 39 & 42.5 & 74.6 & 43.1 & 70.2\tabularnewline
\textbf{FWT\_17~\cite{fwt}} & 19 & 9 (\textcolor{red}{$\downarrow$} 10) & 22 (\textcolor{green}{$\uparrow$} 3) &  & 39.2 & 51.3 & 47.6 &  & 42.6 & 36.4 & 45.2 & 74.9 & 39.4 & 71.9\tabularnewline
\textbf{FAMNet~\cite{famnet}} & 20 & 5 (\textcolor{red}{$\downarrow$} 15) & 19 (\textcolor{red}{$\downarrow$} 1) &  & 39 & 52 & 48.7 &  & 41.7 & 36.8 & 43.7 & 75.9 & 40.6 & 69.2\tabularnewline
\textbf{MHT\_DAM\_17~\cite{mht_dam}} & 21 & 14 (\textcolor{red}{$\downarrow$} 7) & 23 (\textcolor{green}{$\uparrow$} 2) &  & 38.9 & 50.7 & 47.2 &  & 41.9 & 36.5 & 44.5 & 75.1 & 38 & 80\tabularnewline
\textbf{OTCD\_1\_17~\cite{otcd_1}} & 22 & 19 (\textcolor{red}{$\downarrow$} 3) & 21 (\textcolor{red}{$\downarrow$} 1) &  & 38.6 & 48.6 & 47.9 &  & 40 & 37.5 & 42.1 & 75.5 & 40.2 & 73.8\tabularnewline
\textbf{FPSN~\cite{fpsn}} & 23 & 28 (\textcolor{green}{$\uparrow$} 5) & 20 (\textcolor{red}{$\downarrow$} 3) &  & 38.2 & 44.9 & 48.4 &  & 38.8 & 38 & 41.8 & 71.9 & 41.3 & 70.2\tabularnewline
\textbf{GMPHDOGM17~\cite{gmphd_ogm}} & 24 & 16 (\textcolor{red}{$\downarrow$} 8) & 24 (-) &  & 38.2 & 49.9 & 47.1 &  & 41.7 & 35.2 & 44.2 & 74.9 & 40.8 & 61.5\tabularnewline
\textbf{MTDF17~\cite{mtdf}} & 25 & 17 (\textcolor{red}{$\downarrow$} 8) & 26 (\textcolor{green}{$\uparrow$} 1) &  & 37.8 & 49.6 & 45.2 &  & 42.5 & 34.2 & 45.8 & 71.8 & 36.1 & 75.9\tabularnewline
\textbf{MASS~\cite{mass}} & 26 & 25 (\textcolor{red}{$\downarrow$} 1) & 25 (\textcolor{red}{$\downarrow$} 1) &  & 36.9 & 46.9 & 46 &  & 39 & 35.2 & 41.6 & 73.1 & 38.4 & 70.2\tabularnewline
\textbf{LM\_NN\_17~\cite{lm_nn_17}} & 27 & 27 (-) & 27 (-) &  & 36.6 & 45.1 & 43.2 &  & 37.2 & 36.4 & 38.7 & 78.3 & 37.9 & 81.2\tabularnewline
\textbf{PHD\_GM~\cite{phd_gm}} & 28 & 18 (\textcolor{red}{$\downarrow$} 10) & 28 (-) &  & 36.2 & 48.8 & 43.2 &  & 41.1 & 32.1 & 43.7 & 74.2 & 34.9 & 69.5\tabularnewline
\textbf{EAMTT\_17~\cite{eamtt_17}} & 29 & 35 (\textcolor{green}{$\uparrow$} 6) & 29 (-) &  & 34.7 & 42.6 & 41.8 &  & 36.8 & 33.2 & 39.2 & 72.2 & 35.5 & 72.7\tabularnewline
\textbf{SORT17~\cite{sort17}} & 30 & 34 (\textcolor{green}{$\uparrow$} 4) & 30 (-) &  & 34.1 & 43.1 & 39.8 &  & 37.4 & 31.4 & 39.8 & 73.7 & 32.8 & 79.4\tabularnewline
\textbf{IOU17~\cite{iou}} & 31 & 26 (\textcolor{red}{$\downarrow$} 5) & 31 (-) &  & 33.7 & 45.5 & 39.4 &  & 38 & 30.2 & 40.1 & 74.8 & 32.6 & 73.7\tabularnewline
\textbf{HISP\_T17~\cite{hisp_t}} & 32 & 29 (\textcolor{red}{$\downarrow$} 3) & 33 (\textcolor{green}{$\uparrow$} 1) &  & 33.4 & 44.6 & 38.8 &  & 38.7 & 29.2 & 41.1 & 74 & 30.7 & 76.1\tabularnewline
\textbf{GMPHD\_SHA~\cite{gmphd_sha}} & 33 & 33 (-) & 32 (\textcolor{red}{$\downarrow$} 1) &  & 33.2 & 43.7 & 39.2 &  & 37 & 30.2 & 39.3 & 73.2 & 31.8 & 76.9\tabularnewline
\textbf{GMPHD\_DAL~\cite{gm_phd_dal}} & 34 & 30 (\textcolor{red}{$\downarrow$} 4) & 35 (\textcolor{green}{$\uparrow$} 1) &  & 31.5 & 44.4 & 36.2 &  & 38 & 26.4 & 40 & 75.3 & 28.9 & 69.6\tabularnewline
\textbf{GM\_PHD\_D~\cite{gm_phd}} & 35 & 32 (\textcolor{red}{$\downarrow$} 3) & 36 (\textcolor{green}{$\uparrow$} 1) &  & 30.7 & 44 & 34.2 &  & 38 & 25.1 & 40 & 75.3 & 26.3 & 75.9\tabularnewline
\textbf{GMPHD\_KCF~\cite{gmphd_kcf}} & 36 & 37 (\textcolor{green}{$\uparrow$} 1) & 34 (\textcolor{red}{$\downarrow$} 2) &  & 30.4 & 39.6 & 36.6 &  & 35.9 & 26.1 & 39.4 & 67.2 & 27.6 & 72.4\tabularnewline
\textbf{GMPHD\_N1Tr~\cite{gm_phd_n1t}} & 37 & 36 (\textcolor{red}{$\downarrow$} 1) & 37 (-) &  & 30.4 & 42.1 & 33.9 &  & 36.4 & 25.7 & 38.2 & 75.7 & 27 & 76.1\tabularnewline
\bottomrule
\end{tabular}

\caption{\textbf{Results on MOT17} for all published (peer reviewed) trackers.}
\label{Table:MOT17}
\end{table*}

\section{Evaluating Trackers with HOTA on MOTChallenge}
\label{sec:evaluation}

In order to see how HOTA compares to other metrics for real state-of-the-art trackers, we evaluated HOTA on trackers submitted to the \textit{MOTChallenge} MOT17 benchmark, and compared these HOTA scores with the MOTA and IDF1 scores. We cannot compare to Track-mAP because this metric requires trackers to supply confidence scores which is not the case for the MOTChallenge benchmark. 

We restrict our evaluation to only those methods that are published in peer reviewed journals and conferences.
We evaluate 37 different trackers \cite{mpntrack,etc,tracktor,ehaf,sas_mot,yoonkj,dman,jcc,nota,strn,jbnot,motdt,edmt,mht_blstm,am_adm,ham_intp15,phd_gsdl,fwt,famnet,mht_dam,otcd_1,fpsn,gmphd_ogm,mtdf,mass,lm_nn_17,phd_gm,eamtt_17,sort17,iou,hisp_t,gmphd_sha,gm_phd_dal,gm_phd,gmphd_kcf,gm_phd_n1t} on MOT17~\cite{MOT16}.
This includes all of the trackers for which the relevant bibliographic information was available when this analysis was performed on the 1st April 2020.

\PAR{Ranking Methods by HOTA.}
In Table~\ref{Table:MOT17} we show all of the published trackers, ranked according to our proposed HOTA metric. 
For fine-grained analysis, we also show the detection accuracy DetA, the association accuracy AssA, the detection recall DetRe, the detection precision DetPr, the association recall AssRe and the association precision AssPr. For a definition of these metrics, see Sec.~\ref{sec:decomposition}.
We also show the scores for MOTA and IDF1 metrics, and add an indicator to how the rankings change in compared to HOTA ranking.

\PAR{Comparing Trends across Metrics.}
In Fig.~\ref{trends} we show how the scores for different metrics and sub-metrics vary across all of the trackers that we evaluated on MOT17.

Fig.~\ref{trends} \textit{(left)} shows results for HOTA along with the two previously used metrics MOTA and IDF1. 
We observe, (i) that although these metrics do not always agree, they do follow a similar loose trend. 
(ii) it is clear according to all three metrics which trackers are performing well and which do not. 
This is not surprising, as all three metrics are aiming to quantify the quality of tracking. 
This can also be seen in Tab.~\ref{Table:MOT17} where the top-performing methods according to each metric is always within the top four methods over all other metrics. 
(iii) there are some significant differences in ranking between HOTA and both MOTA and IDF1. 
(iv) in general, HOTA aligns better in ranking with IDF1 than MOTA.
This is not surprising, as HOTA and IDF1 both aim to measure long-term tracking quality, whereas MOTA is only able to capture short-term tracking success. 
This is also reflected in the table, where the change in rankings for MOTA is larger compared to IDF1.

In Fig.~\ref{trends} \textit{(middle)}, the HOTA score is compared with its two major components, the DetA and AssA scores which measure detection success and association success, respectively. 
HOTA is computed as the geometric mean of the two and thus is always between the two values. 
We make the following observations. (i) both detection accuracy and association accuracy improve as trackers get better; (ii) top-performers are better at association than detection, while poor performing trackers are better at detection; 
(iii) there is larger variability over the association score over different trackers compared to the detection scores. This is expected as all trackers used the given public detections as input proposals. 

In Fig.~\ref{trends} \textit{(right)} we compare four different components of HOTA: the \textit{detection precision} and \textit{recall}, and \textit{association precision} and \textit{recall}. 
As can be seen, (i) precision values are higher than recall for both detection and association;  (ii) precision values are mostly within a similar range across all trackers, whereas recall values show an obvious trend to decrease as the tracker performance is dropping. 

\begin{figure*}
\centering
\includegraphics[width=0.22\linewidth]{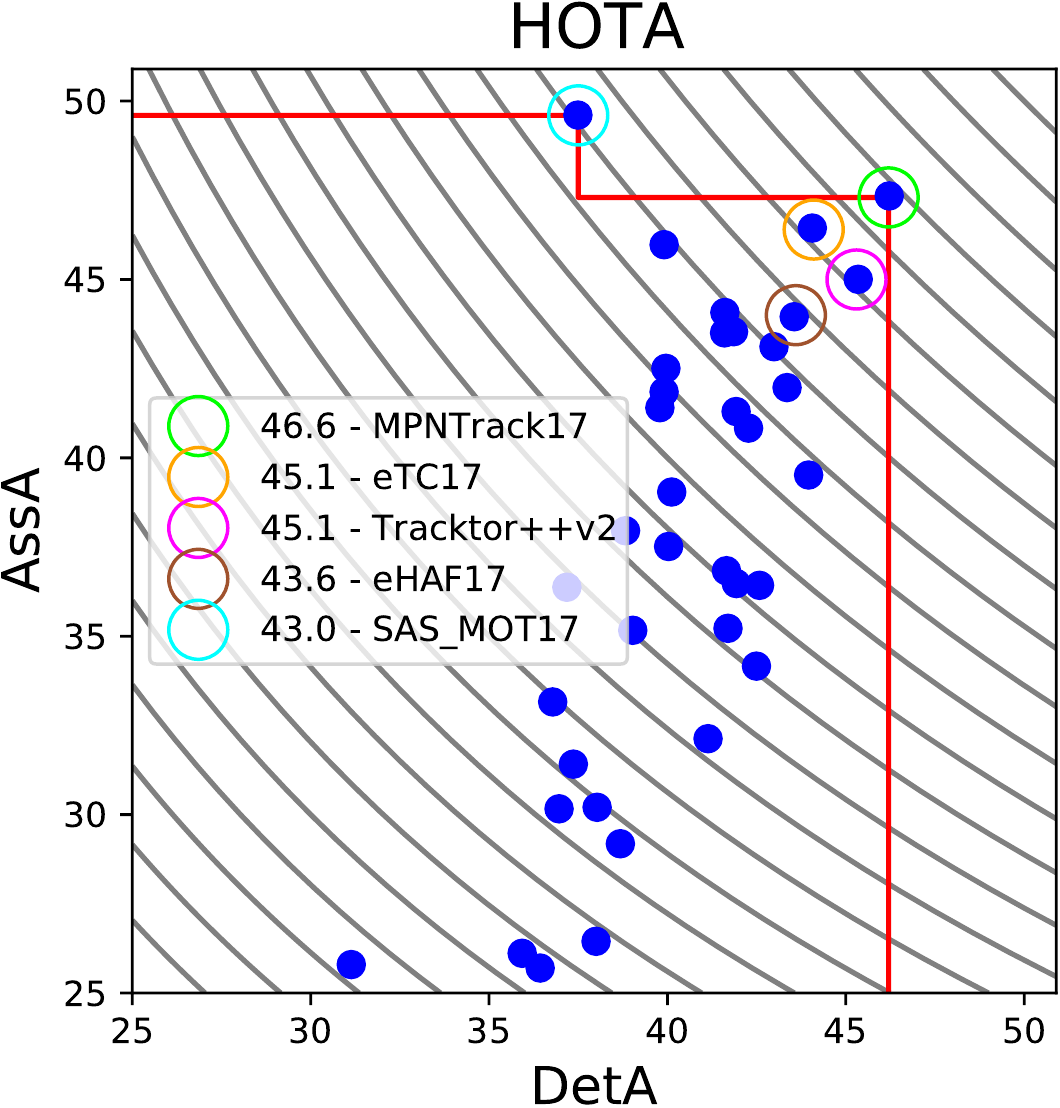}
\includegraphics[width=0.2345\linewidth]{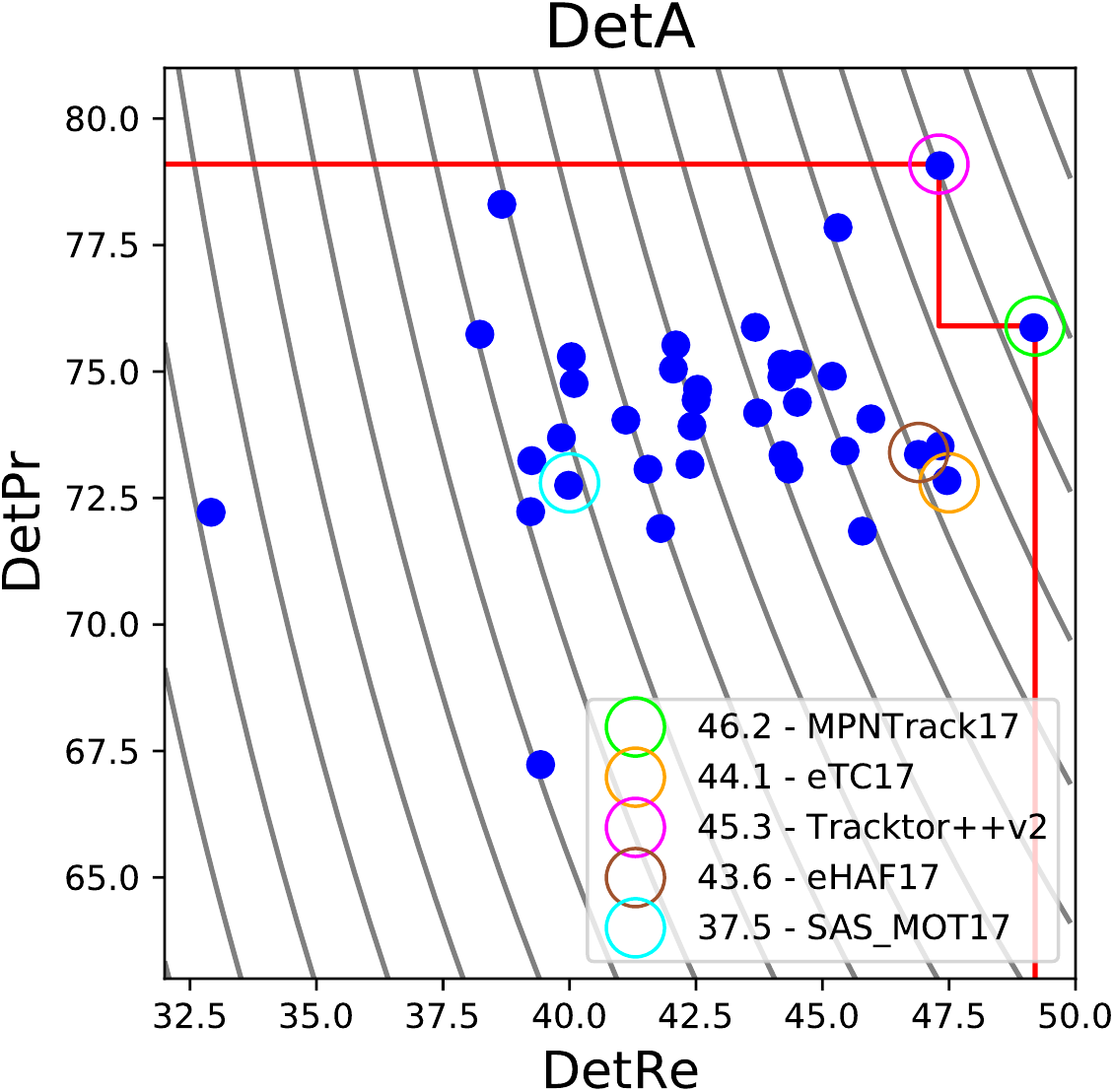}
\includegraphics[width=0.22\linewidth]{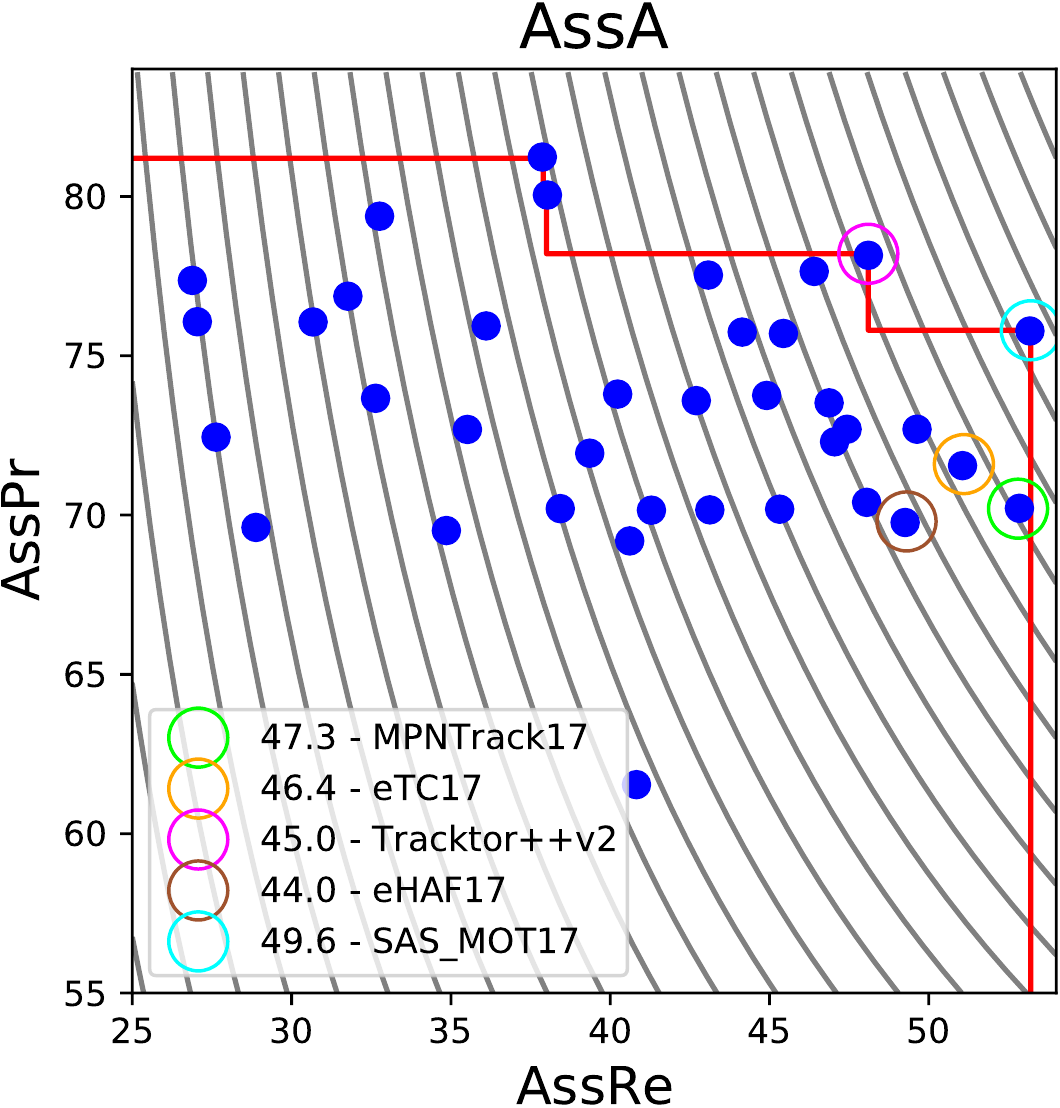}
\includegraphics[width=0.288\linewidth]{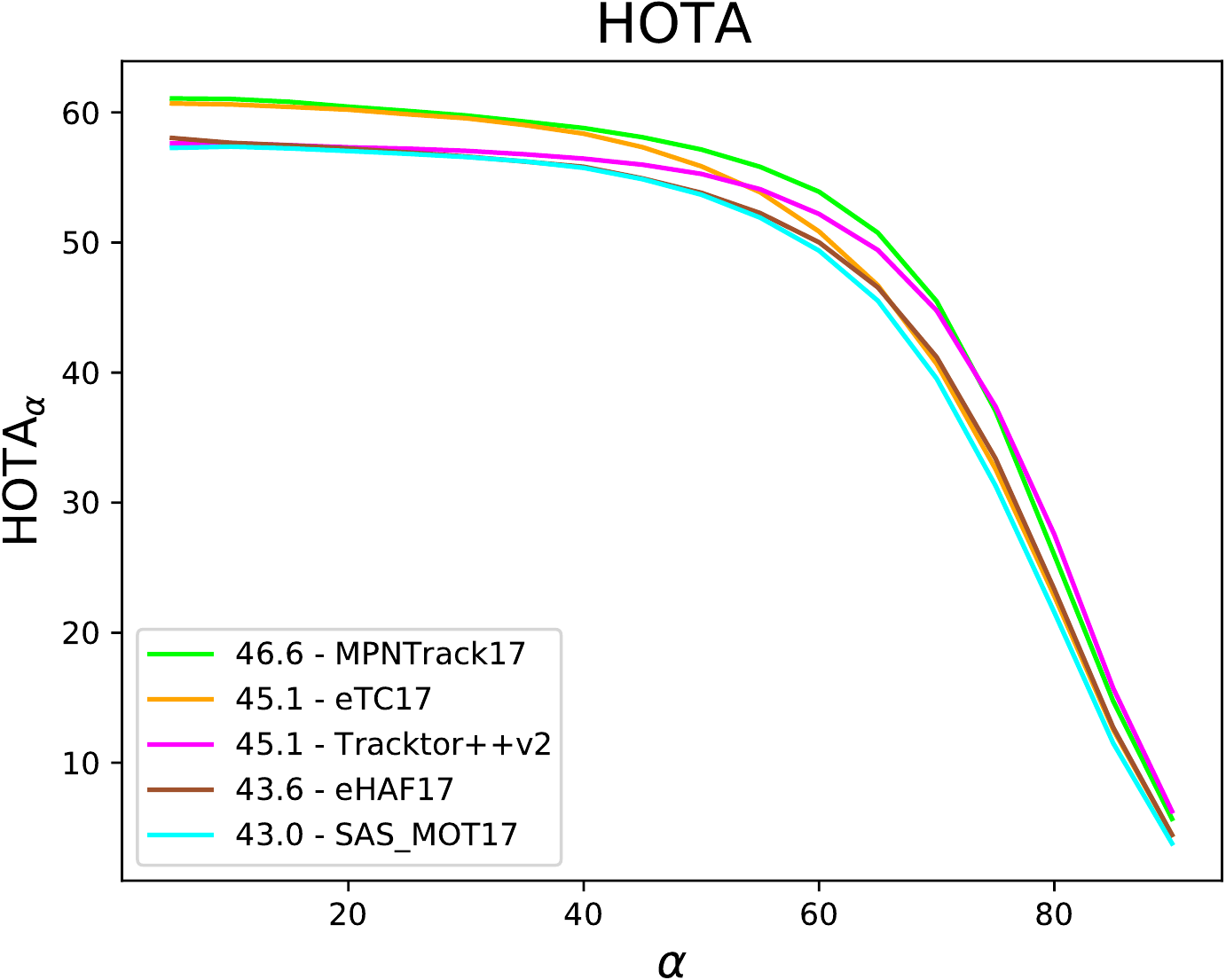}
\caption{Comparison between sub-metrics showing results for all peer reviewed trackers on MOT17. Each of the four plots shows a different decomposition of metrics into their corresponding sub-metrics for evaluating different aspects of tracking. The grey curves are level sets contours of constant score. The red staircase function shows the Pareto front. Only the top-5 trackers by HOTA are shown in the legend and far right plot for clarity.}
\label{fig:pareto}
\end{figure*}

\PAR{Analysing the State-of-the-Art in Multiple Dimensions.}
HOTA combines the different aspects of tracking in a balanced way suitable for ranking trackers. However it is also informative to compare trackers along all of the different dimensions of tracking. In Fig~\ref{fig:pareto} we compare trackers along a number of different dimensions within the sub-metric space of the HOTA family of metrics. 

This analysis allows one to clearly see the benefits and pitfalls of certain trackers, and allows for the selection of top performing trackers for different applications that may have different requirements. Any tracker along the multi-dimensional Pareto front can be considered to be state-of-the-art in at least one aspect of tracking performance. 

The fourth subplot shows how the HOTA score varies over the localisation threshold $\alpha$ for the top five ranked trackers. By showing performance over the range of all thresholds we are able to analyse and compare different properties of trackers that are not otherwise apparent by using a single evaluation score, such as which trackers perform very well when matches are allowed to be loosely localized and those that still perform well when a higher standard of localization is required.

What is also clear from this analysis is that the set of lowest-level sub-metrics (DetRe, DePr, AssRe, AssPr) are not enough on their own to tell the whole story about the results between different trackers. One is able to gain a greater level of understanding by examining the higher-level metrics which are combinations of these sub-metrics (DetA, AssA and HOTA). This highlights one of the key benefits of HOTA compared to previous evaluation approaches, that it simultaneously is able to measure different aspects of tracking performance, while being able to combine these together into unified representative scores.

\begin{figure}
\centering
\includegraphics[width=\linewidth]{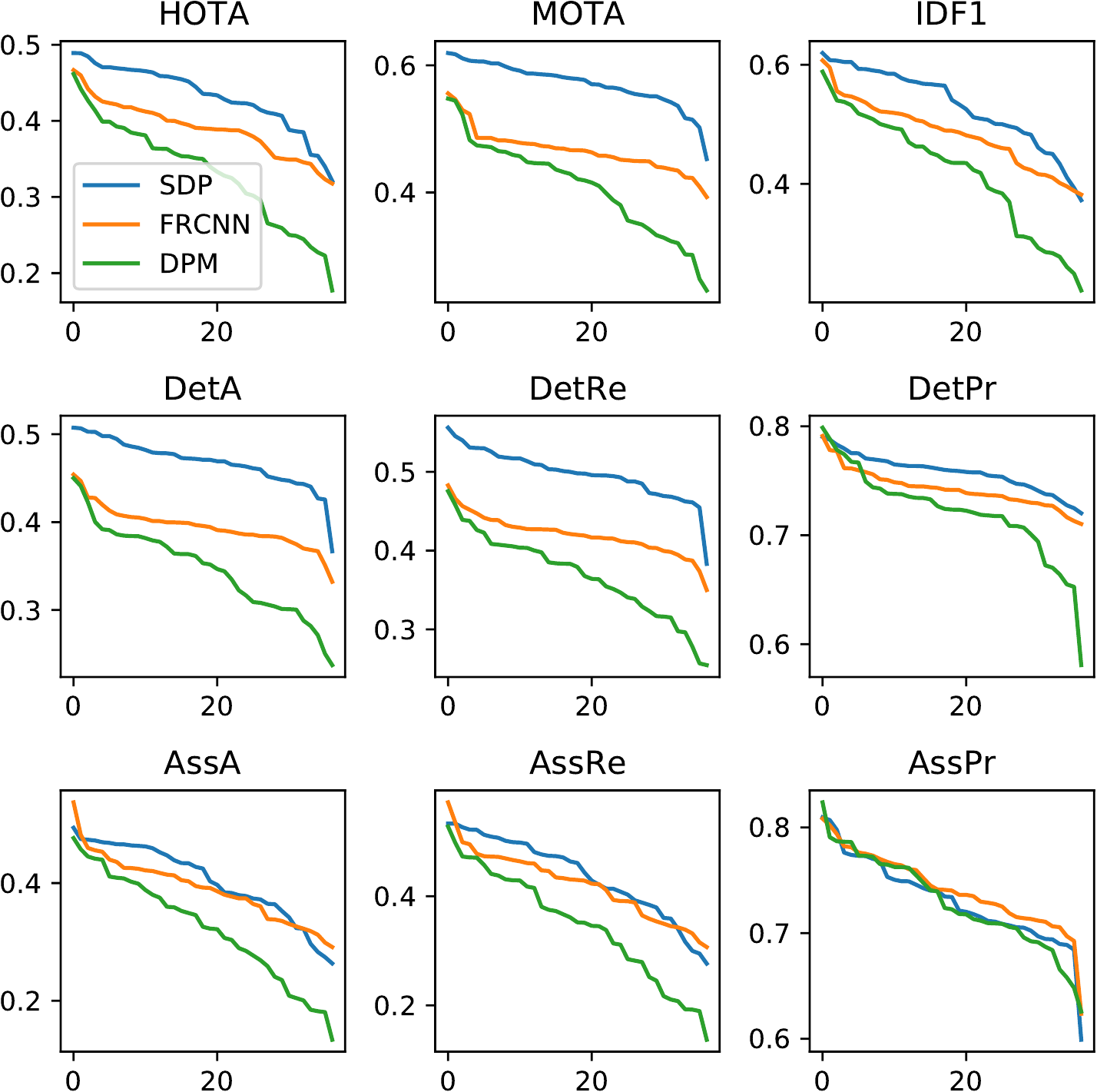}
\caption{Trends for various metrics when using different detectors as input on \textbf{MOT17}. All $37$ trackers from Table~\ref{Table:MOT17} are shown, and are ordered separately for each plot by the metric used.}
\label{detections}
\end{figure}

\PAR{Analyzing Metrics across Detectors.}
The MOT17 benchmark requires methods to produce results using the same tracking method with a set of three different input detections. Thus it is possible to analyze how different performance metrics behave when using different detectors. 
Fig.~\ref{detections} shows such an analysis for HOTA, MOTA and IDF1, as well as for all of the sub-metrics of HOTA. 
For all main metrics, using a better input detector improves the score. 
Of the three main metrics, MOTA is by the most affected by the choice of input detector. On the other hand, HOTA and IDF1 exhibit similar trends when using different detectors as input. 
In fact, MOTA exhibits similar trends to DetA. 
This is because, as discussed in Sec.~\ref{MOTA}, MOTA is mostly a proxy for detection accuracy and thus is highly correlated with DetA. 
As expected, the association scores are far less dependent on the detector input, although it can be seen that better detectors still aid better association. This is not surprising -- having more correct detections allows for more correct associations to be made.
Precision values for both detection and association are less affected by the choice of detector compared to recall values.

\PAR{Do the Metrics Disagree where we expect them too?}
In Sec.~\ref{sec:comparison} we laid out a number of theoretical problems of both the MOTA and IDF1 metrics and discussed how HOTA addresses these issues. 
In that analysis, we argued that MOTA and IDF1 are two ends of the metric spectrum, with HOTA being the middle ground. 
One of the main issues with MOTA is that it does not adequately score association and mostly only depends on detection accuracy, while IDF1 is exactly the opposite -- heavily relying on accurate association while exhibiting non-intuitive behavior with regards to detection quality. 

\begin{figure}
\centering
\includegraphics[width=\linewidth]{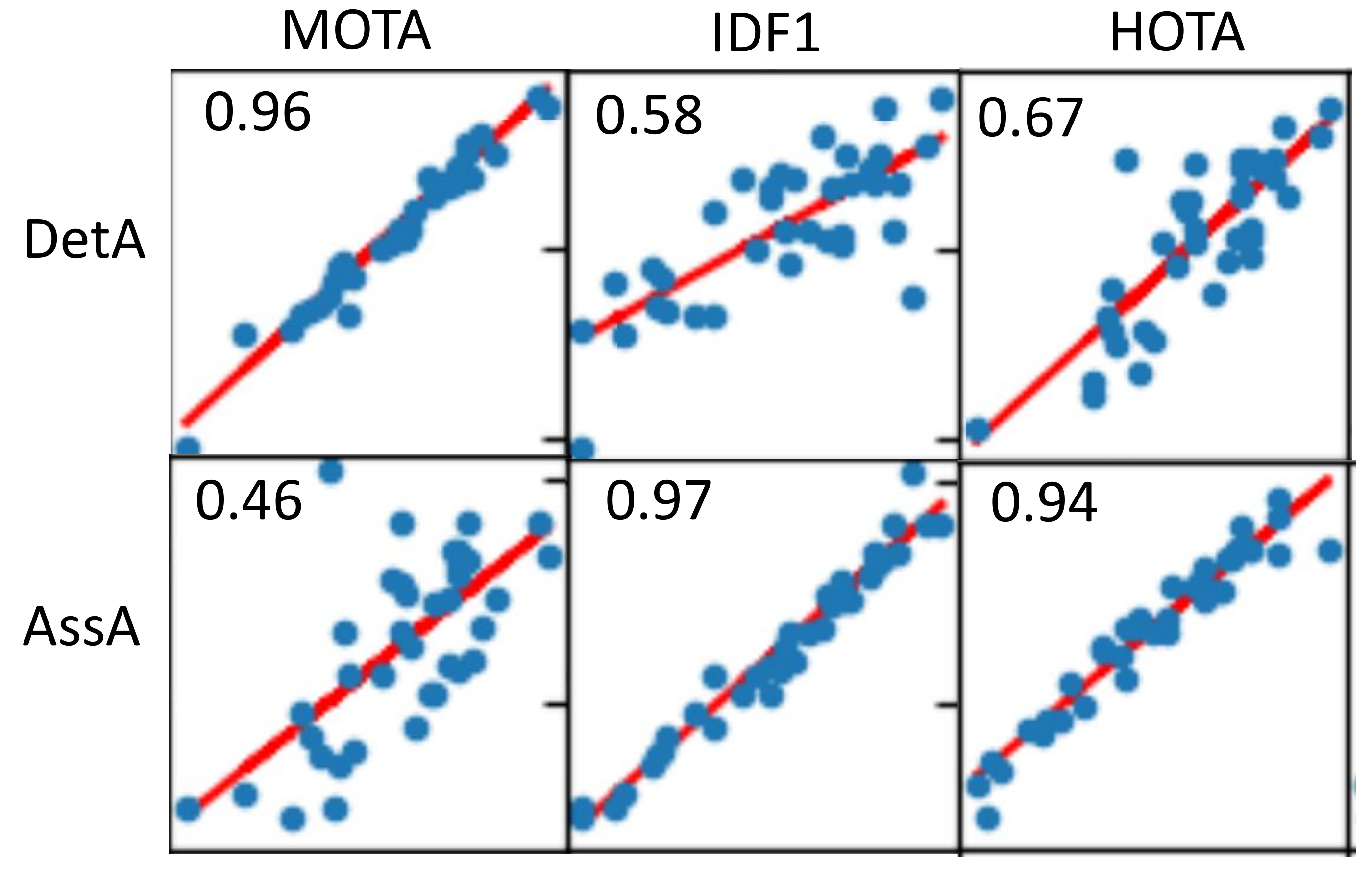}
\caption{Plotting each of the 3 main metrics against both the detection score and the association score for trackers on MOT17. The line of best fit is plotted in red, and the coefficient of determination (R$^2$) is shown in the top left of each plot.}
\label{new_correlation}
\end{figure}

In Fig.~\ref{new_correlation} we plot the MOTA, IDF1 and HOTA scores for all trackers on MOT17 against the DetA and AssA sub-scores, which measure the detection and association accuracy, respectively. 
We obtain the coefficient of determination (R$^2$) by fitting the line-of-best-fit and determining the strength of correlation between these metrics. 
As can be seen, the theoretical analysis of the weaknesses of MOTA and IDF1 is reflected in these results. 
Our observations are the following. 
(i) MOTA highly correlates with the detection score (0.96) while exhibiting low correlation with the association score (0.46). 
(ii) IDF1 exhibits almost the opposite behaviour, correlating strongly with the association score (0.97), but shows low correlation with the detection score (0.58). %
(iii) HOTA is between these two extremes, correlating reasonably strongly with both detection (0.67) and association (0.94). 
This explains why in many cases in Tab.~\ref{Table:MOT17}
the HOTA score causes trackers to increase in rank compared to MOTA while simultaneously decreasing in rank compared to IDF1 (and vice versa).

Note that the correlation is stronger for association than detection. 
This is not because association is assigned a higher weight by HOTA (both are weighted equally). 
The reason for this is that there is a wider range of association scores among the trackers, compared to the range of detection scores. 
This is to be expected, when all trackers are using a set of given public detections as input. 
Thus the variation in tracking scores is more likely to come from association than detection. 

Methods which HOTA ranks higher than MOTA but lower than IDF1 are those for which the association is more accurate than the detection. 
An example of this is SAS\_MOT17~\cite{sas_mot} which rises $26$ places in HOTA compared to MOTA but falls $2$ places compared to IDF1. 
This method specifically focuses on performing accurate association (and they also analyse how IDF1 is better correlated with better association than MOTA), at the cost of detection accuracy. 
Thus this method performs poorly on MOTA and very well according to IDF1, while performing somewhere in between according to HOTA. 

Methods which HOTA ranks higher than IDF1 but lower than MOTA are those for which the detection is more accurate than the association. 
An example of this is JBNOT~\cite{jbnot} which drops $8$ places compared to MOTA, but rises $5$ places \wrt. to IDF1.  This method focuses on improving detection recall, particularly during occlusions by using body joint detectors. 
However this does not perform association as well as many other methods. 
Therefore, this method ranks highly according to MOTA, but poorly according to IDF1. HOTA again places it in-between, taking both detection and association evenly into account.

\begin{figure*}
\centering
\includegraphics[width=0.85\linewidth]{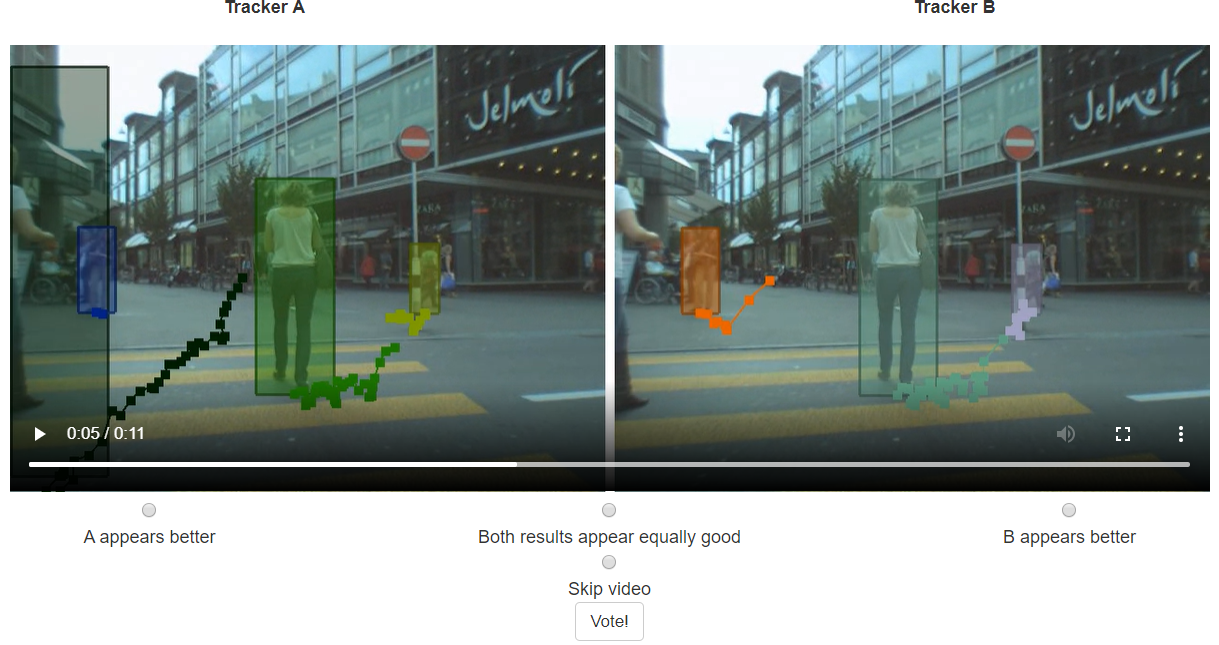}
\caption{An example of user-interface used for the user-experiment and the tracking visualisation.}
\label{ui}
\end{figure*}

\section{Human Visual Assessment Study}
\label{sec:humanstudy}

In previous sections we have provided theoretical analysis as to why HOTA is preferable to other metrics, as well as experimental analysis when using these metrics to evaluate real trackers. 
In this section we take this analysis one step further and perform a large-scale user-study in order to  determine how these metrics align with human judgment of tracking quality.
Our study follows many aspects of the design of \cite{TrackingTrackers} which previously attempted to evaluate tracking metrics using human evaluators. This study evaluated MOTA against a number of simple metrics such as Mostly Tracked, Detection Recall and MOTP. We wish to extend their analysis by running a study comparing HOTA, MOTA and IDF1.

Each of the different metrics has its own set of assumptions about what is important for tracking and evaluates against the best practices according to its own assumptions. Here we seek to answer the question of whether the assumptions for each metric align with the assumptions that humans make when viewing objects tracked through a video. While this is not a perfect proxy for the usefulness of these assumptions for any particular tracking application, it is nonetheless useful to know how each of these sets of assumptions aligns with human ranking.
Furthermore, by specifically recruiting MOT researchers to participate in our study, we are able to evaluate how the assumptions of each metric align with the assumptions of the community of people who would be using these metrics. We believe it is a useful property for metrics to evaluate tracking results in a way that is similar to how experts would rank the results when viewing them.

Such an experiment needs to be carefully designed, such that the experiment imposes as little bias to the results as possible. 
In order to conduct this study, we split the $7$ sequences of the test-set of MOT17 up into $6$ seconds chunks, which gives us $36$ short clips. 
We performed trial evaluations across video lengths of $3$, $6$ and $9$ seconds and found that the optimal span for this study was 6-second clips.
At $9$ seconds the video was too long for the human attention span to adequately judge tracking quality accurately. 
At $3$ seconds the videos were too short for the tracking results to be representative of meaningful tracking scenarios. 
We then evaluated all trackers on the MOT17 benchmark, across these $36$ six second video clips, and evaluated the MOTA, IDF1 and HOTA scores. 
In order to directly compare between metrics in the most user efficient way possible, we designed our experiment as a head-to-head comparison between metrics, such that the user will be presented with a pair of videos showing the tracking results of two different trackers. 
Each video pair consists of the results for two trackers where two of the metrics significantly disagree on which of the trackers performs better. 
Users are then to select which tracker performs better, thus agreeing with one of the two metrics in the head-to-head comparison. 
Users were also given the option to select that both trackers performed equally, or to skip a video pair.

\PAR{Determining the pairs.} 
We determined which pairs to show by first determining pairs that met our head-to-head requirements, and then by sub-selecting valid pairs based on diversity of sequences, detectors and trackers shown. 
Two trackers, A and B, are are a valid pair to compare for the two metrics M1 and M2 (for a particular sub-sequence and detector input) if they meet the following conditions:
\begin{align}
    & \text{S1} \cdot \text{S2} < 0, \;\;\;\;
     |\text{S1}| > 0.05, \;\;\;\;
     |\text{S2}| > 0.05
\end{align}
where S1 and S2 are defined as:
\begin{align}
    & \text{S1} = \text{M1}(\text{A}) - \text{M1}(\text{B}), \;\;\;\;
     \text{S2} = \text{M2}(\text{A}) - \text{M2}(\text{B}) 
\end{align}
The first constraint ensures that the two metrics disagree about which tracker is better (\eg, that the difference between scores for the trackers on one metric should have a different sign than for the other metric).
The second two constraints ensure that for both metrics there is a significant difference between the trackers (at least 0.05) so that any difference in ranking between the metrics is significant.

We evaluated 175 trackers on 108 unique combinations of sub-sequence and input detections (36 sub-sequences and three detection inputs). 
When comparing all pairs of trackers which met the above constraints there were hundreds of valid pairs per combination of metric. 
However we were aiming for a smaller set of videos for the user study, as we wished for each pair to be evaluated by a number of different users. 
We sub-sampled the pairs in such a way that we took at most one pair from each sub-sequence/detection combo and at most one pair that contained each tracker. 
We did this by greedily taking the pairs which minimise $S1 \times S2$, thus finding pairs for which the metrics maximally disagree, and for each chosen pair removing all pairs which contain the same tracker or the same sub-sequence/detection combination from the pool of valid pairs, and iterating greedily until there is no more pairs to choose from. 
This gave us $67$ pairs for HOTA vs MOTA, $51$ pairs for HOTA vs IDF1, and $69$ pairs for MOTA vs IDF1.

\PAR{Result Visualisation.} 
The way tracking results are displayed to users is critically important for such a user study. Depending on how they are displayed, different aspects of tracking could be emphasised. 
As we have seen in Sec.~\ref{sec:comparison} and \ref{sec:evaluation}, MOTA scores depend more on the quality of detection while IDF1 scores depend more on the quality of association, thus it is important that the visualisation method makes both types of potential errors as obvious as possible to users.

\begin{figure}
\centering
\includegraphics[width=\linewidth]{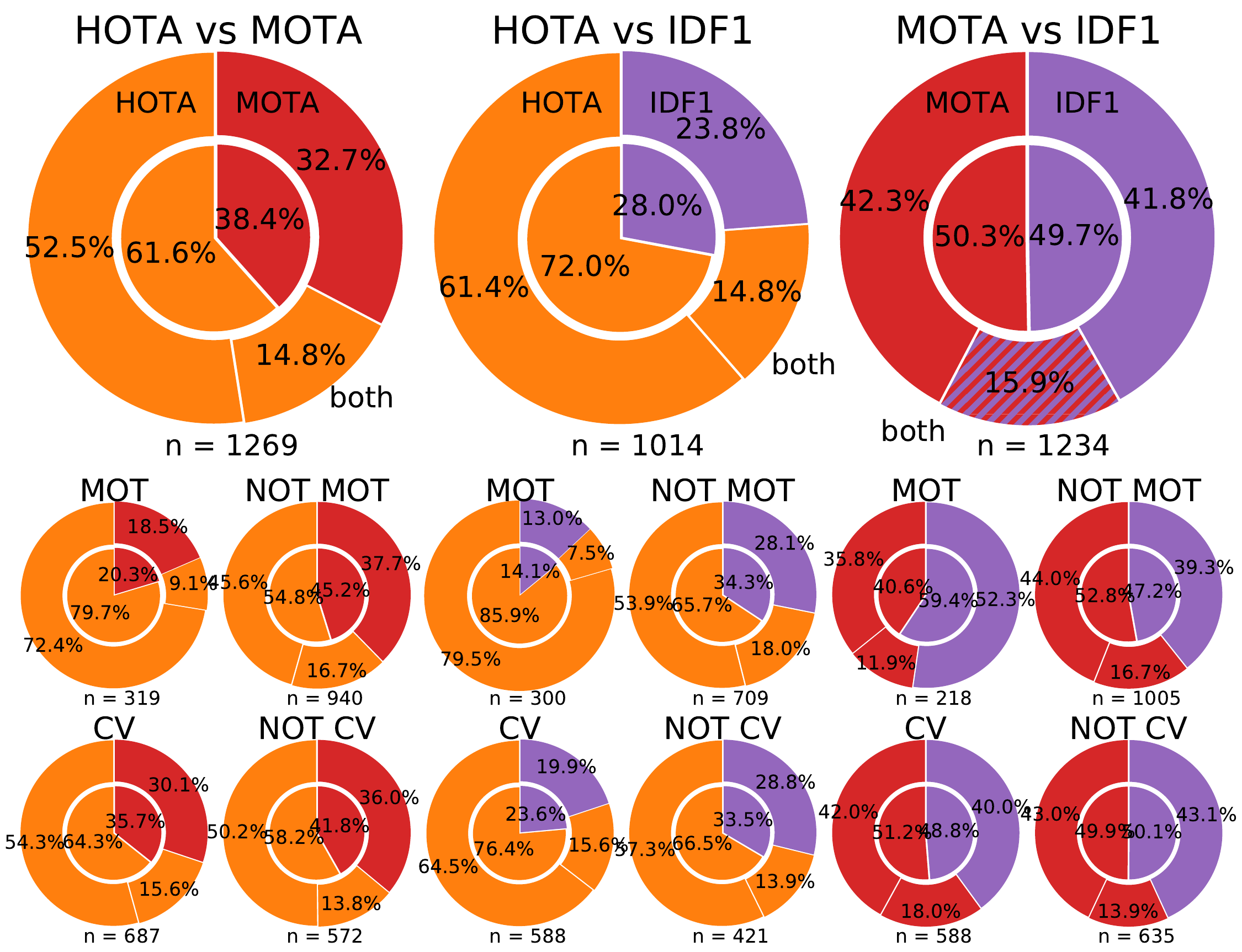}
\caption{Results of our user study showing which metrics better agree with human judgment. Experiments are set up in a head-to-head manner where for a pair of videos the two metrics in the pair disagree on which is better. Users selected the better tracking results, or alternatively rated both the same. \textit{Inner circles are the results excluding those that ranked both equally, while outer circles include these results.} `n=' shows the number of video pairs evaluated for each comparison. Results from all users are shown in the big circles. Results are also shown by user self-reported domain, using information about whether the user is a multi-object tracking researcher (MOT) and whether they are a computer vision researcher (CV).}
\label{userstudy}
\end{figure}

If the visualisation fairly balances the visual saliency of detection and association, then MOTA and IDF1 should perform equally well when compared head-to-head in the user-study, as each is representative of these two different error types. Fig.~\ref{userstudy} shows that this ends up being the case.

An example of our tracking visualisations (along with the user interface for rating trackers) can be seen in Figure~\ref{ui}. 

As can be seen we have made detection as obvious as possible by showing bounding boxes with both a thick colored border and a slightly transparent fill. 
We have also made association as obvious as possible by showing a tracking history of the bottom of each bounding box (in 2D pixel space) as coloured points with lines joining them. 
This history remains shown for the whole history of a track, and only disappears when that object is no longer present in the current frame. 
Such a visualisation style allows for both a quick understanding of the properties of trackers, as well as allowing for a conscientious user to take their time and understand all of the complex detections and associations present.

We play videos to users at half the natural frame-rate for easier video clarity. 
Users are also able to move around frames of the video by either clicking or dragging with the mouse, or by using the arrow keys. 
Pairs of trackers were shown to users in a random order, and users could evaluate as many pairs as they desired. 
The videos within each pair were shuffled so that the placement had no impact on which metric was under evaluation.

\PAR{Results of the User Study.}
We obtained user study results from $230$ participants, $62$ of which are multi-object tracking researchers, and $122$ of which are computer vision researchers. On average each user evaluated 9.02 pairs of trackers, for a total of $2075$ unique tracker comparisons. 
On average users took $2$ minutes and $13$ seconds to evaluate each tracking pair, spending on average $20$ minutes evaluating trackers. This is the equivalent of $80$ hours spent evaluating tracking results.

A visualisation of results of the head-to-head comparisons between trackers can be seen in Figure~\ref{userstudy}. Note that some pairs were used in multiple metric head-to-heads.
As can be seen, looking at all participants, HOTA outperforms MOTA by agreeing with human evaluators $61.6\%$ of the time compared to $38.4\%$ for MOTA (when excluding those that voted for both). 
In the comparison of HOTA and IDF1, human evaluators agree even more that HOTA is a better tracking metric, agreeing with HOTA $72.0\%$ of the time compared with only $28.0\%$ for IDF1. 
In the head-to-head for MOTA vs IDF1 each metric agreed with users around $50\%$ of the time.

For researchers who work in multi-object tracking the levels of agreement with the HOTA metric compared to both alternatives are much higher. 
Compared to MOTA, users agreed with HOTA $79.3\%$ of the time. 
Compared to IDF1, users agreed with HOTA $85.9\%$ of the time.

These results show that HOTA better aligns with human judgment of the accuracy of tracking results than previous metrics.
The fact that MOT researchers agree even more consistently with HOTA is a strong indication that HOTA is able to successfully evaluate trackers in a way that is relevant for the multi-object tracking community.
Evaluating trackers is a difficult task for humans, with often many objects present and extremely complex scenes. 
The `correct' answer is usually not obvious (as shown by users taking on average more than $2$ minutes per pair). 
However researchers in this field have experience working with such data and know what to look for in good tracking results. 
As such, the fact that HOTA agrees so strongly with the judgment of MOT researchers is a strong indication of the usefulness for the HOTA metric.

\section{Conclusion}
In this paper, we introduce HOTA (Higher Order Tracking Accuracy), a novel metric for evaluating multi-object tracking. 
Previously used metrics only capture part of what is important for tracking. 
MOTA is unable to capture association accurately. 
On the other hand, IDF1 and Track-mAP perform non-intuitively in regards to detection. 
HOTA tackles these problems with a simple, elegant formulation that equally weights detection and association accuracy. 

We argue analytically and experimentally why our proposed metric is preferable over the alternatives, testing HOTA using state-of-the-art trackers on the \textit{MOTChallenge} benchmark. 
Furthermore, we perform a large-scale user study and demonstrate that human visual assessment of tracking accuracy aligns better with HOTA compared to both MOTA and IDF1.

We believe that HOTA will change the nature of tracking research, laying the groundwork for new algorithms to be designed and benchmarked against a metric that measures both detection and association quality.

\begin{acknowledgements}
The authors would like to thank Tobias Fischer, Achal Dave, Pavel Tokmakov, Jack Valmadre, Alex Bewley, Joao Henriques and Fisher Yu, as well as the anonymous reviewers, for their helpful comments on our manuscript. 
For partial funding of this project, JL and BL would like to acknowledge the ERC Consolidator Grant DeeViSe (ERC-2017-COG-773161) and a Google Faculty Research Award. AO, PD and LLT would like to acknowledge the Humboldt Foundation through the Sofja
Kovalevskaja Award. PT would like to acknowledge CCAV project Streetwise and EPSRC/MURI grant EP/N019474/1.
\end{acknowledgements}

\bibliographystyle{spmpsci}      %
\bibliography{abbrev_short,paper}   %

\begin{thebibliography}{10}
\providecommand{\url}[1]{{#1}}
\providecommand{\urlprefix}{URL }
\expandafter\ifx\csname urlstyle\endcsname\relax
  \providecommand{\doi}[1]{DOI~\discretionary{}{}{}#1}\else
  \providecommand{\doi}{DOI~\discretionary{}{}{}\begingroup
  \urlstyle{rm}\Url}\fi

\bibitem{PoseTrack}
Andriluka, M., Iqbal, U., Insafutdinov, E., Pishchulin, L., Milan, A., Gall,
  J., Schiele, B.: Posetrack: A benchmark for human pose estimation and
  tracking.
\newblock In: Proceedings of the IEEE Conference on Computer Vision and Pattern
  Recognition, pp. 5167--5176 (2018)

\bibitem{lm_nn_17}
Babaee, M., Li, Z., Rigoll, G.: A dual cnn--rnn for multiple people tracking.
\newblock Neurocomputing \textbf{368}, 69--83 (2019)

\bibitem{hisp_t}
Baisa, N.L.: Online multi-target visual tracking using a hisp filter.
\newblock In: Proceedings of the 13th International Joint Conference on
  Computer Vision, Imaging and Computer Graphics Theory and Applications -
  Volume 5: VISAPP,, pp. 429--438 (2018)

\bibitem{gm_phd_dal}
{Baisa}, N.L.: Online multi-object visual tracking using a gm-phd filter with
  deep appearance learning.
\newblock In: 2019 22th International Conference on Information Fusion
  (FUSION), pp. 1--8 (2019)

\bibitem{gm_phd_n1t}
Baisa, N.L., Wallace, A.: Development of a n-type gm-phd filter for multiple
  target, multiple type visual tracking.
\newblock Journal of Visual Communication and Image Representation \textbf{59},
  257 -- 271 (2019)

\bibitem{OSPA-ST}
Bento, J., Zhu, J.J.: A metric for sets of trajectories that is practical and
  mathematically consistent.
\newblock arXiv preprint arXiv:1601.03094  (2016)

\bibitem{tracktor}
Bergmann, P., Meinhardt, T., Leal-Taix{\'e}, L.: Tracking without bells and
  whistles.
\newblock In: ICCV (2019)

\bibitem{CLEARMOT}
Bernardin, K., Stiefelhagen, R.: Evaluating multiple object tracking
  performance: the clear mot metrics.
\newblock EURASIP Journal on Image and Video Processing \textbf{2008}, 1--10
  (2008)

\bibitem{sort17}
Bewley, A., Ge, Z., Ott, L., Ramos, F., Upcroft, B.: Simple online and realtime
  tracking.
\newblock In: 2016 IEEE International Conference on Image Processing (ICIP),
  pp. 3464--3468 (2016)

\bibitem{iou}
Bochinski, E., Eiselein, V., Sikora, T.: High-speed tracking-by-detection
  without using image information.
\newblock In: AVSS (2017)

\bibitem{mpntrack}
Bras{\'o}, G., Leal-Taix{\'e}, L.: Learning a neural solver for multiple object
  tracking.
\newblock In: CVPR (2020)

\bibitem{nuScenes}
Caesar, H., Bankiti, V., Lang, A.H., Vora, S., Liong, V.E., Xu, Q., Krishnan,
  A., Pan, Y., Baldan, G., Beijbom, O.: nuscenes: A multimodal dataset for
  autonomous driving.
\newblock In: Proceedings of the IEEE/CVF Conference on Computer Vision and
  Pattern Recognition, pp. 11621--11631 (2020)

\bibitem{AMI}
Carletta, J., Ashby, S., Bourban, S., Flynn, M., Guillemot, M., Hain, T.,
  Kadlec, J., Karaiskos, V., Kraaij, W., Kronenthal, M., et~al.: The ami
  meeting corpus: A pre-announcement.
\newblock In: International workshop on machine learning for multimodal
  interaction, pp. 28--39. Springer (2005)

\bibitem{Argoverse}
Chang, M.F., Lambert, J., Sangkloy, P., Singh, J., Bak, S., Hartnett, A., Wang,
  D., Carr, P., Lucey, S., Ramanan, D., Hays, J.: Argoverse: 3d tracking and
  forecasting with rich maps.
\newblock In: The IEEE Conference on Computer Vision and Pattern Recognition
  (CVPR) (2019)

\bibitem{edmt}
Chen, J., Sheng, H., Zhang, Y., Xiong, Z.: Enhancing detection model for
  multiple hypothesis tracking.
\newblock In: BMTT Workshop (2017)

\bibitem{nota}
{Chen}, L., {Ai}, H., {Chen}, R., {Zhuang}, Z.: Aggregate tracklet appearance
  features for multi-object tracking.
\newblock IEEE Signal Processing Letters \textbf{26}(11), 1613--1617 (2019)

\bibitem{famnet}
Chu, P., Ling, H.: Famnet: Joint learning of feature, affinity and
  multi-dimensional assignment for online multiple object tracking.
\newblock In: ICCV (2019)

\bibitem{TAO}
Dave, A., Khurana, T., Tokmakov, P., Schmid, C., Ramanan, D.: Tao: A
  large-scale benchmark for tracking any object.
\newblock In: ECCV (2020)

\bibitem{MOT19}
Dendorfer, P., Rezatofighi, H., Milan, A., Shi, J., Cremers, D., Reid, I.,
  Roth, S., Schindler, K., Leal-Taix\'{e}, L.: {CVPR19} tracking and detection
  challenge: {H}ow crowded can it get?
\newblock arXiv:1906.04567 [cs]  (2019)

\bibitem{MOT20}
Dendorfer, P., Rezatofighi, H., Milan, A., Shi, J., Cremers, D., Reid, I.,
  Roth, S., Schindler, K., Leal-Taix\'{e}, L.: Mot20: A benchmark for multi
  object tracking in crowded scenes.
\newblock arXiv:2003.09003[cs]  (2020).
\newblock \urlprefix\url{http://arxiv.org/abs/1906.04567}.
\newblock ArXiv: 2003.09003

\bibitem{InfoMetric}
Edward, K.K., Matthew, P.D., Michael, B.H.: An information theoretic approach
  for tracker performance evaluation.
\newblock In: 2009 IEEE 12th International Conference on Computer Vision, pp.
  1523--1529. IEEE (2009)

\bibitem{gm_phd}
Eiselein, V., Arp, D., P{\"a}tzold, M., Sikora, T.: Real-time multi-human
  tracking using a probability hypothesis density filter and multiple
  detectors.
\newblock In: 2012 IEEE Ninth International Conference on Advanced Video and
  Signal-Based Surveillance (2012)

\bibitem{PETS}
Ellis, A., Ferryman, J.: Pets2010 and pets2009 evaluation of results using
  individual ground truthed single views.
\newblock In: 2010 7th IEEE International Conference on Advanced Video and
  Signal Based Surveillance, pp. 135--142. IEEE (2010)

\bibitem{pascalvoc}
Everingham, M., {Van Gool}, L., Williams, C., Winn, J., Zisserman, A.: The
  pascal visual object classes {(VOC)} challenge.
\newblock IJCV \textbf{88}(2), 303--338 (2010)

\bibitem{mtdf}
Fu, Z., Angelini, F., Chambers, J., Naqvi, S.M.: Multi-level cooperative fusion
  of gm-phd filters for online multiple human tracking.
\newblock IEEE Transactions on Multimedia \textbf{21}(9), 2277--2291 (2019)

\bibitem{phd_gsdl}
Fu, Z., Feng, P., Angelini, F., Chambers, J., Naqvi, S.: Particle phd filter
  based multiple human tracking using online group-structured dictionary
  learning.
\newblock In: IEEE Access (2018)

\bibitem{AveragingMetrics}
Fujita, O.: Metrics based on average distance between sets.
\newblock Japan Journal of Industrial and Applied Mathematics \textbf{30}(1),
  1--19 (2013)

\bibitem{KITTI}
Geiger, A., Lenz, P., Urtasun, R.: Are we ready for autonomous driving? the
  kitti vision benchmark suite.
\newblock In: Conference on Computer Vision and Pattern Recognition (CVPR)
  (2012)

\bibitem{fwt}
Henschel, R., Leal-Taix{\'e}, L., Cremers, D., Rosenhahn, B.: Fusion of head
  and full-body detectors for multi-object tracking.
\newblock Trajnet CVPRW  (2018)

\bibitem{jbnot}
Henschel, R., Zou, Y., Rosenhahn, B.: Multiple people tracking using body and
  joint detections.
\newblock CVPRW  (2019)

\bibitem{mass}
{Karunasekera}, H., {Wang}, H., {Zhang}, H.: Multiple object tracking with
  attention to appearance, structure, motion and size.
\newblock IEEE Access \textbf{7}, 104423--104434 (2019)

\bibitem{VACE}
Kasturi, R., Goldgof, D., Soundararajan, P., Manohar, V., Boonstra, M.,
  Korzhova, V.: Performance evaluation protocol for face, person and vehicle
  detection \& tracking in video analysis and content extraction (vace-ii).
\newblock Computer Science \& Engineering University of South Florida, Tampa
  (2006)

\bibitem{jcc}
Keuper, M., Tang, S., Andres, B., Brox, T., Schiele, B.: Motion segmentation
  amp; multiple object tracking by correlation co-clustering.
\newblock IEEE Transactions on Pattern Analysis and Machine Intelligence pp.
  1--1 (2018)

\bibitem{mht_dam}
Kim, C., Li, F., Ciptadi, A., Rehg, J.M.: Multiple hypothesis tracking
  revisited.
\newblock In: ICCV (2015)

\bibitem{mht_blstm}
Kim, C., Li, F., Rehg, J.M.: Multi-object tracking with neural gating using
  bilinear lstm.
\newblock In: ECCV (2018)

\bibitem{PanSeg}
Kirillov, A., He, K., Girshick, R., Rother, C., Doll{\'a}r, P.: Panoptic
  segmentation.
\newblock In: Proceedings of the IEEE conference on computer vision and pattern
  recognition, pp. 9404--9413 (2019)

\bibitem{gmphd_kcf}
Kutschbach, T., Bochinski, E., Eiselein, V., Sikora, T.: Sequential sensor
  fusion combining probability hypothesis density and kernelized correlation
  filters for multi-object tracking in video data.
\newblock In: AVSS (2017)

\bibitem{LIMA}
Layne, R., Hannuna, S., Camplani, M., Hall, J., Hospedales, T.M., Xiang, T.,
  Mirmehdi, M., Damen, D.: A dataset for persistent multi-target multi-camera
  tracking in rgb-d.
\newblock In: Proceedings of the IEEE Conference on Computer Vision and Pattern
  Recognition Workshops, pp. 47--55 (2017)

\bibitem{MOT15}
Leal-Taix\'{e}, L., Milan, A., Reid, I., Roth, S., Schindler, K.:
  {MOTC}hallenge 2015: {T}owards a benchmark for multi-target tracking.
\newblock arXiv:1504.01942 [cs]  (2015)

\bibitem{TrackingTrackers}
Leal-Taix{\'e}, L., Milan, A., Schindler, K., Cremers, D., Reid, I., Roth, S.:
  Tracking the trackers: an analysis of the state of the art in multiple object
  tracking.
\newblock arXiv preprint arXiv:1704.02781  (2017)

\bibitem{fpsn}
Lee, S., Kim, E.: Multiple object tracking via feature pyramid siamese
  networks.
\newblock IEEE access \textbf{7} (2018)

\bibitem{am_adm}
Lee, S.H., Kim, M.Y., Bae, S.H.: Learning discriminative appearance models for
  online multi-object tracking with appearance discriminability measures.
\newblock IEEE Access \textbf{6}, 67316--67328 (2018)

\bibitem{Monotonicity}
Leichter, I., Krupka, E.: Monotonicity and error type differentiability in
  performance measures for target detection and tracking in video.
\newblock IEEE transactions on pattern analysis and machine intelligence
  \textbf{35}(10), 2553--2560 (2013)

\bibitem{TRAJECTORY}
Li, Y., Huang, C., Nevatia, R.: Learning to associate: Hybridboosted
  multi-target tracker for crowded scene.
\newblock In: 2009 IEEE Conference on Computer Vision and Pattern Recognition,
  pp. 2953--2960. IEEE (2009)

\bibitem{COCO}
Lin, T.Y., Maire, M., Belongie, S., Hays, J., Perona, P., Ramanan, D.,
  Doll{\'a}r, P., Zitnick, C.L.: Microsoft coco: Common objects in context.
\newblock In: European conference on computer vision, pp. 740--755. Springer
  (2014)

\bibitem{otcd_1}
Liu, Q., Liu, B., Wu, Y., Li, W., Yu, N.: Real-time online multi-object
  tracking in compressed domain.
\newblock IEEE Access \textbf{7}, 76489--76499 (2019)

\bibitem{motdt}
Long, C., Haizhou, A., Zijie, Z., Chong, S.: Real-time multiple people tracking
  with deeply learned candidate selection and person re-identification.
\newblock In: ICME (2018)

\bibitem{MOTSFusion}
Luiten, J., Fischer, T., Leibe, B.: Track to reconstruct and reconstruct to
  track.
\newblock IEEE Robotics and Automation Letters \textbf{5}(2), 1803--1810 (2020)

\bibitem{WinningVIS}
Luiten, J., Torr, P., Leibe, B.: Video instance segmentation 2019: A winning
  approach for combined detection, segmentation, classification and tracking.
\newblock In: Proceedings of the IEEE International Conference on Computer
  Vision Workshops, pp. 0--0 (2019)

\bibitem{MOTlitreview}
Luo, W., Xing, J., Milan, A., Zhang, X., Liu, W., Zhao, X., Kim, T.K.: Multiple
  object tracking: A literature review.
\newblock arXiv preprint arXiv:1409.7618  (2014)

\bibitem{sas_mot}
Maksai, A., Fua, P.: Eliminating exposure bias and metric mismatch in multiple
  object tracking.
\newblock In: CVPR (2019)

\bibitem{NonMarkovianGloballyConsistantObjectTracking}
Maksai, A., Wang, X., Fleuret, F., Fua, P.: Non-markovian globally consistent
  multi-object tracking.
\newblock In: Proceedings of the IEEE International Conference on Computer
  Vision, pp. 2544--2554 (2017)

\bibitem{PETSvsVACE}
Manohar, V., Boonstra, M., Korzhova, V., Soundararajan, P., Goldgof, D.,
  Kasturi, R., Prasad, S., Raju, H., Bowers, R., Garofolo, J.: Pets vs. vace
  evaluation programs: A comparative study.
\newblock In: Proceedings of IEEE International Workshop on Performance
  Evaluation of Tracking and Surveillance, pp. 1--6 (2006)

\bibitem{MOT16}
Milan, A., Leal-Taix\'{e}, L., Reid, I., Roth, S., Schindler, K.: {MOT}16: {A}
  benchmark for multi-object tracking.
\newblock arXiv:1603.00831  (2016)

\bibitem{TrackingChallenges}
Milan, A., Schindler, K., Roth, S.: Challenges of ground truth evaluation of
  multi-target tracking.
\newblock In: Proceedings of the IEEE Conference on Computer Vision and Pattern
  Recognition Workshops (2013)

\bibitem{AICityChallenge}
Naphade, M., Anastasiu, D.C., Sharma, A., Jagrlamudi, V., Jeon, H., Liu, K.,
  Chang, M.C., Lyu, S., Gao, Z.: The nvidia ai city challenge.
\newblock In: 2017 IEEE SmartWorld, Ubiquitous Intelligence \& Computing,
  Advanced \& Trusted Computed, Scalable Computing \& Communications, Cloud \&
  Big Data Computing, Internet of People and Smart City Innovation
  (SmartWorld/SCALCOM/UIC/ATC/CBDCom/IOP/SCI), pp. 1--6. IEEE (2017)

\bibitem{ETISEO}
Nghiem, A.T., Bremond, F., Thonnat, M., Valentin, V.: Etiseo, performance
  evaluation for video surveillance systems.
\newblock In: 2007 IEEE Conference on Advanced Video and Signal Based
  Surveillance, pp. 476--481. IEEE (2007)

\bibitem{MDAM}
Rahmathullah, A.S., Garc{\'\i}a-Fern{\'a}ndez, {\'A}.F., Svensson, L.: A metric
  on the space of finite sets of trajectories for evaluation of multi-target
  tracking algorithms.
\newblock arXiv preprint arXiv:1605.01177  (2016)

\bibitem{70sTrack1}
Reid, D.: An algorithm for tracking multiple targets.
\newblock IEEE transactions on Automatic Control \textbf{24}(6), 843--854
  (1979)

\bibitem{IDF1}
Ristani, E., Solera, F., Zou, R., Cucchiara, R., Tomasi, C.: Performance
  measures and a data set for multi-target, multi-camera tracking.
\newblock In: European Conference on Computer Vision, pp. 17--35. Springer
  (2016)

\bibitem{ImageNet}
Russakovsky, O., Deng, J., Su, H., Krause, J., Satheesh, S., Ma, S., Huang, Z.,
  Karpathy, A., Khosla, A., Bernstein, M., Berg, A.C., Fei-Fei, L.: {ImageNet
  Large Scale Visual Recognition Challenge}.
\newblock International Journal of Computer Vision (IJCV)  (2015)

\bibitem{phd_gm}
Sanchez-Matilla, R., Cavallaro, A.: A predictor of moving objects for
  first-person vision.
\newblock In: IEEE International Conference on Image Processing (ICIP) (2019)

\bibitem{eamtt_17}
Sanchez-Matilla, R., Poiesi, F., Cavallaro, A.: Online multi-target tracking
  with strong and weak detections.
\newblock In: Computer Vision -- ECCV 2016 Workshops, pp. 84--99 (2016)

\bibitem{ehaf}
Sheng, H., Zhang, Y., Chen, J., Xiong, Z., Zhang, J.: Heterogeneous association
  graph fusion for target association in multiple object tracking.
\newblock IEEE Transactions on Circuits and Systems for Video Technology
  (2018)

\bibitem{GMME}
Shitrit, H.B., Berclaz, J., Fleuret, F., Fua, P.: Tracking multiple people
  under global appearance constraints.
\newblock In: 2011 International Conference on Computer Vision, pp. 137--144.
  IEEE (2011)

\bibitem{70sTrack2}
Singer, R., Sea, R., Housewright, K.: Derivation and evaluation of improved
  tracking filter for use in dense multitarget environments.
\newblock IEEE Transactions on Information Theory \textbf{20}(4), 423--432
  (1974)

\bibitem{PurityMetric}
Smith, K., Gatica-Perez, D., Odobez, J.M., Ba, S.: Evaluating multi-object
  tracking.
\newblock In: 2005 IEEE Computer Society Conference on Computer Vision and
  Pattern Recognition (CVPR'05)-Workshops, pp. 36--36. IEEE (2005)

\bibitem{70sTrack4}
Smith, P., Buechler, G.: A branching algorithm for discriminating and tracking
  multiple objects.
\newblock IEEE Transactions on Automatic Control \textbf{20}(1), 101--104
  (1975)

\bibitem{gmphd_sha}
Song, Y., Jeon, M.: Online multiple object tracking with the hierarchically
  adopted gm-phd filter using motion and appearance.
\newblock IEEE/IEIE The International Conference on Consumer Electronics (ICCE)
  Asia  (2016)

\bibitem{gmphd_ogm}
Song, Y., Yoon, K., Yoon, Y., Yow, K., Jeon, M.: Online multi-object tracking
  with gmphd filter and occlusion group management.
\newblock IEEE Access  (2019)

\bibitem{70sTrack3}
Stein, J., Blackman, S.: Generalized correlation of multi-target track data.
\newblock IEEE Transactions on Aerospace and Electronic Systems \textbf{11}(6),
  1207--1217 (1975)

\bibitem{CLEARworkshop}
Stiefelhagen, R., Bernardin, K., Bowers, R., Garofolo, J., Mostefa, D.,
  Soundararajan, P.: The clear 2006 evaluation.
\newblock In: International evaluation workshop on classification of events,
  activities and relationships, pp. 1--44. Springer (2006)

\bibitem{Waymo}
Sun, P., Kretzschmar, H., Dotiwalla, X., Chouard, A., Patnaik, V., Tsui, P.,
  Guo, J., Zhou, Y., Chai, Y., Caine, B., et~al.: Scalability in perception for
  autonomous driving: Waymo open dataset.
\newblock arXiv pp. arXiv--1912 (2019)

\bibitem{MOTS}
Voigtlaender, P., Krause, M., Osep, A., Luiten, J., Sekar, B.B.G., Geiger, A.,
  Leibe, B.: Mots: Multi-object tracking and segmentation.
\newblock In: Proceedings of the IEEE Conference on Computer Vision and Pattern
  Recognition (2019)

\bibitem{CHIL}
Waibel, A., Steusloff, H., Stiefelhagen, R., Watson, K.: Computers in the human
  interaction loop.
\newblock In: Computers in the Human Interaction Loop, pp. 3--6. Springer
  (2009)

\bibitem{etc}
Wang, G., Wang, Y., Zhang, H., Gu, R., Hwang, J.N.: Exploit the connectivity:
  Multi-object tracking with trackletnet.
\newblock In: ACM International Conference on Multimedia, pp. 482--490 (2019)

\bibitem{PANDA}
Wang, X., Zhang, X., Zhu, Y., Guo, Y., Yuan, X., Xiang, L., Wang, Z., Ding, G.,
  Brady, D., Dai, Q., et~al.: Panda: A gigapixel-level human-centric video
  dataset.
\newblock In: Proceedings of the IEEE/CVF Conference on Computer Vision and
  Pattern Recognition, pp. 3268--3278 (2020)

\bibitem{UADETRAC}
Wen, L., Du, D., Cai, Z., Lei, Z., Chang, M.C., Qi, H., Lim, J., Yang, M.H.,
  Lyu, S.: Ua-detrac: A new benchmark and protocol for multi-object detection
  and tracking.
\newblock Computer Vision and Image Understanding \textbf{193}, 102907 (2020)

\bibitem{AMOTA}
Weng, X., Wang, J., Held, D., Kitani, K.: 3d multi-object tracking: A baseline
  and new evaluation metrics.
\newblock In: IROS (2020)

\bibitem{TrajectoryOrig}
Wu, B., Nevatia, R.: Tracking of multiple, partially occluded humans based on
  static body part detection.
\newblock In: 2006 IEEE Computer Society Conference on Computer Vision and
  Pattern Recognition (CVPR'06), vol.~1, pp. 951--958. IEEE (2006)

\bibitem{TrackClustering}
Wu, C.W., Zhong, M.T., Tsao, Y., Yang, S.W., Chen, Y.K., Chien, S.Y.:
  Track-clustering error evaluation for track-based multi-camera tracking
  system employing human re-identification.
\newblock In: Proceedings of the IEEE Conference on Computer Vision and Pattern
  Recognition Workshops, pp. 1--9 (2017)

\bibitem{strn}
Xu, J., Cao, Y., Zhang, Z., Hu, H.: Spatial-temporal relation networks for
  multi-object tracking.
\newblock ICCV  (2019)

\bibitem{VIS}
Yang, L., Fan, Y., Xu, N.: Video instance segmentation.
\newblock In: Proceedings of the IEEE International Conference on Computer
  Vision, pp. 5188--5197 (2019)

\bibitem{yoonkj}
YOON, K., GWAK, J., SONG, Y.M., YOON, Y.C., JEON, M.: Oneshotda: Online
  multi-object tracker with one-shot-learning-based data association.
\newblock IEEE Access \textbf{8}, 38060--38072 (2020)

\bibitem{ham_intp15}
Yoon, Y., Boragule, A., Song, Y., Yoon, K., Jeon, M.: Online multi-object
  tracking with historical appearance matching and scene adaptive detection
  filtering.
\newblock In: IEEE AVSS (2018)

\bibitem{PETSmetrics}
Young, D.P., Ferryman, J.M.: Pets metrics: On-line performance evaluation
  service.
\newblock In: 2005 IEEE International Workshop on Visual Surveillance and
  Performance Evaluation of Tracking and Surveillance, pp. 317--324. IEEE
  (2005)

\bibitem{BDD}
Yu, F., Chen, H., Wang, X., Xian, W., Chen, Y., Liu, F., Madhavan, V., Darrell,
  T.: Bdd100k: A diverse driving dataset for heterogeneous multitask learning.
\newblock In: Proceedings of the IEEE/CVF Conference on Computer Vision and
  Pattern Recognition, pp. 2636--2645 (2020)

\bibitem{SolutionPath}
Yu, S.I., Meng, D., Zuo, W., Hauptmann, A.: The solution path algorithm for
  identity-aware multi-object tracking.
\newblock In: Proceedings of the IEEE conference on computer vision and pattern
  recognition, pp. 3871--3879 (2016)

\bibitem{dman}
Zhu, J., Yang, H., Liu, N., Kim, M., Zhang, W., Yang, M.H.: Online multi-object
  tracking with dual matching attention networks.
\newblock In: ECCV (2018)

\bibitem{VisDrone}
Zhu, P., Wen, L., Du, D., Bian, X., Hu, Q., Ling, H.: Vision meets drones:
  Past, present and future.
\newblock arXiv preprint arXiv:2001.06303  (2020)

\end{thebibliography}

\end{document}